\documentclass{bmvc2k}

\usepackage{tikz}
\usetikzlibrary{shapes}
\usetikzlibrary{arrows, arrows.meta}
\usepackage{dirtytalk}
\usepackage{amsmath}
\usepackage{amssymb}
\usepackage{dsfont}
\usepackage{bm}
\usepackage{subcaption}
\usepackage{graphicx}
\usepackage{booktabs} 
\usepackage{tabularx}
\usepackage{multirow}

\title{Selection of Source Images Heavily Influences the Effectiveness of\\Adversarial Attacks}

\addauthor{Utku Ozbulak}{utku.ozbulak@ugent.be}{1}
\addauthor{Esla Timothy Anzaku}{eslatimothy.anzaku@ghent.ac.kr}{1}
\addauthor{Wesley De Neve}{wesley.deneve@ugent.be}{1}
\addauthor{Arnout Van Messem}{arnout.vanmessem@uliege.be}{2}

\addinstitution{
 Ghent University\\
 Ghent, Belgium\\
 \vspace{0.25em}
 Ghent University Global Campus\\
 Incheon, South Korea
 \vspace{0.5em}
}
\addinstitution{
 University of Li\`ege\\
 Li\`ege, Belgium
}

\runninghead{Ozbulak et al.}{Influence of source images on adversarial attacks}

\begin{document}

\maketitle

\begin{abstract}
Although the adoption rate of deep neural networks (DNNs) has tremendously increased in recent years, a solution for their vulnerability against adversarial examples has not yet been found. As a result, substantial research efforts are dedicated to fix this weakness, with many studies typically using a subset of source images to generate adversarial examples, treating every image in this subset as equal. We demonstrate that, in fact, not every source image is equally suited for this kind of assessment. To do so, we devise a large-scale model-to-model transferability scenario for which we meticulously analyze the properties of adversarial examples, generated from every suitable source image in ImageNet by making use of three of the most frequently deployed attacks. In this transferability scenario, which involves seven distinct DNN models, including the recently proposed vision transformers, we reveal that it is possible to have a difference of up to $12.5\%$ in model-to-model transferability success, $1.01$ in average $L_2$ perturbation, and $0.03$ ($8/225$) in average $L_{\infty}$ perturbation when $1,000$ source images are sampled randomly among all suitable candidates. We then take one of the first steps in evaluating the robustness of images used to create adversarial examples, proposing a number of simple but effective methods to identify unsuitable source images, thus making it possible to mitigate extreme cases in experimentation and support high-quality benchmarking. In support of future research efforts, we make our code and the statistics for all evaluated source images  as well as the list of identified fragile source images publicly available in \url{https://github.com/utkuozbulak/imagenet-adversarial-image-evaluation}.
\end{abstract}

\vspace{-1em}
\section{Introduction}
\label{sec:intro}
\vspace{-0.5em}

Thanks to recent advances in the field of deep neural networks, a wide range of problems that were once thought to be hard challenges found easy-to-adopt solutions~\cite{Alexnet,attention}. Indeed, many deep learning libraries now come with built-in solutions and pre-trained models, further increasing the adoption rate of such networks in the area of computer vision~\cite{tensorflow2015-whitepaper,scipy_cite,pytorch-whitepaper}. In spite of receiving a large amount of research attention, a number of fundamental flaws of DNNs still remain unsolved. One of those flaws is their vulnerability to adversarial attacks, where small changes in inputs may lead to large changes in predictions~\cite{LBFGS}. 

\begin{figure}[t]
\centering
\begin{tikzpicture}[thick,scale=0.6, every node/.style={scale=0.9}]
\scriptsize
    \draw[line width=0.5mm] (0, 0) -- (13.3, 0);
    \node[align=center] at (6.65, -1.3) {\footnotesize Number of source images used from ImageNet to create adversarial examples};
    \def\xsize{2.6} 
    \def\x{0.2} 
    \draw (\x, -0.3) -- (\x, 0.1);
    \draw (\x, -0.3) -- (\x+\xsize, -0.3);
    \draw (\x+\xsize, -0.3) -- (\x+\xsize, 0.1);
    \node[align=center,rotate=0] at (\x + \xsize/2, -0.5) {\scriptsize 1 - 500};    
    \def\xsize{4.9}
    \def\x{3} 
    \draw (\x, -0.3) -- (\x, 0.1);
    \draw (\x, -0.3) -- (\x+\xsize, -0.3);
    \draw (\x+\xsize, -0.3) -- (\x+\xsize, 0.1);
    \node[align=center,rotate=0] at (\x + \xsize/2, -0.5) {\scriptsize 501 - 2,000}; 
    \def\xsize{2.4}
    \def\x{8.1} 
    \draw (\x, -0.3) -- (\x, 0.1);
    \draw (\x, -0.3) -- (\x+\xsize, -0.3);
    \draw (\x+\xsize, -0.3) -- (\x+\xsize, 0.1);
    \node[align=center,rotate=0] at (\x + \xsize/2, -0.5) {\scriptsize 2,001 - 10,000};   
    \def\x{10.7} 
    \def\xsize{2.4}
    \draw (\x, -0.3) -- (\x, 0.1);
    \draw (\x, -0.3) -- (\x+\xsize, -0.3);
    \draw (\x+\xsize, -0.3) -- (\x+\xsize, 0.1);
    \node[align=center,rotate=0] at (\x + \xsize/2, -0.5) {\scriptsize All or more data};   
    \node[align=left] at (0.5, 2)  {\rotatebox{90}{{\citet{chen2017ead}}}};
    \node[align=left] at (1, 2)  {\rotatebox{90}{\citet{xu2018structured_new_local_adv_attack}}};
    \node[align=left] at (1.5, 2)  {\rotatebox{90}{\citet{LAVAN}}};
    \node[align=left] at (2, 2)  {\rotatebox{90}{\citet{croce2019sparse}}};
    \node[align=left] at (2.5, 2)  {\rotatebox{90}{\citet{finlay2019logbarrier}}};
    \node[align=left] at (3.1, 2)  {\rotatebox{90}{\citet{baluja2017adversarial}}};
    \node[align=left] at (3.6, 2)  {\rotatebox{90}{\citet{ilyas2018black_black_black_box}}};
    \node[align=left] at (4.1, 2)  {\rotatebox{90}{\citet{su2018robustness_18_imagenet_models_evaluation}}};
    \node[align=left] at (4.6, 2)  {\rotatebox{90}{\citet{brendel2019accurate}}};
    \node[align=left] at (5.1, 2)  {\rotatebox{90}{\citet{hu2019new}}};
    \node[align=left] at (5.6, 2)  {\rotatebox{90}{\citet{guo2019simple_black_black_box}}};
    \node[align=left] at (6.1, 2)  {\rotatebox{90}{\citet{input_transform_def3}}};
    \node[align=left] at (6.6, 2)  {\rotatebox{90}{\citet{zhao2020towards_large_yet_imperceivable}}};
    \node[align=left] at (7.1, 2)  {\rotatebox{90}{\citet{ozbulak2020perturbation}}};
    \node[align=left] at (7.6, 2)  {\rotatebox{90}{\citet{dong2020benchmarking}}};
    \node[align=left] at (8.2, 2)  {\rotatebox{90}{\citet{xie2017mitigating}}};
    \node[align=left] at (8.7, 2)  {\rotatebox{90}{\citet{adv_def_blur_input}}};
    \node[align=left] at (9.2, 2)  {\rotatebox{90}{\citet{liao2018defense}}};
    \node[align=left] at (9.7, 2)  {\rotatebox{90}{\citet{skip_connections_easier}}};
    \node[align=left] at (10.2, 2)  {\rotatebox{90}{\citet{shamsabadi2020colorfool}}};
    \node[align=left] at (10.8, 2)  {\rotatebox{90}{\scriptsize Dezfooli et al. \cite{moosavi2017universal}}};
    \node[align=left] at (11.3, 2)  {\rotatebox{90}{\citet{IFGS}}};
    \node[align=left] at (11.8, 2)  {\rotatebox{90}{\citet{kannan2018adversarial}}};
    \node[align=left] at (12.3, 2)  {\rotatebox{90}{\citet{yan2018deep}}};
    \node[align=left] at (12.8, 2)  {\rotatebox{90}{\citet{xie2020adversarial}}};
\end{tikzpicture}
\vspace{1em}
\caption{A number of studies that work with images taken from the ImageNet validation set, grouped based on the number of source images used for creating adversarial examples.}
\label{fig:imagenet_data}
\vspace{-1em}
\end{figure}
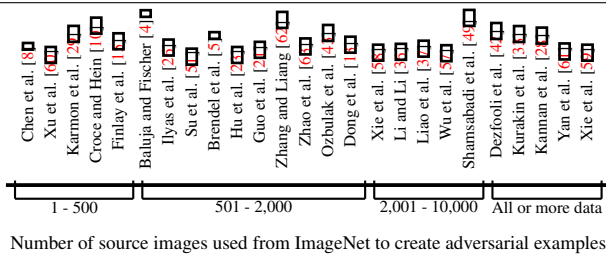

Although adversarial attacks have been recognized to be a threat for all domains that make use of DNNs, the domain of vision in particular is said to be the one that suffers from adversarial attacks the most, since the perturbation is often invisible to the bare eye. Moreover, continuous deployment of DNNs for mission-critical tasks such as self-driving cars and medical diagnosis tools further amplify this threat since the adversarial examples are not easily detectable~\cite{self_driving_adversarial,medical_classification_adv,self_driving_car_seg_adv,medical_diagnosis_robust}. 

In recent years, numerous adversarial defenses were proposed in order to prevent adversarial attacks or detect adversarial examples~\cite{logit_def5_grosse2017statistical,logit_def3_koo2019hawkeye,adv_def_blur_input,logit_def4_roth2019odds}. Proposed defenses often claim a certain level of robustness against adversarial examples that have an amount of perturbation less than a selected norm~\cite{Croce2020Provable}. Since the topic of adversariality is closely linked with security, reproducibility of newly proposed techniques is of utmost importance. As a result, there have been a number of impactful studies that analyze the correctness and reliability of newly proposed adversarial defenses~\cite{Carlini_CVPR_Def,athalye2018_obfuscated,DBLP:journalsCarliniW17,CarliniTeam_2020_defense_evaluation}. In this context, Carlini and Wagner~\cite{DBLP:journalsCarliniW17}, for instance, demonstrated that most of the defenses proposed for MNIST~\cite{lecun1998gradient} do not even generalize to CIFAR~\cite{CIFAR}. This observation prompted research on the suitability of datasets for adversarial research~\cite{not_all_datasets_equal}, with Carlini and Wagner further suggesting that the usage of larger datasets such as ImageNet~\cite{ILSVRC15:rus} may be necessary, given the lack of generalization of defenses proposed for smaller datasets~\cite{DBLP:journalsCarliniW17}. 

Even though the results obtained with ImageNet are more convincing, working with ImageNet is much more challenging than, for example, working with MNIST or CIFAR. Indeed, not only does ImageNet contain more images than the other two, the images themselves are also larger. In addition, DNNs that achieve state-of-the-art results for ImageNet are also much bigger than their counterparts that achieve state-of-the-art results for MNIST or CIFAR, thus posing a challenge in terms of computational power needed. As a result, most of the studies that work with ImageNet only use a subset of images in order to create adversarial examples, unless that research is performed by a large industry lab that can afford the computational power (see Figure~\ref{fig:imagenet_data}). 

Although the studies of~\cite{su2018robustness_18_imagenet_models_evaluation,CarliniTeam_2020_defense_evaluation} hinted that not all source images may be equally suitable for adversarial example creation, most of the studies that work with adversarial examples often randomly sample source images among the ones that are correctly classified. As such, any image that is correctly classified by the models of interest is thought to be suitable and equal in terms of model-to-model transferability and the required perturbation to achieve adversariality. To the best of our knowledge, an in-depth analysis of source image suitability of adversarial examples in large-scale model-to-model scenarios has not been conducted yet. Hence, approaching the problem of adversarial examples from a different angle and following the directions of \cite{motivating_rules_for_adv_research,su2018robustness_18_imagenet_models_evaluation,CarliniTeam_2020_defense_evaluation}, instead of analyzing the effectiveness of attacks, the durability of defenses, or the robustness of models, our study focuses on the source images used to create adversarial examples, hereby investigating the impact of image selection on (1) the success of model-to-model adversarial transferability and (2) the required perturbation to achieve this transferability.

With the help of large-scale experiments, we demonstrate that, even when the most-studied adversarial attacks for benchmarking are used, model-to-model transferability successes of adversarial examples, as well as the amount of required perturbation to achieve this transferability, heavily depend on the source images used to create those adversarial examples. Moreover, we present a case study that shows how the experimental results obtained may lead to misleading conclusions when making use of certain subsets of source images.


\vspace{-1em}
\section{Adversarial attacks}
\vspace{-0.5em}
Assuming an $M$-class setting in which a data point and its categorical association are defined as $\bm{x} \in \mathbb{R}^k$ and $\bm{y} \in \mathbb{R}^M$, respectively, with $y_c = 1$ and $y_m = 0 \,, \forall \, m \in \{0,\ldots, M\} \char`\\ \{c\}$, let $g$ be a classification function that maps inputs onto categorical predictions. In this setting, we define the output $g(\theta, \bm{x}) \in \mathbb{R}^M$ as the logits obtained by a prediction model/classifier using the parameters $\theta$. The given data point is then classified into the category with the largest logit value:  $G(\theta, \bm{x}) = \arg \max_t(g(\theta, \bm{x})_t)$. If $G(\theta, \bm{x}) = \arg \max_t (\bm{y}_t)$, then this classification is correct.

For the given setting, a perturbation $\Delta_{x}$ bounded by the $L_p$ ball centered at $\bm{x}$ with a radius $\epsilon$, formulated as $\mathcal{B}(\bm{x})_{\epsilon}^{p} := \{ \hat{\bm{x}}: ||\Delta_{x}||_p:=|| \bm{x} - \hat{\bm{x}} ||_p \leq \epsilon \}$, is said to be an \textit{adversarial perturbation} if $G(\theta, \bm{x}) \neq G(\theta, \hat{\bm{x}})$. In this case, $\hat{\bm{x}}$ is also said to be an adversarial example. 

Since the discovery of adversarial examples, a plethora of attacks using a wide range of perturbation generation methods has been proposed~\cite{7789548,moosavi2016deepfool,xu2018structured_new_local_adv_attack}. Early research efforts in the field mostly made use of L-BFGS optimization~\cite{LBFGS}, Fast Gradient Sign Method (FGSM)~\cite{Goodfellow-expharnessing}, and Iterative Fast Gradient Sign Method (IFGSM)~\cite{IFGS}. However, Projected Gradient Descent (PGD)~\cite{PGD_attack} , the Carlini \& Wagner's Attack (CW)~\cite{CW_Attack}, and Momentum Iterative Fast Gradient Sign Method (MI-FGSM)~\cite{dong2018boosting} have taken the place of the aforementioned attacks in recent research efforts, thanks to the superior results obtained by the latter three. Following these findings, the study presented in this paper also uses these three attacks for examining the fragility of source images.

PGD can be seen as a generalization of FGSM and IFGSM. In particular, this attack aims at finding an adversarial example $\hat{\bm{x}}$ that satisfies $||\hat{\bm{x}} - \bm{x}||_{\infty} < \epsilon$, where the perturbation is defined within an $L_{\infty}$ ball centered at $\bm{x}$ with a radius $\epsilon$. The adversarial example is iteratively generated as follows: $[\hat{\bm{x}}]_{n+1} =  \Pi_{\epsilon}\Big([\hat{\bm{x}}]_{n} - \alpha \ \text{sign} \big(\nabla_x J(g(\theta,[\hat{\bm{x}}]_n)_c)  \big)\Big) \,,$ with $[\hat{\bm{x}}]_1 = \bm{x}$, where the perturbation is calculated using the signature of the gradient of the cross-entropy loss, $\text{sign} (\nabla_x J(g(\theta, [\hat{\bm{x}}])_c))$, originating from the target class $c$. In this setting, $\alpha$ controls the exercised perturbation at each iteration and $\Pi_{\epsilon}$ is a function that controls the $L_{\infty}$ limit imposed on the perturbation.

CW, on the other hand, is a complex attack that aims to find a perturbation within a small $L_2$ norm as follows: $\min_{\hat{\bm{x}}} \, f(\hat{\bm{x}}, c) + ||\hat{\bm{x}} - \bm{x} ||_2 \,$, where $f$ is a preferred loss function. Given \cite{CW_Attack,su2018robustness_18_imagenet_models_evaluation}, we use the following loss: $f(\hat{\bm{x}}, c) = \max_k \{ \max_{c\neq k}\{g(\theta, \hat{\bm{x}})_c - g(\theta, \hat{\bm{x}})_k \} - \kappa\} \,,$ selecting the target class with $c$ and adjusting the confidence of the attack with $\kappa$.

The overall structure of MI-FGSM is similar to that of PGD and IFGSM. However, instead of adding perturbation directly to the image, it integrates the gradient of the cross-entropy loss into a variable that acts as a momentum term: \mbox{$[\bm{\tau}]_{n+1} = \mu [\bm{\tau}]_n + \frac{J(g(\theta,[\hat{\bm{x}}]_n)_c)}{ ||J(g(\theta,[\hat{\bm{x}}]_n)_c)||_1}$}, where $\mu$ is the multiplier for already-accumulated gradient in past iterations. Unlike the previous two attacks, the perturbation is generated from this momentum term $\bm{\tau}$ instead of the gradient itself, and iteratively added to the image as follows: $[\hat{\bm{x}}]_{n+1} =  \Pi_{\epsilon}\Big([\hat{\bm{x}}]_{n} - \alpha \ \text{sign} (\bm{\tau}) \Big) \,,$ with $[\hat{\bm{x}}]_1 = \bm{x}$.

Given that adversarial examples are trivial to generate in white-box cases~\cite{Carlini_CVPR_Def,DBLP:journalsCarliniW17}, and given the recent focus on the importance of adversarial evaluation in black-box scenarios~\cite{ilyas2018black_black_black_box,tu2019autozoom_black_black_box}, our study mainly focuses on analyzing the properties of adversarial examples that achieve model-to-model transferability. In this context, an adversarial example created by a model is said to achieve model-to-model adversarial transferability if it is also incorrectly classified by another model, provided that the source image used to create the adversarial example is initially correctly classified by both models.

\vspace{-1em}
\section{Experimental setup}
\label{sec:ExperimentalSetup}

\hspace{0.425cm}\textbf{Models}\,\textendash\,In this study, we use five different deep learning architectures that see frequent use in the literature. The considered models are: AlexNet~\cite{Alexnet}, SqueezeNet~\cite{squeezenet}, VGG-16~\cite{VGG}, ResNet-50~\cite{resnet}, and DenseNet-121~\cite{densenet}. In addition to these models, we also include two recently proposed vision transformer models that achieve state-of-the-art results on ImageNet~\cite{dosovitskiy2021an}: Vision Transformer Base$/16-224$ (ViT-B) and Vision Transformer Large$/16-224$ (ViT-L). From here on, each model will be denoted by its set of parameters $\theta_{i}, i \in \{1,\dots,7\}$, and multiple models will be denoted by $\Theta = \{\theta_1, \ldots, \theta_7\}$.

\textbf{Data}\,\textendash\,We follow the approach used by previous studies on adversariality, leveraging the images in the ImageNet validation set for generating adversarial examples. In this paper, these unperturbed images are referred to as \textit{source images}. Further adopting previously used methods, we only rely on images that are correctly classified by all selected models in order to conduct trustworthy experiments on adversarial transferability, thus ensuring $G(\theta_i, \bm{x}) = \arg \max_t (\bm{y}_t), \forall i \in \{1,\ldots, 7\}$. By doing so, we filter out images that are hard to correctly classify for at least one of our models, thus limiting the hypothesis space and allowing us to perform a best-case analysis. After this filtering operation, we are left with a set of $19,025$ source images, which approximately corresponds to $38\%$ of the ImageNet validation set. We will refer to this set of $19,025$ source images as:
\begin{align}
\mathbb{S} = \{\bm{x} \mid G(\theta_i, \bm{x}) = \arg \max_t(\bm{y}_t); i \in \{1,\ldots,7\}\} \,.
\end{align}

\textbf{Adversarial perturbation}\,\textendash\,Although the methods used to identify perturbation in images are not a perfect match for how humans perceive noise, $L_p$ norms (with $p \in \{0, 2, \infty\}$) are commonly used since the early days of research on adversarial examples~\cite{CW_Attack,motivating_rules_for_adv_research,DBLP:journals/corr/PapernotMWJS15}. We adopt both $L_{2}$ and $L_{\infty}$ norms for measuring the added perturbation. In terms of the used $L_{\infty}$ perturbation budget, another large-scale study on adversarial transferability~\cite{su2018robustness_18_imagenet_models_evaluation} uses $\epsilon_{[0, 1]} \in \{0.1, 0.2, 0.3\}$, which approximately corresponds to $\epsilon_{[0, 255]} \in \{25, 45, 67\}$ in discretized settings. We observed that using $\epsilon_{[0, 255]} \geq 45$ leads to adversarial examples that come with large perturbation budgets. 
In light of this observation, we limit the perturbation on an $L_{\infty}$ ball to $38$ (i.e., $\epsilon_{[0, 255]} = 38$, $\epsilon_{[0, 1]} = 0.15$), thus ensuring that the perturbation is not excessive. Further details on the calculation of $L_p$ norms and the employed attacks, as well as a comparison of perturbation visibility, can be found in the supplementary material (Section~A).

For PGD and MI-FGSM, we perform the attack with $50$ iterations and allow a perturbation budget of $\epsilon_{[0, 1]} = 0.15$. For MI-FGSM, we follow the work of~\cite{dong2018boosting} and use $\mu = 1$. For CW, we use $\kappa=20$ (as suggested by the authors of the attack). Using a randomly selected class that differs from the true class of the source image, we perform the aforementioned attacks on source images. In order to avoid cases where the \mbox{image/target} class combination is challenging, if an attack does not succeed within the allocated number of iterations, we select another class, and we perform the attack on the same image up to five times. At each iteration of the adversarial attack, we analyze whether or not the prediction for the image changed for the other six models (i.e., evaluating the non-targeted transferability) and then save the adversarial examples with the smallest perturbation. By doing so, we aim at finding the least-required perturbation, as exercised by all three attacks, that is sufficient to convert a source image into an adversarial one. 

\textbf{Non-adversarial perturbation}\,\textendash\,In addition to the adversarial attacks, we also make use of commonly used image distortion techniques in order to measure the robustness of source images. For this analysis, we employ (1) uniform noise, (2) Gaussian noise, and (3) change in contrast to create \say{adversarial examples}, where all of these additive types of noise respect the $L_{\infty}$ limit put in place for the adversarial attacks. Details on the usage of these operations can be found in the supplementary material (Section~B).

\vspace{-1em}
\section{Methodology for the source image analysis}
\vspace{-0.5em}

In this section, we explain the notation and methodology used for the analysis of source images in adversarial scenarios. We denote by $\hat{\bm{x}}^{(\text{A}):i \to j}$ an adversarial example that is created through the addition of adversarial perturbation with the attack $(\text{A}) \in \{\text{PGD}, \text{CW}, \text{MI-FGSM}\}$, calculated from the model $\theta_i$, but that is misclassified by model $\theta_j$, thus achieving adversarial transferability. We then denote the set of all adversarial examples that achieve adversarial transferability, created through the usage of source image $\bm{x}$, as follows: 
\begin{align}
\widehat{\mathcal{X}}^{(\text{A})} := \{ \hat{\bm{x}}^{(\text{A}):i \to j} \, | \, i,j = 1,\ldots,7 \, ; \, i \neq j\} \,.
\end{align}

We measure the added perturbation with $L_{\{2,\infty\}}$ norms. Moreover, we denote the least amount of perturbation required to convert a source image into an adversarial example for a particular target model ($j$), regardless of which other model it is generated from, by:
\begin{align}
d_p(\theta_j, \widehat{\mathcal{X}}^{(\text{A})}) = &\min_{i \in \{ 1,\ldots,7\} \setminus \{j\}}  || \bm{x} - \hat{\bm{x}}^{(\text{A}):i \to j}||_p  \,,
\end{align}
where $p$ denotes the selected norm. We also measure the minimum amount of perturbation required to convert a source image into an adversarial example for any model as follows:
\begin{align}
D_p(\Theta, \widehat{\mathcal{X}}^{(\text{A})}) = &\min_{j \in \{ 1,\ldots,7\}} d_p(\theta_j, \widehat{\mathcal{X}}^{(\text{A})}) \,.
\end{align}
Another important benchmark is the transferability count of adversarial examples created from individual source images. Since we have seven models, and since we are only interested in model-to-model transferability, we count the successful model-to-model transfers for adversarial examples generated from a source image $\bm{x}$ and the attack $A$ as follows:
\begin{align}
T(\Theta, \widehat{\mathcal{X}}^{(\text{A})}, \bm{y}) = \sum_{i,j=1 ,\, i \neq j}^{7}  \mathds{1}_{\{G(\theta_{j}, \, \hat{\bm{x}}^{(A): i \to j}) \, \neq \, \arg \max_t (\bm{y}_t)\}} \,.
\end{align}
For each source image and attack, this (untargeted) transferability count $T(\Theta, \widehat{\mathcal{X}}^{(\text{A})}, \bm{y})$ can take a value between $0$ and $42$. In this context, having zero model-to-model transferability means that none of the adversarial examples generated from a particular source image achieved adversarial transferability and $42$ means that the adversarial examples created from a source image achieved adversarial transferability in all model-to-model scenarios.

\begin{figure}[t!]
\centering
\begin{subfigure}{0.32\textwidth}
\includegraphics[width=\linewidth]{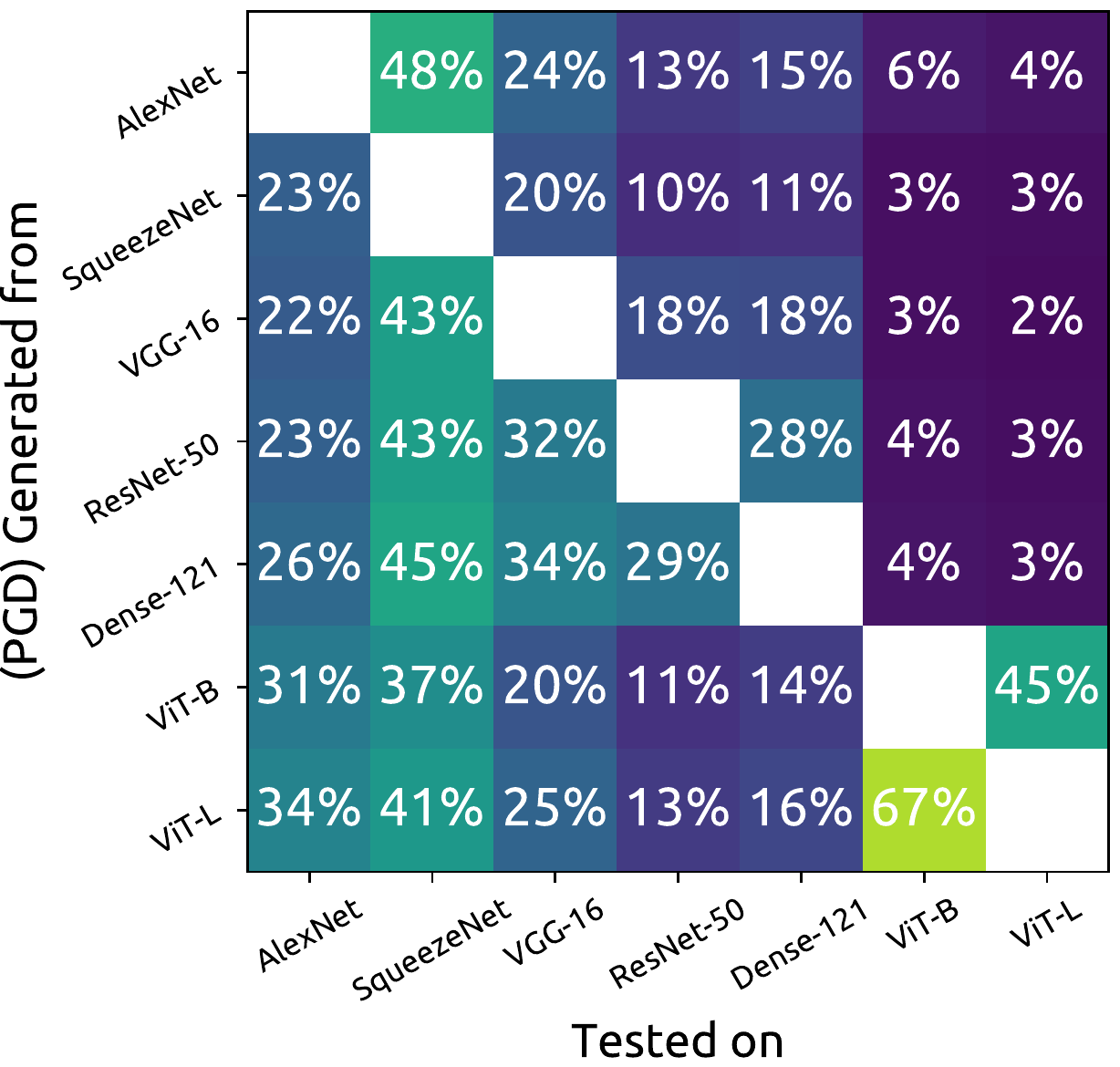}
\caption{All images $(\mathbb{S})$}
\label{fig:transferability_matrix}
\end{subfigure}
\begin{subfigure}{0.32\textwidth}
\includegraphics[width=\linewidth]{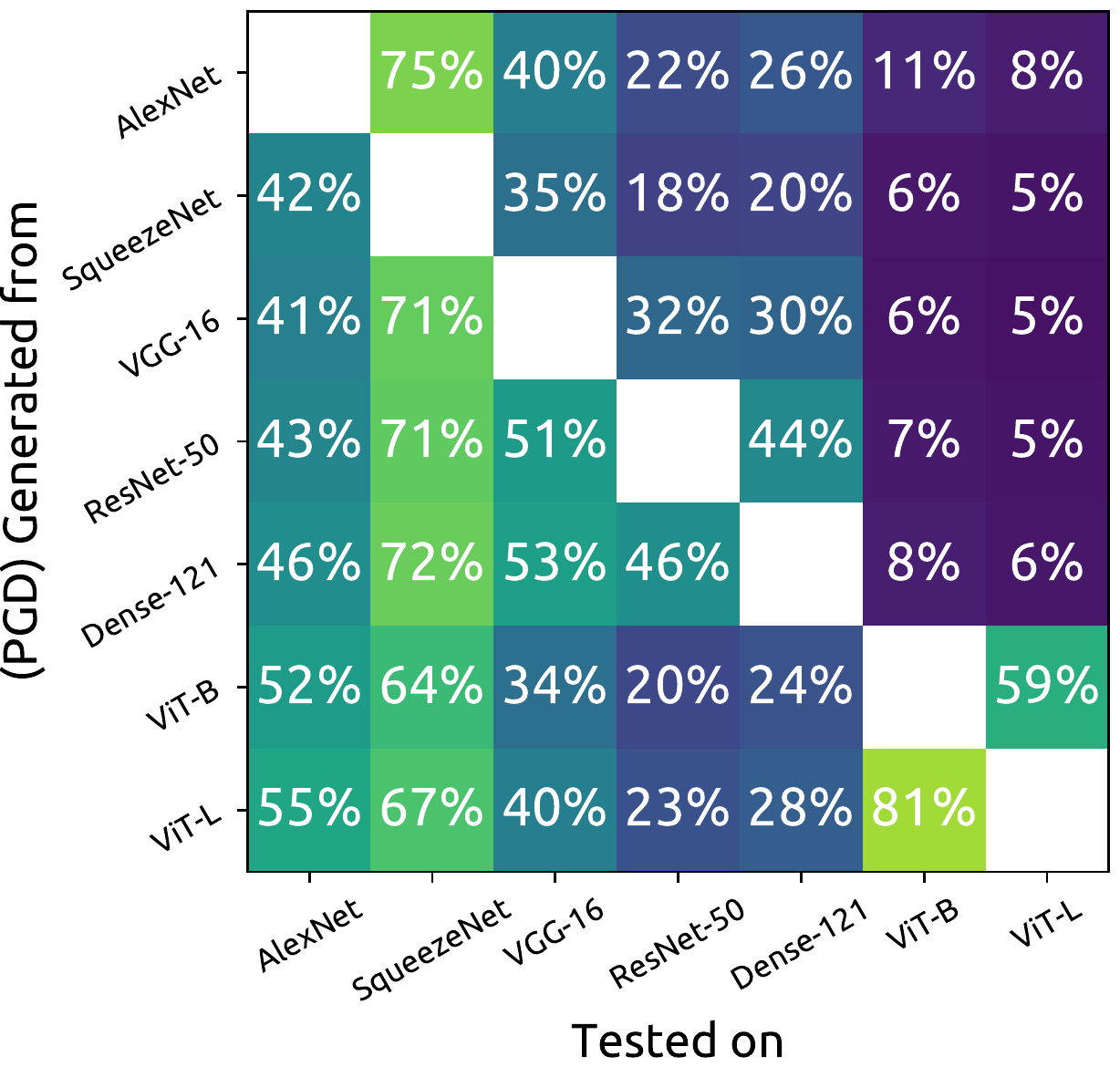}
\caption{Fragile images $(\mathbb{S}_f)$}
\label{fig:trans-nonadv}
\end{subfigure}
\begin{subfigure}{0.32\textwidth}
\includegraphics[width=\linewidth]{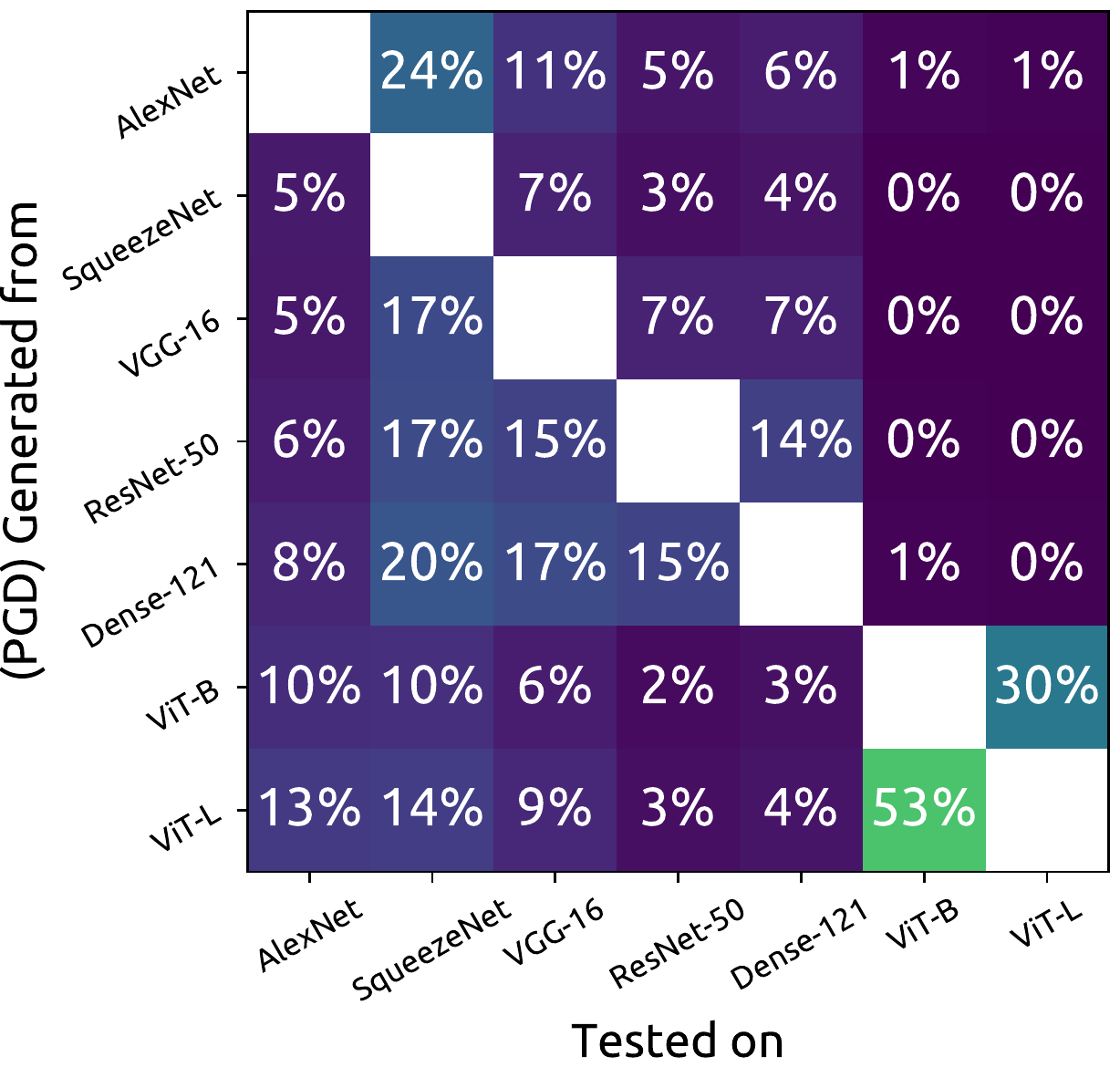}
\caption{Hard images $(\mathbb{S}_h)$}
\label{fig:trans-onlyadv}
\end{subfigure}
\vspace{1em}
\caption{Proportion of source images in (left) $\mathbb{S}$, (center) $\mathbb{S}_f$, and (right) $\mathbb{S}_h$ that achieved (untargeted) adversarial transferability with the usage of PGD.}
\label{fig:trans-matrix-overall}
\vspace{-1em}
\end{figure}

\vspace{-1em}
\section{Experimental results}
\label{sec:Experimental Results}
\vspace{-0.5em}

Through the methodology described above, we successfully created $825,005$ adversarial examples that achieve adversarial transferability for at least one model-to-model scenario (excluding white-box cases). Specifically, $173,542$, $115,688$, and $535,775$ adversarial examples were produced with PGD, CW, and MI-FGSM, respectively. In the remainder of this paper, we provide and discuss experimental results for these $825,005$ adversarial examples, as well as for the $19,025$ source images used to obtain them. 

Since MI-FGSM is able to create a large number of adversarial examples that achieve model-to-model transferability compared to the other two attacks, experimental results obtained through the usage of all adversarial examples may be skewed towards adversarial examples created with MI-FGSM. For the sake of precise experimentation, when we inspect all adversarial examples for an experiment, we provide the same experiment in the supplementary material using adversarial examples created with individual attacks.

\vspace{-0.5em}
\subsection{Model-to-model transferability}
\label{Model-to-model transferability}
\vspace{-0.5em}

In Figure~\ref{fig:transferability_matrix}, we show the model-to-model transferability success ratio of adversarial examples generated with PGD. Specifically, we provide details for the source and target models of all $173,542$ adversarial examples that achieved (untargeted) adversarial transferability. 

\begin{figure}[t!]
\centering
\begin{subfigure}{0.43\textwidth}
\includegraphics[width=\textwidth]{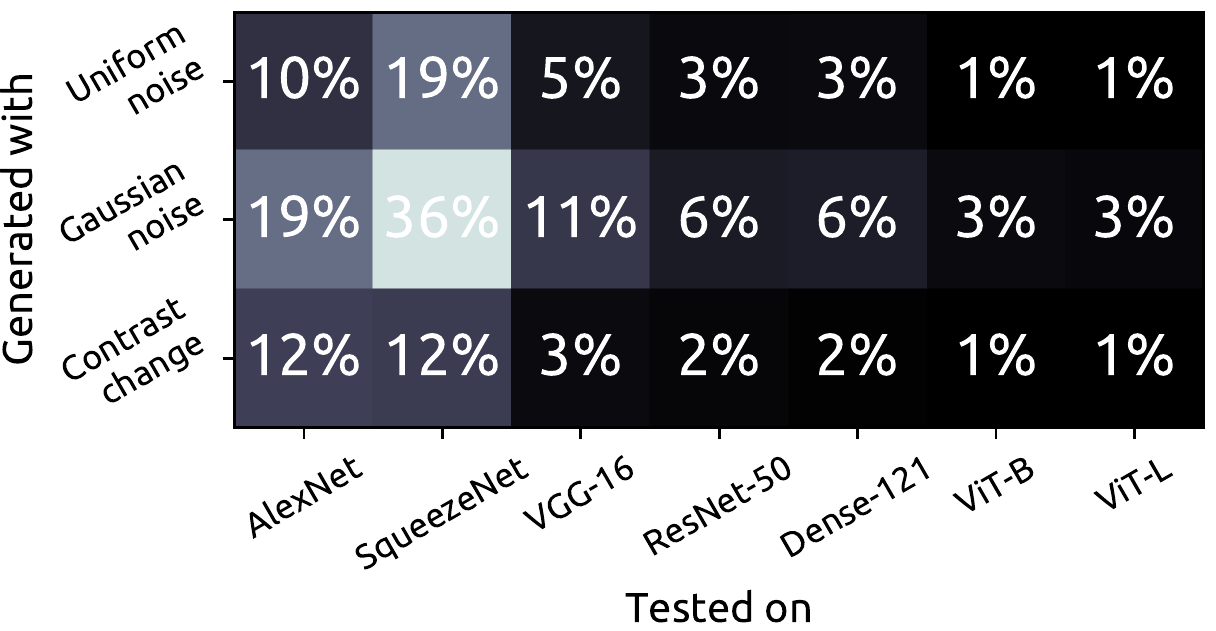}
\caption{Non-adversarial noise and transferability}
\label{fig:non_adv_transferability_matrix}
\end{subfigure}
\begin{subfigure}{0.47\textwidth}
\includegraphics[width=\textwidth]{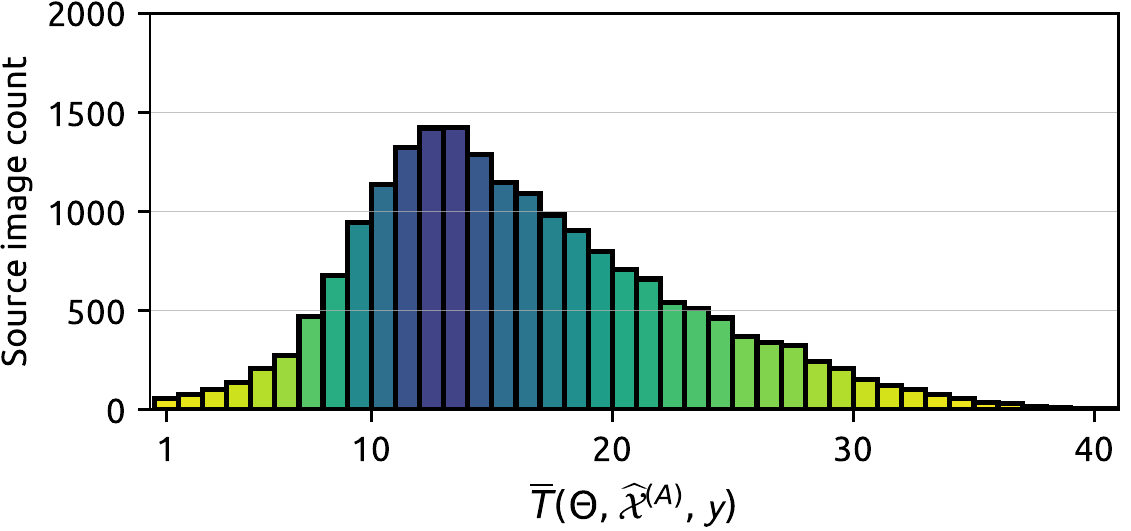}
\caption{Source images that attained transferability}
\label{fig:adv_trans_cnt}
\end{subfigure}
\vspace{1em}
\caption{(left) Number (proportion) of source images that became ``adversarial examples'' through the addition of non-adversarial noise and (right) histogram  of transferability count of source images of source images and their transferability count according to $\overline{T}(\Theta, \widehat{\mathcal{X}}^{(\text{A})}, \bm{y})$.}
\end{figure}

In order to answer the question of whether or not adversarial transferability success can be influenced by source image selection, let us continue with an unusual experiment. In Figure~\ref{fig:non_adv_transferability_matrix}, we show the number of source images that had their predictions changed for the models listed on the $x$-axis through the application of the non-adversarial perturbations listed on the $y$-axis. Surprisingly, relying on common noise generation methods that do not require any special setup, we observe that a large portion of source images have their classification changed in a limited $L_{\infty}$ ball setting. Specifically, $9,615$ unique source images, corresponding to approximately $50\%$ of the source images ($\mathbb{S}$), become \say{adversarial examples} for at least one model through the introduction of non-adversarial noise.

Combining the two experiments (Figure~\ref{fig:transferability_matrix} and Figure~\ref{fig:non_adv_transferability_matrix}) that have been discussed thus far, let us divide $\mathbb{S}$ into two sets $\mathbb{S}_f$ and $\mathbb{S}_h$, where the former contains \emph{fragile source images} that had, at least once and for any model, their prediction changed with the application of non-adversarial noise ($9,615$ source images) and where the latter contains the remaining images ($9,410$ source images), with $\mathbb{S} = \mathbb{S}_f \cup \mathbb{S}_h$. According to this separation, we provide Figure~\ref{fig:trans-nonadv} and Figure~\ref{fig:trans-onlyadv}, where we show the model-to-model transferability of the adversarial examples originating from the source images in $\mathbb{S}_f$ and $\mathbb{S}_h$, respectively. As can be seen, even though we use a similar number of source images taken from the same dataset for both $\mathbb{S}_f$ and $\mathbb{S}_h$, we obtain outcomes that are completely different in terms of adversarial transferability success. We present detailed versions of all transferability matrices, as well as the results obtained for CW and MI-FGSM, in the supplementary material (Section~C).

The reason for the large discrepancy between the results presented in Figure~\ref{fig:trans-nonadv} and Figure~\ref{fig:trans-onlyadv} is the fragility of a subset of the source images. Compared to the other, non-fragile images, these fragile source images have their predictions easily changed for a large number of models, even when other conditions are held the same (e.g., attacks and models). In order to lay bare the fragility of these source images, we perform an aggregate analysis of their average transferability per attack, leading to a histogram of $\overline{T}(\Theta, \widehat{\mathcal{X}}^{(\text{A})}, \bm{y})$ for all source images in $\mathbb{S}$, as shown in Figure~\ref{fig:adv_trans_cnt}. This histogram illustrates that, with one of the employed attacks, a large portion of the adversarial examples achieve adversarial transferability between $10$ to $20$ times. However, an intriguing observation can be made for the leftmost and the rightmost side of this figure, where $585$ source images achieve adversarial transferability less than $5$ times and where $1,743$ sources images achieve transferability more than $25$ times. These images that, through the added perturbation, do not easily become adversarial examples, as well as the fragile source images, which easily change predictions between models and which achieve unnaturally high model-to-model transferability, will be our main focus for the remainder of this paper.

\begin{figure}[t!]
\centering
\includegraphics[width=0.5\textwidth]{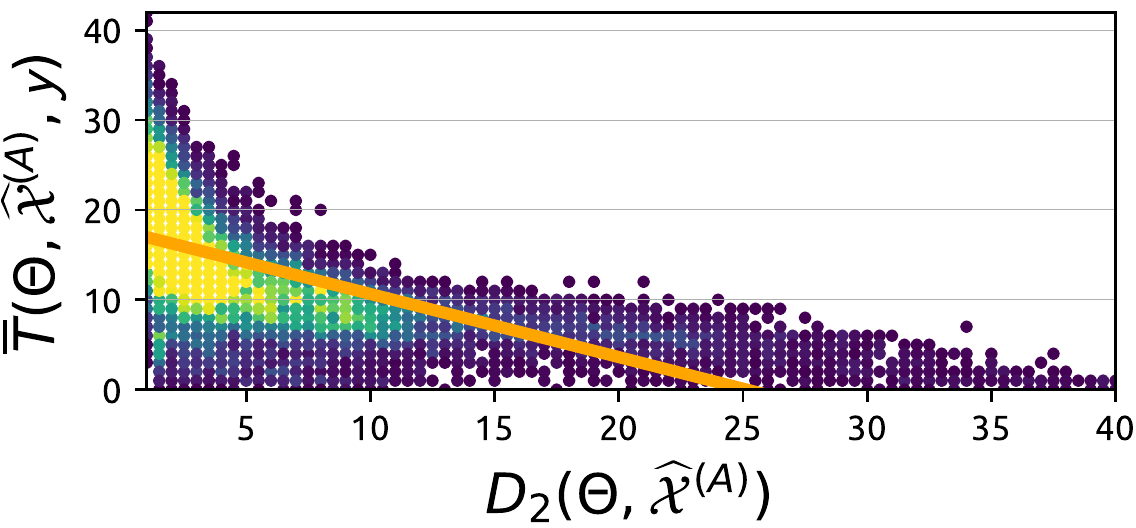}
\includegraphics[width=0.35\textwidth]{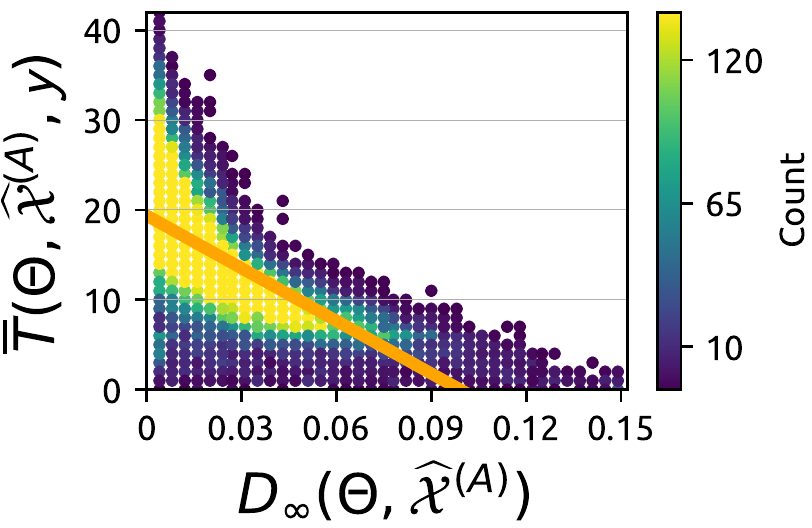}
\vspace{0.75em}
\caption*{\qquad\,\,\,\,\, Correlation: $-0.67$ \qquad\qquad\,\,\,\,\,\,\,\,\,\,\,\,\,\,\, Correlation: $-0.61$}
\vspace{1em}
\caption{Scatter plot of $D_p(\Theta, \widehat{\mathcal{X}}^{(A)})$, the minimum amount of perturbation required for each source image, against average adversarial transferability count $\overline{T}(\Theta, \widehat{\mathcal{X}}^{\text{(A)}}, \bm{y})$, for $p=2$ (left) and $p=\infty$ (right). The regression line is shown in orange.}
\label{fig:transferability_perturbation_main}
\vspace{-1em}
\end{figure}

\vspace{-0.5em}
\subsection{Adversarial perturbation}

Another important aspect of model-to-model adversarial transferability is how easy a source image becomes an adversarial example, since the robustness of adversarial defenses, as well as recently proposed models, are certified under an $L_p$ norm perturbation. In Figure~\ref{fig:transferability_perturbation_main}, we perform a correlation analysis between $\overline{T}(\Theta, \widehat{\mathcal{X}}^{(A)}, \bm{y})$ and the minimum required $L_p$ perturbation to achieve adversarial transferability $D_{\{2,\infty\}}(\Theta, \widehat{\mathcal{X}}^{(\text{A})})$. In this context, we observe a mild negative correlation between added noise and transferability count, where the adversarial examples originating from source images that achieve higher transferability counts are also the ones that require less perturbation. These results hint that the fragile images we have identified do not only achieve high adversarial transferability, but that they also do so with smaller perturbation budgets.

\begin{figure*}[t!]
\centering
\rotatebox[origin=l]{90}{\phantom{--}\scriptsize\underline{$T(\cdot)\geq1$}}\hspace{0.5em}
\includegraphics[width=0.4\linewidth]{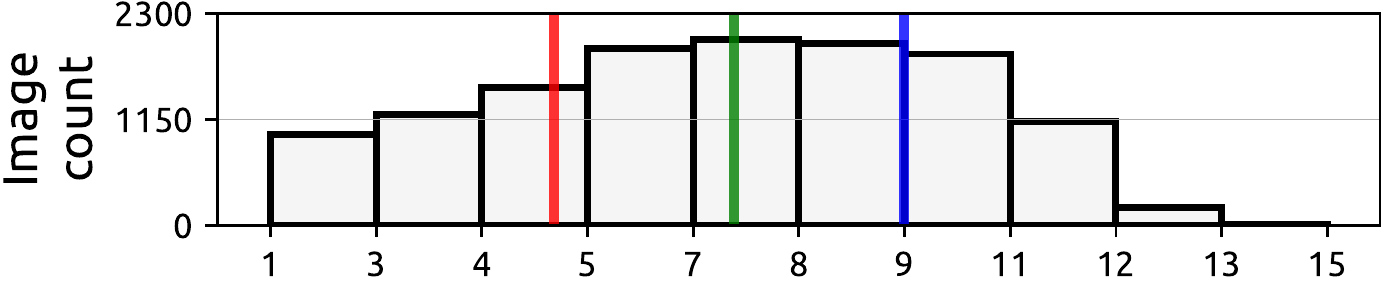}\hspace{1em}
\includegraphics[width=0.4\linewidth]{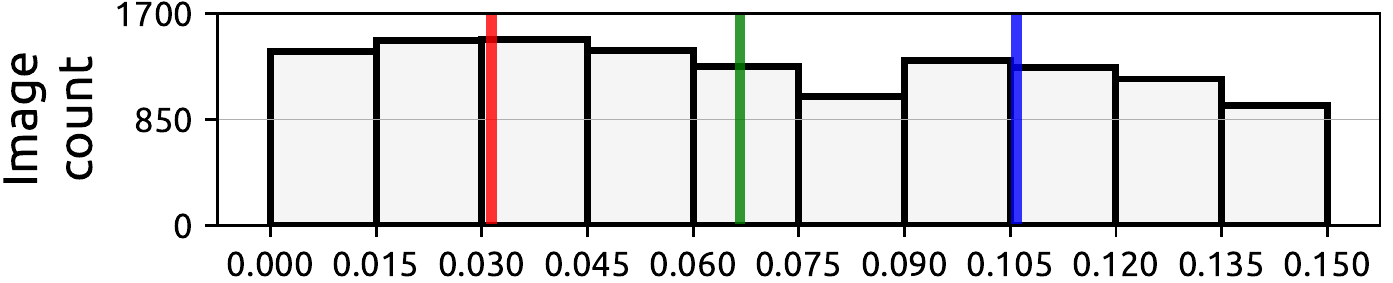}
\\
\vspace{0.2em}
\rotatebox[origin=l]{90}{\phantom{--}\scriptsize\underline{$T(\cdot)\geq20$}}\hspace{0.5em}
\includegraphics[width=0.4\linewidth]{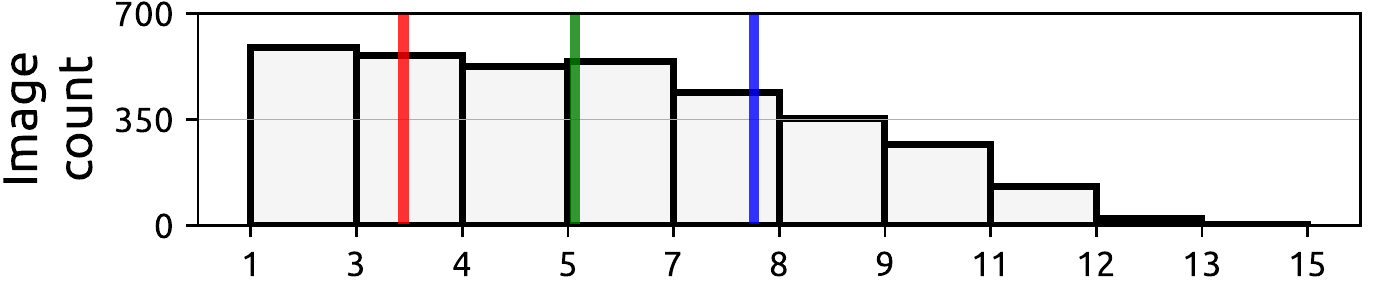}\hspace{1em}
\includegraphics[width=0.4\linewidth]{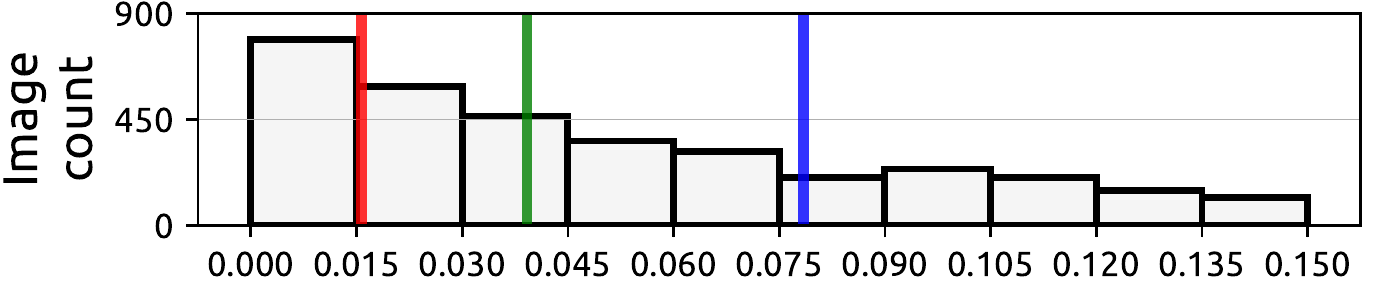}
\\
\vspace{0.2em}
\hspace{0.1em}
\rotatebox[origin=l]{90}{\phantom{--.}\scriptsize\underline{$T(\cdot)\geq30$}}\hspace{0.5em}
\includegraphics[width=0.4\linewidth]{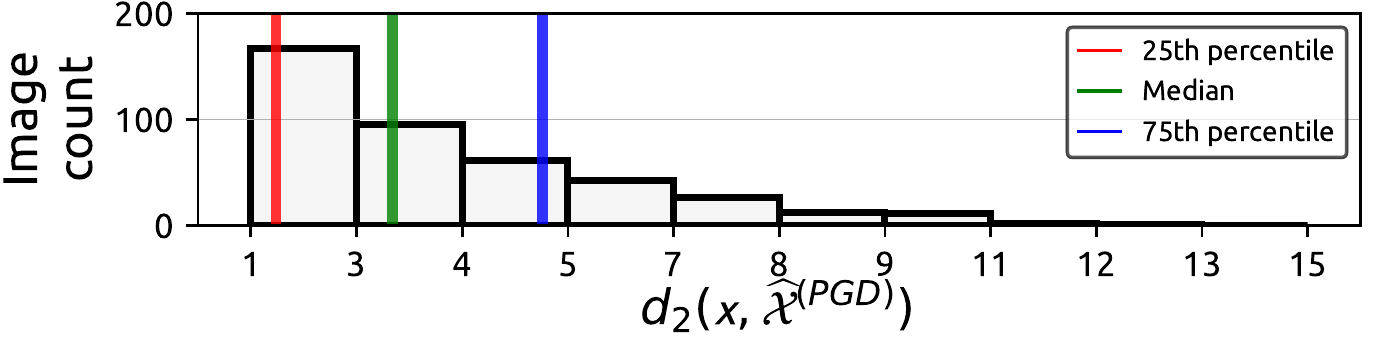}\hspace{1em}
\includegraphics[width=0.4\linewidth]{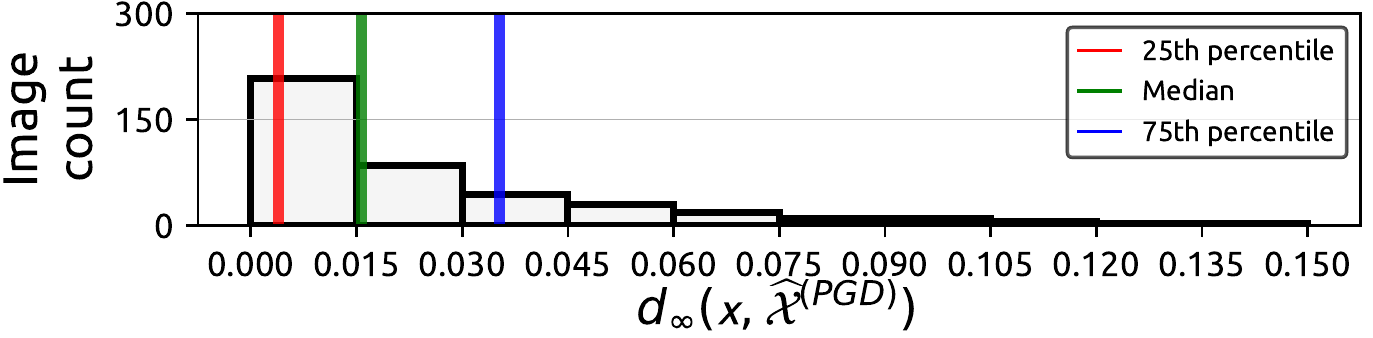}
\vspace{1em}
\caption{Source images that achieved adversarial transferability to ViT-B are selected based on transferability count, with $T(\Theta, \widehat{\mathcal{X}}^{\text{(PGD)}}, \bm{y}) \geq \{1, 20, 30\}$. The minimum amount of perturbation required for creating adversarial examples from these source images is histogrammed, measuring the perturbation using $d_p(\bm{x}, \widehat{\mathcal{X}}^{\text{(PGD)}})$, with $p\in \{2,\infty\}$. The median perturbation, as well as the $25$th and the $75$th percentile, are provided in order to improve interpretability.}
\label{fig:pert_norm}
\end{figure*}

In order to solidify these observations regarding perturbation and transferability, we part from an aggregate analysis to a more granular one and investigate the perturbations of adversarial examples that achieve transferability for each model individually. In Figure~\ref{fig:pert_norm}, we provide for ViT-B the smallest required $L_{\{2,\infty\}}$ perturbation for source images progressively filtered with $T(\Theta, \widehat{\mathcal{X}}, \bm{y}) \geq \{1, 20, 30\}$. Note that, as $T(\Theta, \widehat{\mathcal{X}}, \bm{y})$ increases, the distribution of the perturbation shifts towards zero, thus confirming our previous observations. These results indicate that source images that achieve high transferability counts are, most likely, also the ones that require less perturbation. Similar results, as available in the supplementary material (Section~D and Section~E), can be observed for the other models.

\vspace{-1.5em}
\section{Source image suitability}

Our experiments indicate that, while a certain portion of images never becomes adversarial, another portion of images can be easily turned into adversarial examples using relatively small perturbation budgets. Given the importance of research reproducibility, this leads to the question of how much variance can be observed when randomly sampling source images. In order to answer this question, we randomly sample $1,000$ source images from $\mathbb{S}$ (since this number seems to be the most commonly selected number in Figure~\ref{fig:imagenet_data}), subsequently inspecting the transferability success and $L_{\{2,\infty\}}$ perturbation norms of the adversarial examples generated for the individual model-to-model transferability cases. We perform the aforementioned routine $10,000$ times. As a result, we calculate the lowest, the highest, and the average transferability, as well as the $L_{\{2, \infty\}}$ perturbations. Overall, we observe that, while the average case closely matches the usage of all available source images, it is possible to have differences between the lowest and the highest case of up to $12.5\%$ in transferability, $1.01$ in $L_2$ norm perturbation, and $0.03$ (i.e., $8/255$) in $L_{\infty}$ norm perturbation. These results indicate that, even when random sampling is used, it is indeed possible to have conflicting results, depending on the source images selected.

In Section~\ref{Model-to-model transferability}, we demonstrated that one way to identify source images that are fragile to adversarial attacks is to perform a large-scale analysis of model-to-model transferability using all possible source images. However, such an approach is not scalable, unless an abundance of computational power is available, thus forcing us to investigate alternate methods for the identification of these atypical source images. An important piece of information we have for each source image is the vector of prediction probabilities obtained through the softmax function, $P(\theta, \bm{x}) = [e^{g(\theta, \bm{x})_c} / \sum_{k=1}^{M} e^{g(\theta, \bm{x})_k}]_{c \in \{1,\ldots,M\}}$. The softmax output in conjunction with various error quantification methods has seen a significant use in recent research efforts on measuring the robustness and calibrated nature of DNNs~\cite{guo2017calibration}.
Relying on the knowledge obtained from these studies, we use the following error quantification methods for evaluating the suitability of source images: the error made for the correct class, as calculated by (1) $1- \max (P(\theta, \bm{x}))$, (2) mean squared error (MSE), (3) Wasserstein distance (WD), and (4) the ratio of probabilities (Q) (that is, the second-largest to the largest one). Details on the way the different errors are calculated can be found in the supplementary material (Section~F).

\begin{table}[t]
\centering
\caption{Correlation coefficients between various estimates of errors in source image predictions and properties of adversarial examples (transferability and perturbation) created from those source images are given for PGD, CW, and MI-FGSM.}
\scriptsize
\vspace{1em}
\begin{tabular}{lccccccccc}
	\cmidrule[1pt]{1-10}
    \multirow{3}{*}{\shortstack{Error\\measurement}} &  \multicolumn{3}{c|}{PGD} & \multicolumn{3}{c|}{CW}  &  \multicolumn{3}{c}{MI-FGSM} \\ 
    \cmidrule[0.25pt]{2-10}
    ~ &  $T(\cdot)$ & $d_2(\cdot)$ & $d_{\infty}(\cdot)$ &  $T(\cdot)$ & $d_2(\cdot)$ & $d_{\infty}(\cdot)$ &  $T(\cdot)$ & $d_2(\cdot)$ & $d_{\infty}(\cdot)$ \\
    \cmidrule[0.5pt]{1-10} 
     $\text{Q}(P(\theta, \bm{x}))$		        & $0.58$ 	& $\bf -0.64$ 	& $\bf -0.58$ 	& $\bf 0.57$ & $\bf -0.59$ & $\bf -0.66$ & $0.42$ & $-0.54$ & $-0.54$\\
     $1 - \max (P(\theta, \bm{x}))$	            & $\bf 0.61$ 	& $-0.60$ 	& $-0.57$ & $\textbf{0.57}$ & $-0.54$ & $-0.63$ & $\bf 0.43$ & $\bf -0.58$ & $\bf -0.57$\\
     $\text{MSE}(P(\theta, \bm{x}), \bm{y})$	& $0.56$ 	& $-0.57$ 	& $-0.53$ & $0.56$ & $-0.51$ & $-0.61$ & $0.37$ & $-0.51$ & $-0.53$\\
     $\text{WD}(P(\theta, \bm{x}), \bm{y})$		& $0.33$ 	& $-0.35$ 	& $-0.37$ & $0.33$ & $-0.32$ & $-0.37$ & $0.29$ & $-0.38$ & $-0.38$\\
\cmidrule[1pt]{1-10}
\end{tabular}
\label{tbl:dist_correlation_table}
\end{table}

In Table~\ref{tbl:dist_correlation_table}, we provide the correlation between (a) the error measurement for the prediction of source images and (b) the properties of adversarial examples originating from those images (i.e., transferability and perturbation). Even though we use a large number of data points for this analysis, we still find a moderate correlation between multiple error estimates and adversarial properties. In particular, the simple approach of $\text{Q}(\cdot)$ has the largest correlation when it comes to estimating perturbations, while having a comparably large correlation with transferability. Based on Table~\ref{tbl:dist_correlation_table}, we observe that, when $P(\theta, \bm{x})$ for a source image has its second-largest prediction closer to the largest one, adversarial examples originating from that source image are more likely to achieve adversarial transferability while requiring less perturbation. This leads to the question whether or not these error estimates can be used to identify fragile images, thus alleviating the need for large-scale experimentation. To answer this question, we devise the following experimental procedure.



In order to approximate the adversarial properties of source images, we group source images according to the $Q(P(\theta, \bm{x}))$-value obtained. Specifically, we sort $\mathbb{S}$ according to $Q(P(\theta, \bm{x}))$ and create subsets based on certain percentiles of $Q(\cdot)$. Doing so, we observe the results for the same experimental routine described above (i.e., $1,000$ source images sampled $10,000$ times), but with a small difference: only the source images that have $Q(\cdot)$ larger than the $75$th and $90$th percentile ($\mathbb{S}_{Q>{\{75, 90\}}}$), as well as source images that have $Q(\cdot)$ smaller than the $10$th and $25$th percentile ($\mathbb{S}_{Q<{\{10,25\}}}$), are selected.

We observe that source images with lower error estimates, as measured through $Q(P(\theta, \bm{x}))$, are harder to convert to adversarial examples, whereas the ones with higher $Q(P(\theta, \bm{x}))$ estimates are easier to convert. Furthermore, the required amount of perturbation for creating adversarial examples also differs greatly between the lower and the upper end of $Q(\cdot)$, with source images having a lower $Q(\cdot)$ requiring more perturbation, and vice versa. These results indicate that error estimates based on the prediction of source images can be used as a proxy for the properties of adversarial examples originating from these source images.

Finally, we measure the adversarial properties obtained with source images filtered from both ends, with \mbox{$\mathbb{S} \setminus (\mathbb{S}_{Q<P} \cup \mathbb{S}_{Q>100-P})$}. Using this approach, overall, we are able to reduce the difference between the highest and the lowest transferability from $12.5\%$ to $7.6\%$, the difference in $L_2$ norm perturbation from $1.01$ to $0.71$, and the difference in $L_{\infty}$ norm perturbation from $0.03$ to $0.01$, thus pointing to a more stable experimentation that is closer to the average case. Moreover, when we filter the same number of images from both ends (e.g., \mbox{$\mathbb{S} \setminus (\mathbb{S}_{Q<10} \cup \mathbb{S}_{Q>90})$}), the average transferability goes down slightly compared to using all available source images, while the average amount of required perturbation goes up slightly. These results indicate that the usage of $Q(\cdot)$ is more reliable in identifying fragile (easy) source images than hard source images. Consequently, we believe that the way error measurements are performed can be further improved, for instance through the usage of more complex analysis that takes into account the categories of source images.

Comprehensive experimental results for the experiments detailed in this section are provided in the supplementary material (See Table~I to Table~VI).

\vspace{-1em}
\section{Conclusions and outlook}
\label{Conclusions}
\vspace{-0.5em}

With the help of large-scale experiments, we exposed the fragility of a subset of source images to adversariality, with the adversarial examples created from these fragile images achieving high transferability rates for relatively small perturbation budgets. We then took one of the first steps to identify unusual source images that are either very hard or very easy to convert to adversarial examples, with the goal of supporting high-quality experimentation. 

Given the security concerns associated with adversarial examples, an important item for future work is to evaluate how the observations made in this paper extend to adversarial defenses. In particular, we believe that the fragile images we have identified, given the properties discussed in this paper, can easily be leveraged to circumvent adversarial defenses.

We noted that a large number of adversarial examples are misclassified into categories that are semantically close to the categories of their source image counterparts, thus achieving untargeted adversarial transferability. In the supplementary material (Section~G), we provide a number of qualitative examples of such cases. In that regard, we believe a detailed investigation of this topic, involving the semantic relationships between different categories, is also a promising item for future work, and where this future work item could make use of the hierarchies available in the WordNet database~\cite{WordNet}.

\bibliography{adv_data_paper}

\clearpage
\newpage

\appendix

\begin{center}
\Large
Supplementary Material for:\\Selection of Source Images Heavily Influences the Effectiveness of Adversarial Attacks
\end{center}

\section{$L_p$ norms and perturbation visibility}
\begin{figure*}[ht!]
\centering
\begin{tikzpicture}
\scriptsize
\centering
\def\sety1{0}
\node[inner sep=0pt] (a) at (0, \sety1)
    {\includegraphics[width=1.8cm]{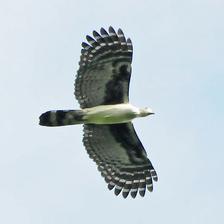}};
\node[align=center] at (0, \sety1-1.1) {Original image};
\node[inner sep=0pt] (a) at (2, \sety1)
    {\includegraphics[width=1.8cm]{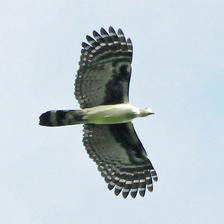}};
\node[align=center] at (2, \sety1-1.1) {$L_2=4.81$};
\node[align=center] at (2-0.06, \sety1-1.4) {$L_{\infty}=0.03$};
\node[inner sep=0pt] (b) at (4, \sety1)
    {\includegraphics[width=1.8cm]{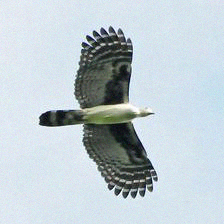}};
\node[align=center] at (4, \sety1-1.1) {$L_2=6.81$};
\node[align=center] at (4-0.06, \sety1-1.4) {$L_{\infty}=0.07$};
\node[inner sep=0pt] (c) at (6, \sety1)
    {\includegraphics[width=1.8cm]{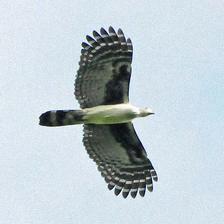}};
\node[align=center] at (6, \sety1-1.1) {$L_2=8.30$};
\node[align=center] at (6-0.06, \sety1-1.4) {$L_{\infty}=0.08$};
\node[inner sep=0pt] (a) at (8, \sety1)
    {\includegraphics[width=1.8cm]{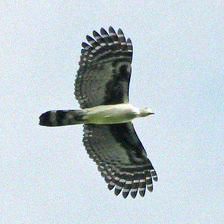}};
\node[align=center] at (8, \sety1-1.1) {$L_2=8.99$};
\node[align=center] at (8-0.06, \sety1-1.4) {$L_{\infty}=0.09$};
\node[inner sep=0pt] (b) at (10, \sety1)
    {\includegraphics[width=1.8cm]{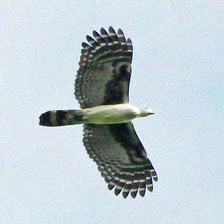}};
\node[align=center] at (10, \sety1-1.1) {$L_2=9.58$};
\node[align=center] at (10-0.06, \sety1-1.4) {$L_{\infty}=0.10$};

\def\sety1{-2.5}
\node[inner sep=0pt] (a) at (0, \sety1)
    {\includegraphics[width=1.8cm]{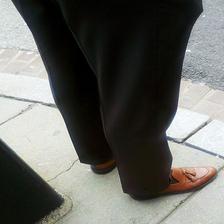}};
\node[align=center] at (0, \sety1-1.1) {Original image};
\node[inner sep=0pt] (a) at (2, \sety1)
    {\includegraphics[width=1.8cm]{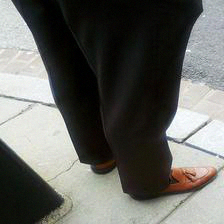}};
\node[align=center] at (2, \sety1-1.1) {$L_2=3.37$};
\node[align=center] at (2-0.06, \sety1-1.4) {$L_{\infty}=0.01$};
\node[inner sep=0pt] (b) at (4, \sety1)
    {\includegraphics[width=1.8cm]{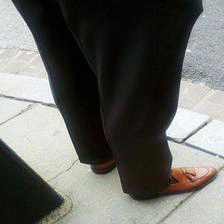}};
\node[align=center] at (4, \sety1-1.1) {$L_2=4.75$};
\node[align=center] at (4-0.06, \sety1-1.4) {$L_{\infty}=0.03$};
\node[inner sep=0pt] (c) at (6, \sety1)
    {\includegraphics[width=1.8cm]{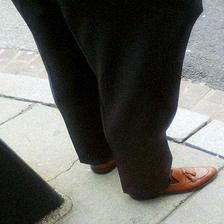}};
\node[align=center] at (6, \sety1-1.1) {$L_2=8.11$};
\node[align=center] at (6-0.06, \sety1-1.4) {$L_{\infty}=0.09$};
\node[inner sep=0pt] (a) at (8, \sety1)
    {\includegraphics[width=1.8cm]{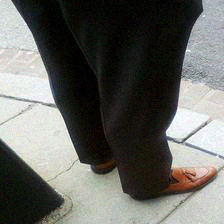}};
\node[align=center] at (8, \sety1-1.1) {$L_2=8.754$};
\node[align=center] at (8-0.06, \sety1-1.4) {$L_{\infty}=0.10$};
\node[inner sep=0pt] (b) at (10, \sety1)
    {\includegraphics[width=1.8cm]{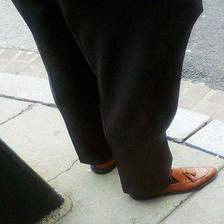}};
\node[align=center] at (10, \sety1-1.1) {$L_2=9.30$};
\node[align=center] at (10-0.06, \sety1-1.4) {$L_{\infty}=0.10$};
\def\sety1{-5}
\node[inner sep=0pt] (a) at (0, \sety1)
    {\includegraphics[width=1.8cm]{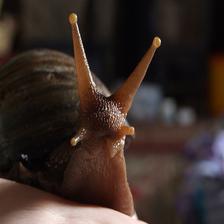}};
\node[align=center] at (0, \sety1-1.1) {Original image};
\node[inner sep=0pt] (a) at (2, \sety1)
    {\includegraphics[width=1.8cm]{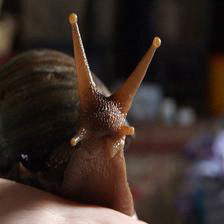}};
\node[align=center] at (2, \sety1-1.1) {$L_2=4.79$};
\node[align=center] at (2-0.06, \sety1-1.4) {$L_{\infty}=0.03$};
\node[inner sep=0pt] (b) at (4, \sety1)
    {\includegraphics[width=1.8cm]{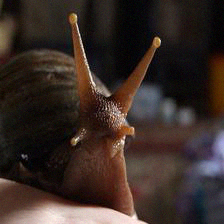}};
\node[align=center] at (4, \sety1-1.1) {$L_2=5.85$};
\node[align=center] at (4-0.06, \sety1-1.4) {$L_{\infty}=0.05$};
\node[inner sep=0pt] (c) at (6, \sety1)
    {\includegraphics[width=1.8cm]{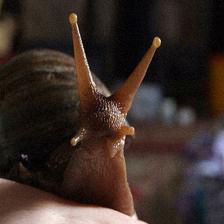}};
\node[align=center] at (6, \sety1-1.1) {$L_2=6.75$};
\node[align=center] at (6-0.06, \sety1-1.4) {$L_{\infty}=0.07$};
\node[inner sep=0pt] (a) at (8, \sety1)
    {\includegraphics[width=1.8cm]{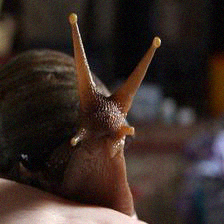}};
\node[align=center] at (8, \sety1-1.1) {$L_2=8.21$};
\node[align=center] at (8-0.06, \sety1-1.4) {$L_{\infty}=0.10$};
\node[inner sep=0pt] (b) at (10, \sety1)
    {\includegraphics[width=1.8cm]{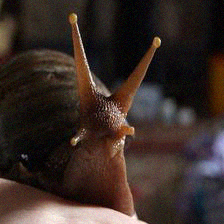}};
\node[align=center] at (10, \sety1-1.1) {$L_2=9.44$};
\node[align=center] at (10-0.06, \sety1-1.4) {$L_{\infty}=0.10$};
\def\sety1{-7.5}
\node[inner sep=0pt] (a) at (0, \sety1)
    {\includegraphics[width=1.8cm]{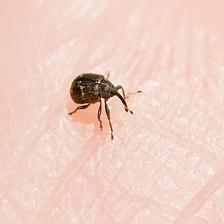}};
\node[align=center] at (0, \sety1-1.1) {Original image};
\node[inner sep=0pt] (a) at (2, \sety1)
    {\includegraphics[width=1.8cm]{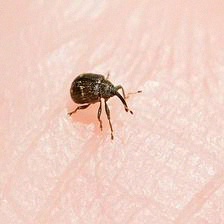}};
\node[align=center] at (2, \sety1-1.1) {$L_2=3.39$};
\node[align=center] at (2-0.06, \sety1-1.4) {$L_{\infty}=0.01$};
\node[inner sep=0pt] (b) at (4, \sety1)
    {\includegraphics[width=1.8cm]{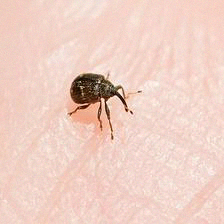}};
\node[align=center] at (4, \sety1-1.1) {$L_2=4.80$};
\node[align=center] at (4-0.06, \sety1-1.4) {$L_{\infty}=0.03$};
\node[inner sep=0pt] (c) at (6, \sety1)
    {\includegraphics[width=1.8cm]{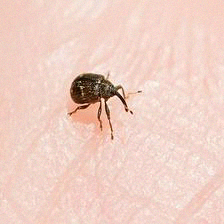}};
\node[align=center] at (6, \sety1-1.1) {$L_2=5.88$};
\node[align=center] at (6-0.06, \sety1-1.4) {$L_{\infty}=0.05$};
\node[inner sep=0pt] (a) at (8, \sety1)
    {\includegraphics[width=1.8cm]{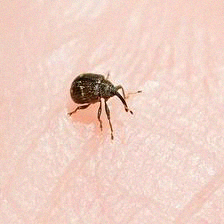}};
\node[align=center] at (8, \sety1-1.1) {$L_2=6.75$};
\node[align=center] at (8-0.06, \sety1-1.4) {$L_{\infty}=0.07$};
\node[inner sep=0pt] (b) at (10, \sety1)
    {\includegraphics[width=1.8cm]{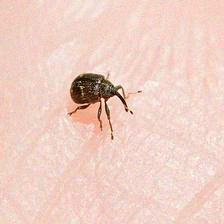}};
\node[align=center] at (10, \sety1-1.1) {$L_2=8.25$};
\node[align=center] at (10-0.06, \sety1-1.4) {$L_{\infty}=0.09$};
\def\sety1{-10}
\node[inner sep=0pt] (a) at (0, \sety1)
    {\includegraphics[width=1.8cm]{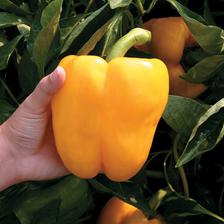}};
\node[align=center] at (0, \sety1-1.1) {Original image};
\node[inner sep=0pt] (a) at (2, \sety1)
    {\includegraphics[width=1.8cm]{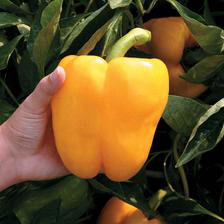}};
\node[align=center] at (2, \sety1-1.1) {$L_2=3.33$};
\node[align=center] at (2-0.06, \sety1-1.4) {$L_{\infty}=0.01$};
\node[inner sep=0pt] (b) at (4, \sety1)
    {\includegraphics[width=1.8cm]{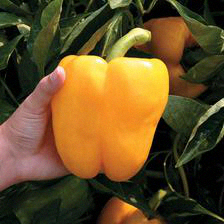}};
\node[align=center] at (4, \sety1-1.1) {$L_2=4.67$};
\node[align=center] at (4-0.06, \sety1-1.4) {$L_{\infty}=0.03$};
\node[inner sep=0pt] (c) at (6, \sety1)
    {\includegraphics[width=1.8cm]{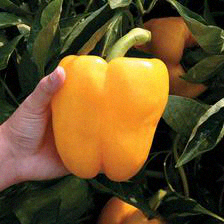}};
\node[align=center] at (6, \sety1-1.1) {$L_2=6.59$};
\node[align=center] at (6-0.06, \sety1-1.4) {$L_{\infty}=0.07$};
\node[inner sep=0pt] (a) at (8, \sety1)
    {\includegraphics[width=1.8cm]{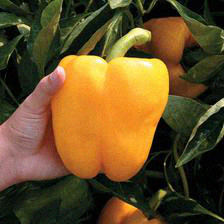}};
\node[align=center] at (8, \sety1-1.1) {$L_2=7.35$};
\node[align=center] at (8-0.06, \sety1-1.4) {$L_{\infty}=0.08$};
\node[inner sep=0pt] (b) at (10, \sety1)
    {\includegraphics[width=1.8cm]{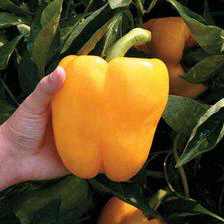}};
\node[align=center] at (10, \sety1-1.1) {$L_2=8.06$};
\node[align=center] at (10-0.06, \sety1-1.4) {$L_{\infty}=0.08$};
\def\sety1{-12.5}
\node[inner sep=0pt] (a) at (0, \sety1)
    {\includegraphics[width=1.8cm]{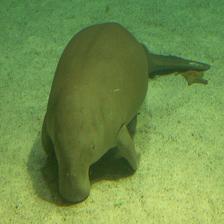}};
\node[align=center] at (0, \sety1-1.1) {Original image};
\node[inner sep=0pt] (a) at (2, \sety1)
    {\includegraphics[width=1.8cm]{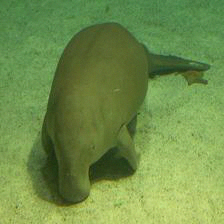}};
\node[align=center] at (2, \sety1-1.1) {$L_2=3.40$};
\node[align=center] at (2-0.06, \sety1-1.4) {$L_{\infty}=0.01$};
\node[inner sep=0pt] (b) at (4, \sety1)
    {\includegraphics[width=1.8cm]{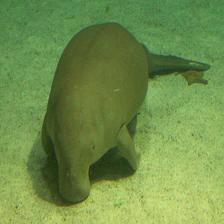}};
\node[align=center] at (4, \sety1-1.1) {$L_2=4.81$};
\node[align=center] at (4-0.06, \sety1-1.4) {$L_{\infty}=0.03$};
\node[inner sep=0pt] (c) at (6, \sety1)
    {\includegraphics[width=1.8cm]{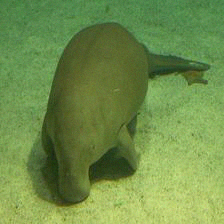}};
\node[align=center] at (6, \sety1-1.1) {$L_2=5.89$};
\node[align=center] at (6-0.06, \sety1-1.4) {$L_{\infty}=0.05$};
\node[inner sep=0pt] (a) at (8, \sety1)
    {\includegraphics[width=1.8cm]{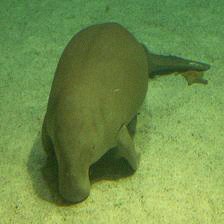}};
\node[align=center] at (8, \sety1-1.1) {$L_2=6.79$};
\node[align=center] at (8-0.06, \sety1-1.4) {$L_{\infty}=0.07$};
\node[inner sep=0pt] (b) at (10, \sety1)
    {\includegraphics[width=1.8cm]{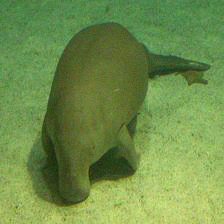}};
\node[align=center] at (10, \sety1-1.1) {$L_2=8.96$};
\node[align=center] at (10-0.06, \sety1-1.4) {$L_{\infty}=0.10$};
\def\sety1{-15}
\node[inner sep=0pt] (a) at (0, \sety1)
    {\includegraphics[width=1.8cm]{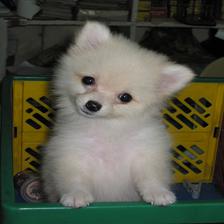}};
\node[align=center] at (0, \sety1-1.1) {Original image};
\node[inner sep=0pt] (a) at (2, \sety1)
    {\includegraphics[width=1.8cm]{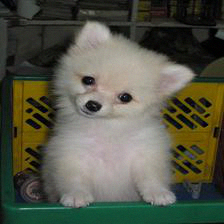}};
\node[align=center] at (2, \sety1-1.1) {$L_2=3.39$};
\node[align=center] at (2-0.06, \sety1-1.4) {$L_{\infty}=0.01$};
\node[inner sep=0pt] (b) at (4, \sety1)
    {\includegraphics[width=1.8cm]{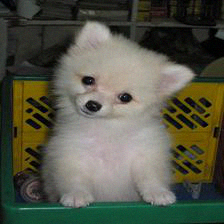}};
\node[align=center] at (4, \sety1-1.1) {$L_2=4.77$};
\node[align=center] at (4-0.06, \sety1-1.4) {$L_{\infty}=0.03$};
\node[inner sep=0pt] (c) at (6, \sety1)
    {\includegraphics[width=1.8cm]{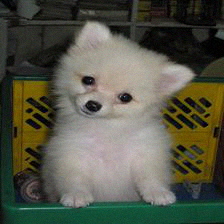}};
\node[align=center] at (6, \sety1-1.1) {$L_2=5.84$};
\node[align=center] at (6-0.06, \sety1-1.4) {$L_{\infty}=0.05$};
\node[inner sep=0pt] (a) at (8, \sety1)
    {\includegraphics[width=1.8cm]{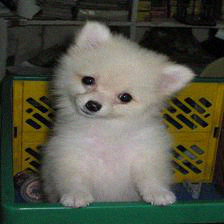}};
\node[align=center] at (8, \sety1-1.1) {$L_2=6.75$};
\node[align=center] at (8-0.06, \sety1-1.4) {$L_{\infty}=0.07$};
\node[inner sep=0pt] (b) at (10, \sety1)
    {\includegraphics[width=1.8cm]{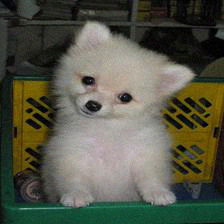}};
\node[align=center] at (10, \sety1-1.1) {$L_2=8.22$};
\node[align=center] at (10-0.06, \sety1-1.4) {$L_{\infty}=0.09$};
\end{tikzpicture}
\vspace{1em}
\caption{Application of adversarial perturbations to images.}
\label{fig:additional_examples-pgd}
\end{figure*}

Although we guarantee the discretization property, in order to maintain comparability with the literature, the perturbation amounts reported in the main text (both $L_2$ and $L_{\infty}$) are calculated based on the assumption that pixel values lie in $[0, 1]$. Based on this, we calculate the $L_2$ and $L_{\infty}$ distance between two vectors with size $k = 3 \times 224 \times 224$ (channel $\times$ height $\times$ width) as follows:
\begin{align}
    L_{2}(\bm{x}, \hat{\bm{x}}) & = ||\bm{x} - \hat{\bm{x}}||_2 \,,\\
    L_{\infty}(\bm{x}, \hat{\bm{x}}) & = \max(|\bm{x} - \hat{\bm{x}}|) \,,
\end{align}
where $\bm{x}$ and $\hat{\bm{x}}$ represent an initial (source) image and its adversarial counterpart, respectively. In Figure~\ref{fig:additional_examples-pgd}, we provide a number of qualitative examples that illustrate the measurement of perturbation visibility.

\section{Non-adversarial perturbations}


In the main text, we compare the adversarial transferability of images modified through adversarial perturbation with that of images changed through non-adversarial noise. The different types of non-adversarial noise we employ are (1) uniform noise, (2) normal noise, and (3) change in contrast. For the aforementioned types of noise, we initialize a vector $\bm{p}=\bm{0}\in\mathbb{R}^k$ that has the same size as the input, filling its values as described below, with all non-adversarial perturbation generation methods respecting the $L_{\infty}$ perturbation limit set for PGD, hereby using $\Pi_{\epsilon=38}$.


\textbf{Uniform noise}\,\textendash\,Similar to the usage of PGD, we rely on an iterative approach for the application of uniform noise. As such, each of the elements of $\bm{p}$ is sampled from a uniform distribution $\mathcal{U}[-1,1]$. However, instead of using the values themselves, we use their signature, applying perturbation as follows:
\begin{align}
[\hat{\bm{x}}]_{n+1} &= \Pi_{\epsilon}( [\hat{\bm{x}}]_{n} + [\bm{p}]_{n}) \,, [p_k]_{n} \sim  \text{sign} (\mathcal{U}[-1, 1]) \,.
\end{align}
with $[\hat{\bm{x}}]_{1} = \bm{x}$. Similar to the usage of PGD, if the \say{adversarial example} created this way does not achieve model-to-model transferability, we perform the same operation four more times.



\textbf{Gaussian noise}\,\textendash\,Instead of an iterative approach, we follow a different methodology for the application of Gaussian noise. We sample only one noise vector, with every element of this vector originating from a Gaussian distribution with zero mean and standard deviation $10$. We then apply this noise vector to the data point at hand as follows:
\begin{align}
\hat{\bm{x}} = \Pi_{\epsilon}( \bm{x} + \bm{p}) \,, p_k \sim  \mathcal{N}(0, 10^2) \,.
\end{align}
If the resulting image does not achieve adversarial transferability, we perform the same operation up to ten times more, with newly sampled values from the same normal distribution.

\textbf{Change in contrast}\,\textendash\,A change in contrast in the image domain means that all pixel values are modified with the same value. To that end, we evaluate all possible values within the allowed $L_{\infty}$ limit, creating a set of adversarial examples originating from an input image as follows:
\begin{align}
\widehat{\mathcal{X}} := \{ \hat{\bm{x}}_b \, | \, \hat{\bm{x}}_b =  \bm{x} + \mathbf{1} * b,  b \in  \{-38, \ldots, 38\} \} \,.
\end{align}

\section{Detailed transferability graphs}
In Figure~\ref{fig:transferability_matrix_sup_all} and Figure~\ref{fig:non_adv_transferability_matrix_all}, we provide the model-to-model transferability plots presented in Figure~2 in the main text, but in a higher resolution and with more details. In addition to the untargeted transferability details provided in the aforementioned figures, in Figure~\ref{fig:transferability_matrix_targeted}, we provide the targeted adversarial transferability success of the produced adversarial examples.

In Figure~\ref{fig:transferability_matrix_PGD}, Figure~\ref{fig:transferability_matrix_CW}, and Figure~\ref{fig:transferability_matrix_MIFGSM}, we provide detailed model-to-model transferability details for (left) fragile and (right) hard images, respectively, as identified with the help of non-adversarial perturbations.

In Figure~3(b) of the main text, we histogrammed $\overline{T}(\Theta, \widehat{\mathcal{X}}^{(A)}, \bm{y})$ for all adversarial examples, hereby displaying the transferability count of the source images. In Figure~\ref{fig:all_hists}, we provide the same information with $T(\Theta, \widehat{\mathcal{X}}^{(A)}, \bm{y})$, but specifically for adversarial examples created through the use of individual attacks.


\section{Correlation between transferability and perturbation}
In Figure~\ref{fig:transferability_perturbation}, we plot the adversarial transferability count for each source image, as obtained with $T(\Theta, \widehat{\mathcal{X}}^{(A)}, \bm{y})$, against the minimum required $L_p$ perturbation to achieve adversarial transferability $D_{\{2,\infty\}}(\Theta, \widehat{\mathcal{X}}^{(\text{A})})$, for all adversarial examples, as well as the subset of adversarial examples produced with individual attacks. Here, we observe a mild negative correlation between the added noise and the transferability count, where the adversarial examples originating from source images that achieve higher transferability counts are also the ones that require less perturbation.

\section{Required perturbation for adversarial transferability}
In Figure~5 of the main text, we provided, for ViT-B, the $L_{\{2, \infty\}}$ norms of adversarial perturbations obtained through the usage of a number of source images, where this number is progressively reduced based on the transferability count of those images. From Figure~\ref{fig:pert_norm_alexnet} to Figure~\ref{fig:pert_norm_vitl}, we provide the same results for the other models and for all adversarial attacks.

\clearpage

\begin{figure*}[t!]
    \centering
        \includegraphics[width=0.48\linewidth]{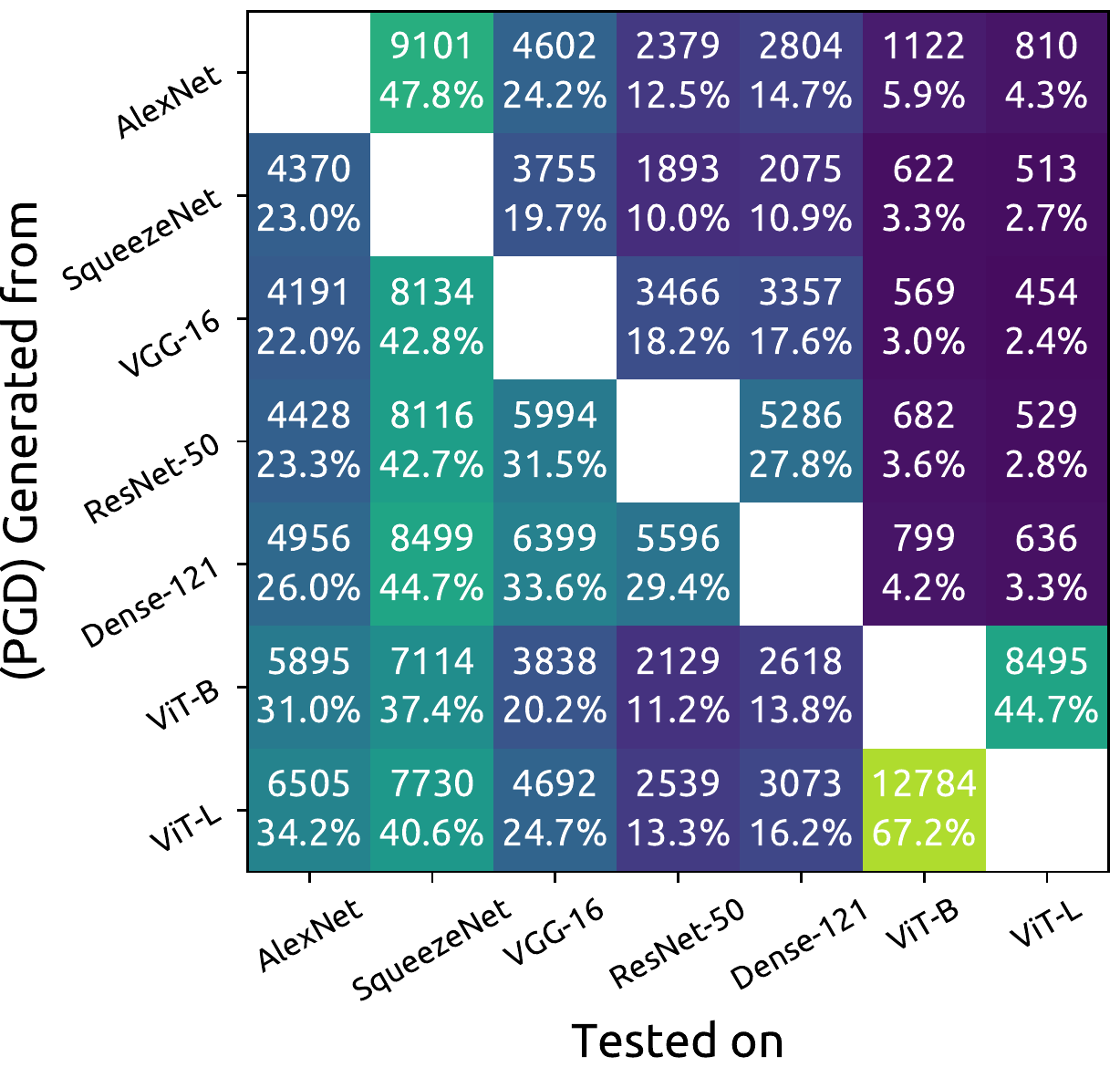}
        \includegraphics[width=0.48\linewidth]{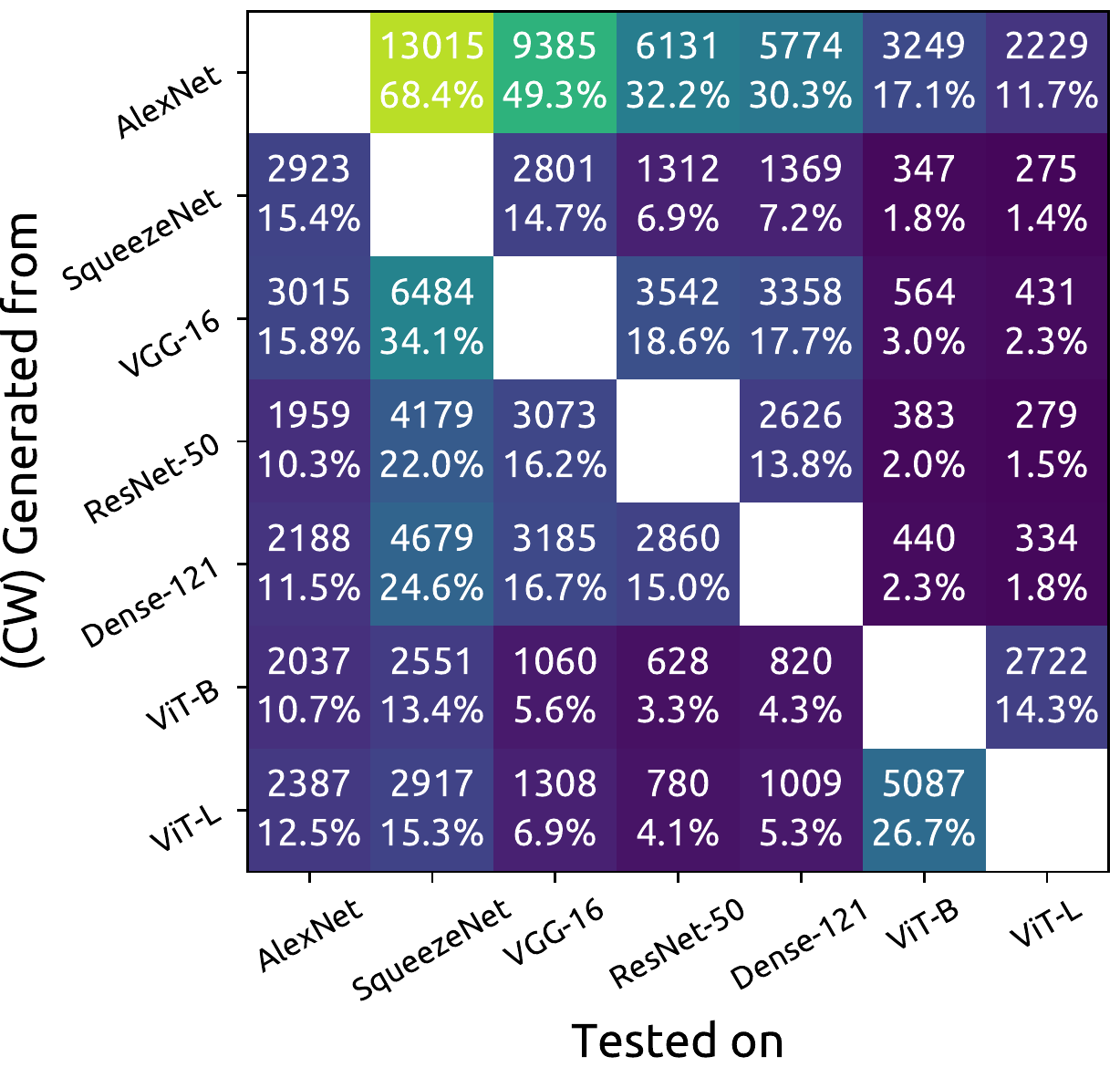}
        \includegraphics[width=0.48\linewidth]{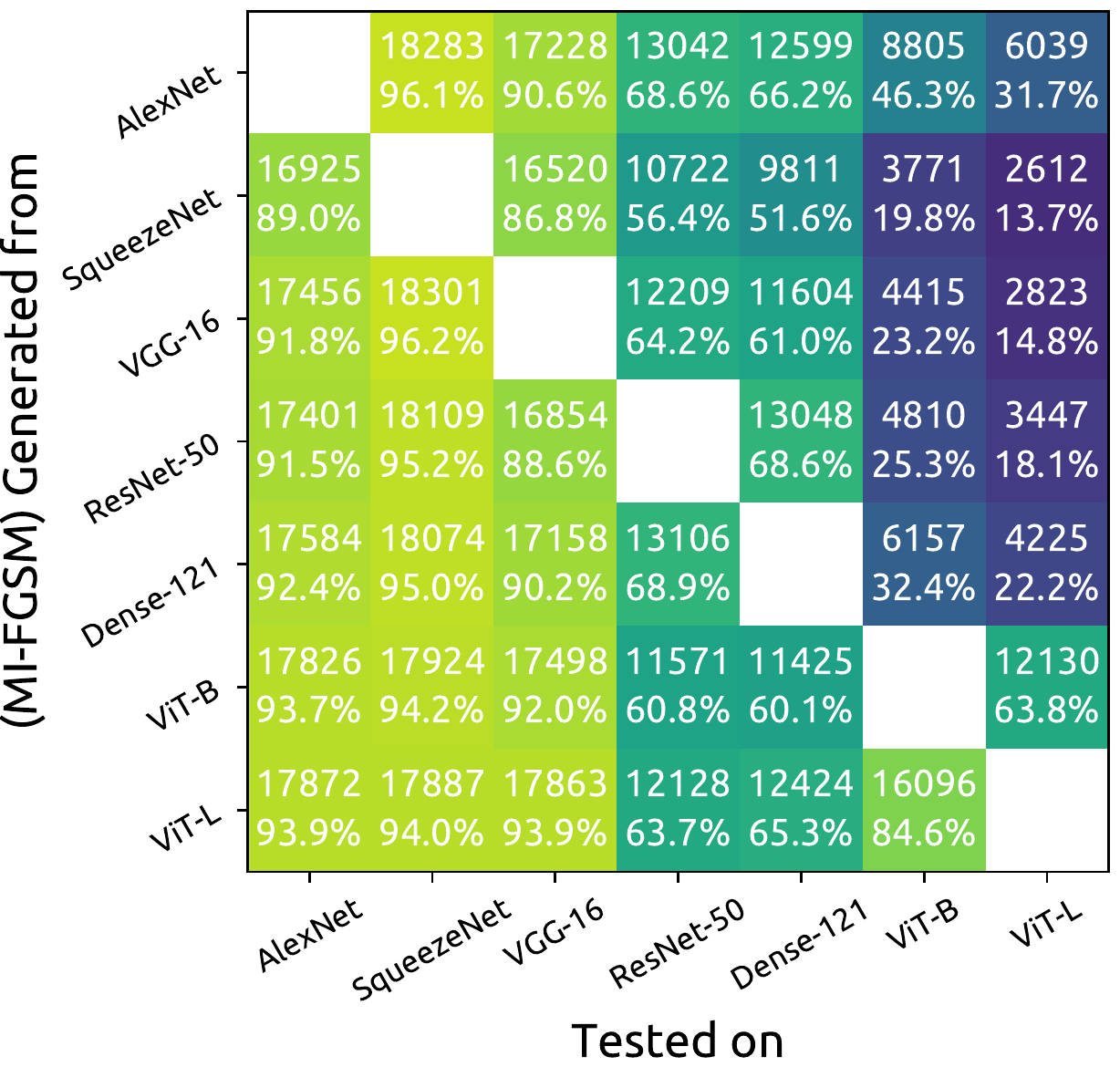}
      \vspace{2em}
      \caption{Number (proportion) of source images that achieved (untargeted) adversarial transferability through the usage of (left) PGD, (right) CW, and (bottom) MI-FGSM. Adversarial examples are generated from the models listed on the $y$-axis and are tested on the models listed on the $x$-axis.}
      \label{fig:transferability_matrix_sup_all}
\end{figure*}

\begin{figure*}[t]
\centering
\includegraphics[width=0.55\linewidth]{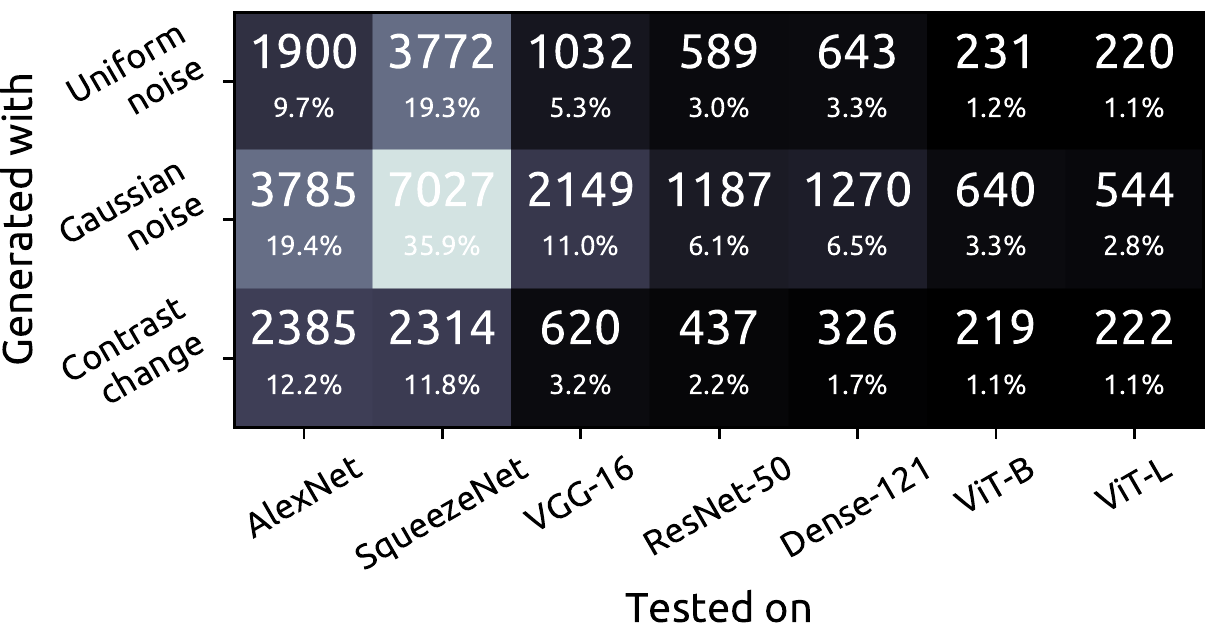}
\vspace{2em}
\caption{Number (proportion) of source images that have their classification changed through the usage of non-adversarial perturbation.}
\label{fig:non_adv_transferability_matrix_all}
\end{figure*}

\clearpage
\begin{figure*}[t!]
    \centering
        \includegraphics[width=0.48\linewidth]{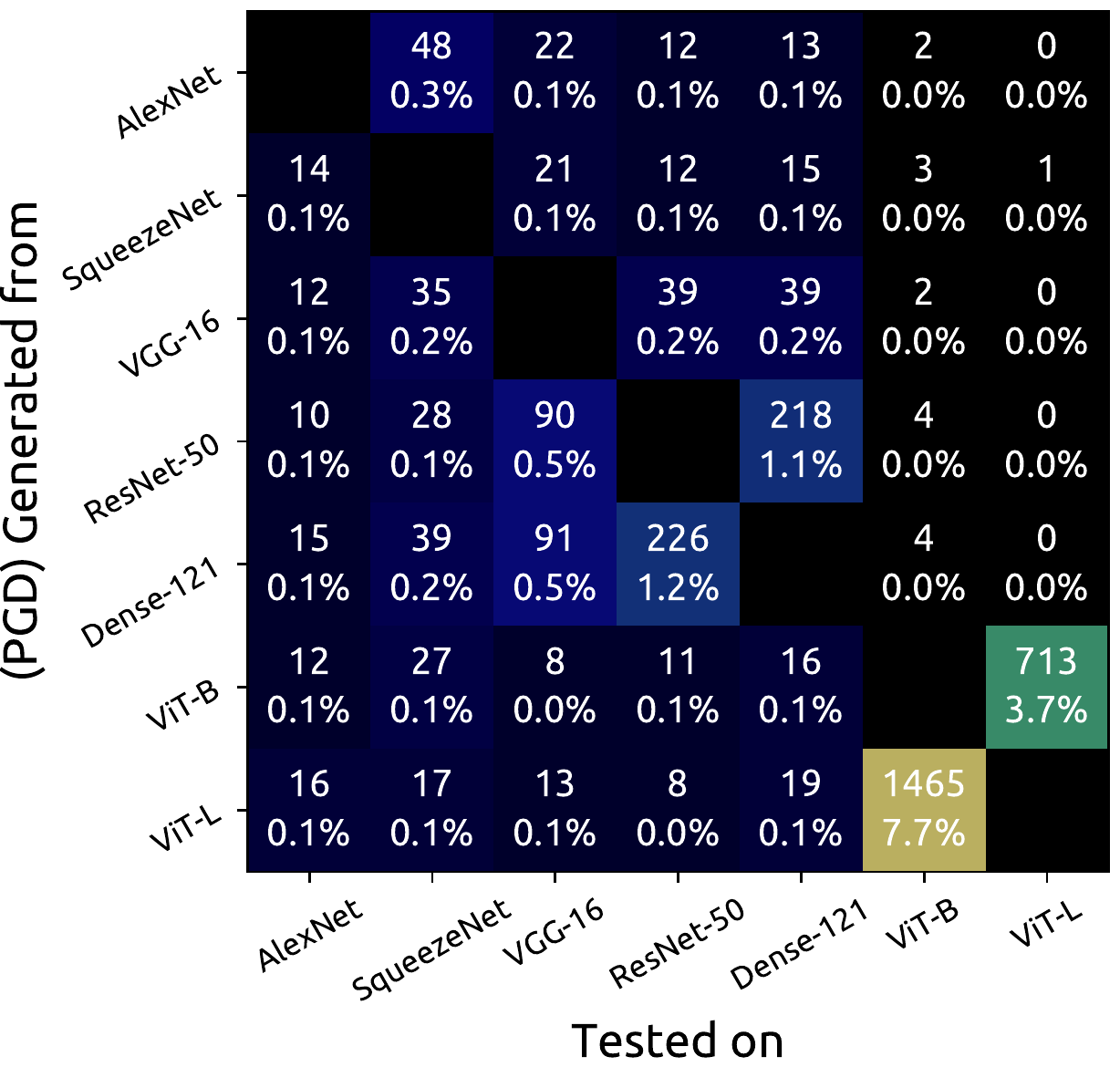}
        \includegraphics[width=0.48\linewidth]{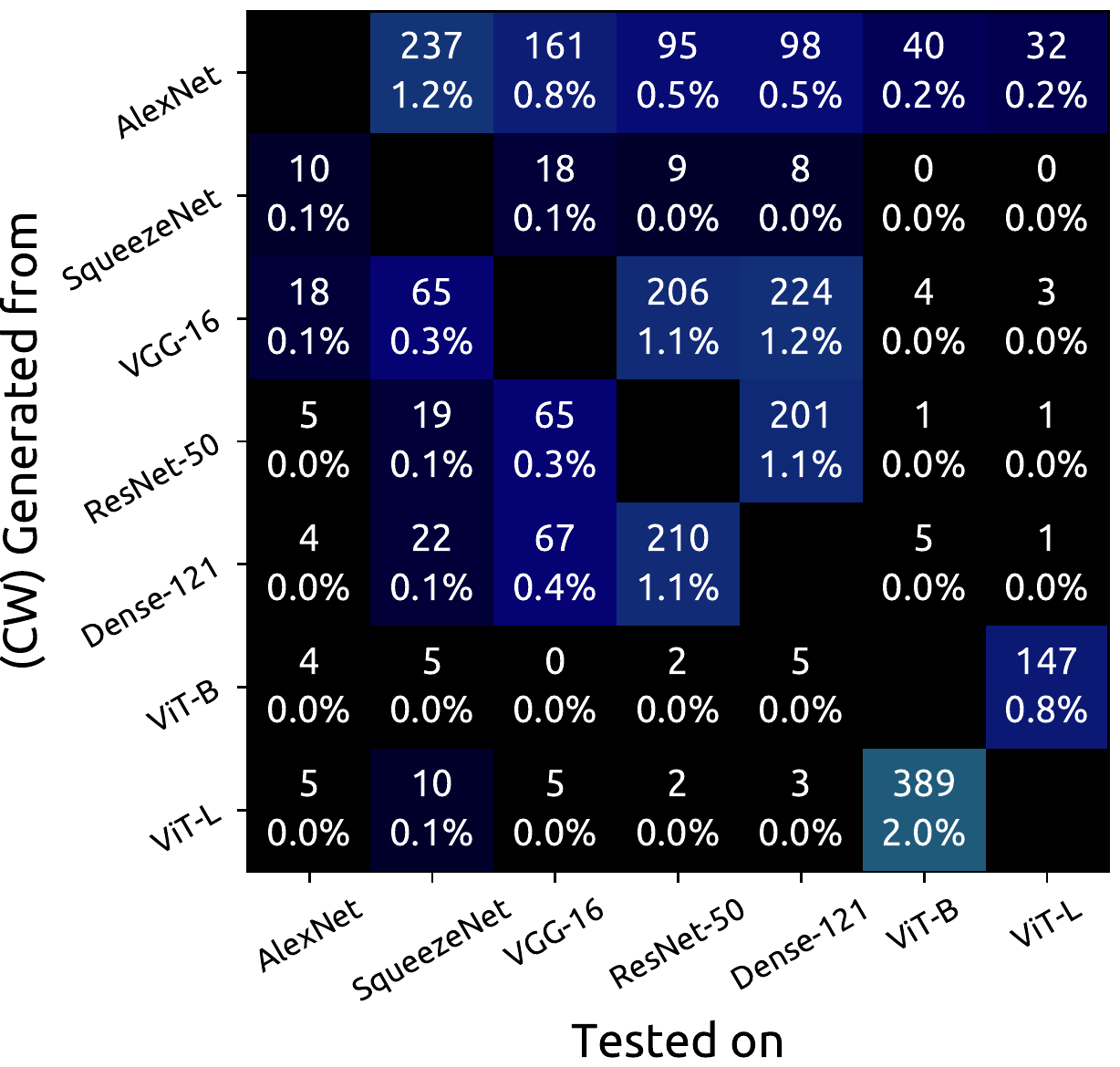}
        \includegraphics[width=0.48\linewidth]{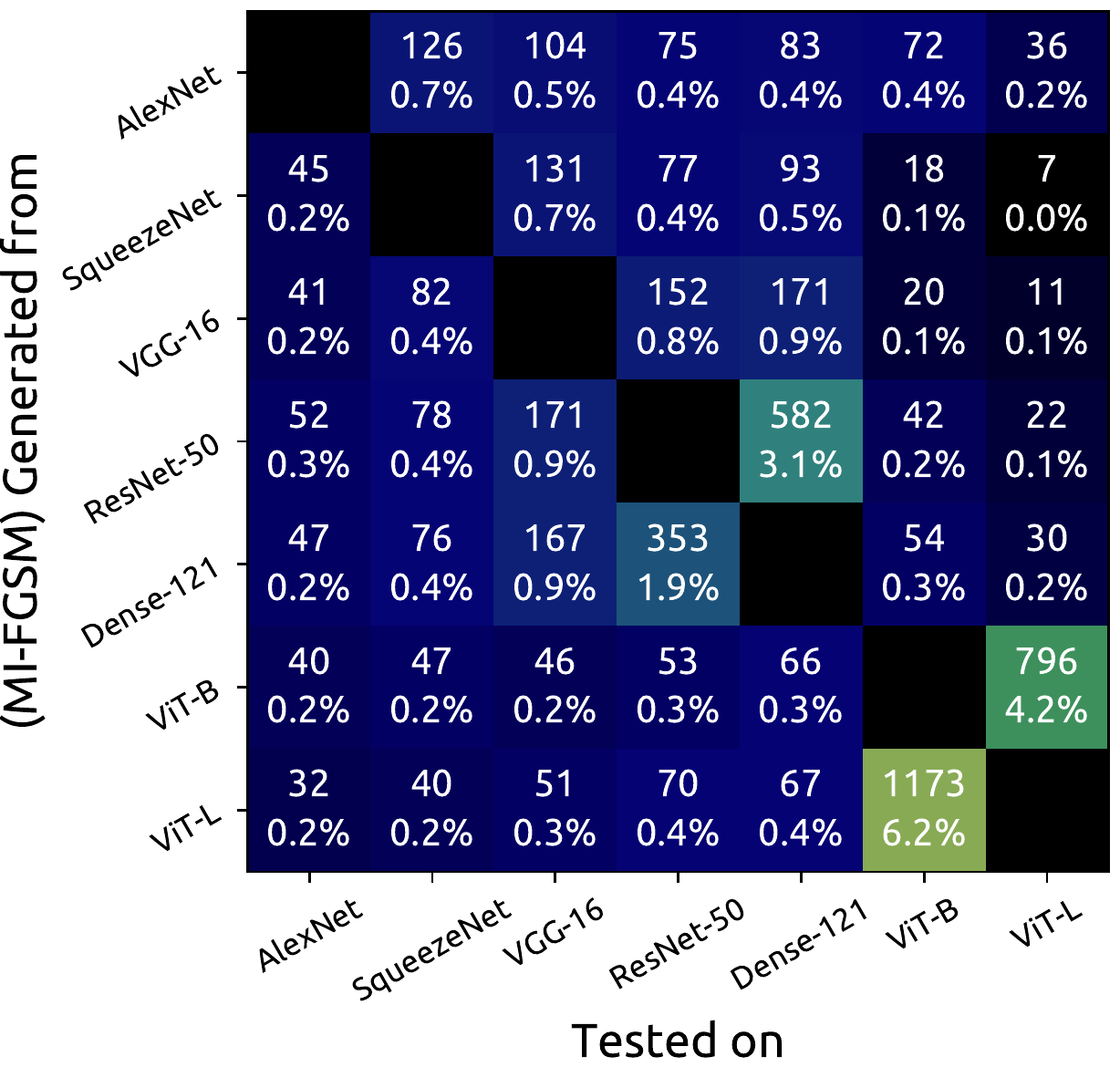}
      \vspace{2em}
      \caption{Number (proportion) of source images that achieved (targeted) adversarial transferability through the usage of (left) PGD, (right) CW, and (bottom) MI-FGSM. Adversarial examples are generated from the models listed on the $y$-axis and are tested on the models listed on the $x$-axis.}
      \label{fig:transferability_matrix_targeted}
\end{figure*}


\begin{figure*}[t!]
    \centering
    \begin{subfigure}{0.48\textwidth}
\includegraphics[width=\linewidth]{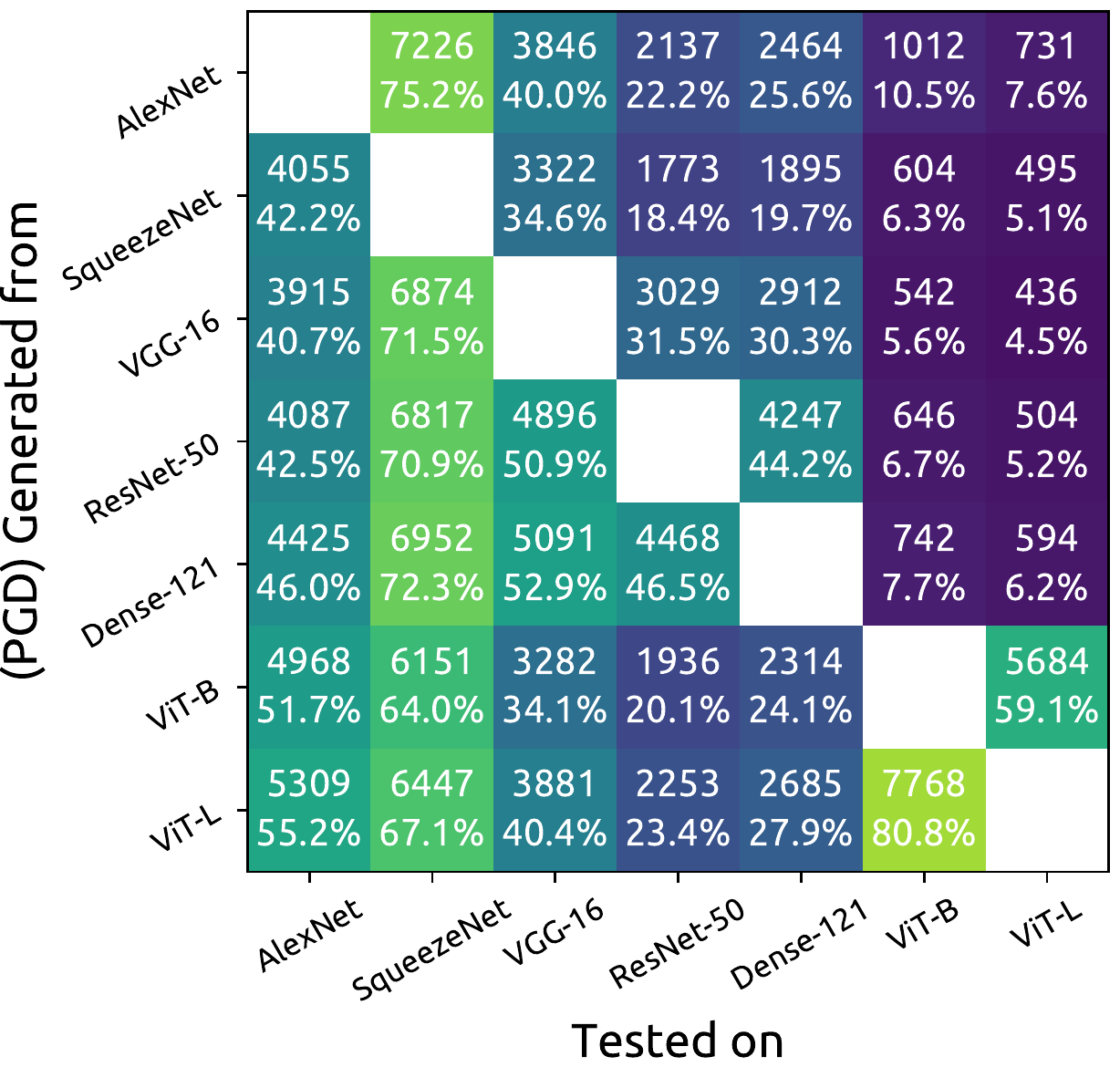}
\caption{Fragile images $(\mathbb{S}_f)$}
\end{subfigure}
    \begin{subfigure}{0.48\textwidth}
\includegraphics[width=\linewidth]{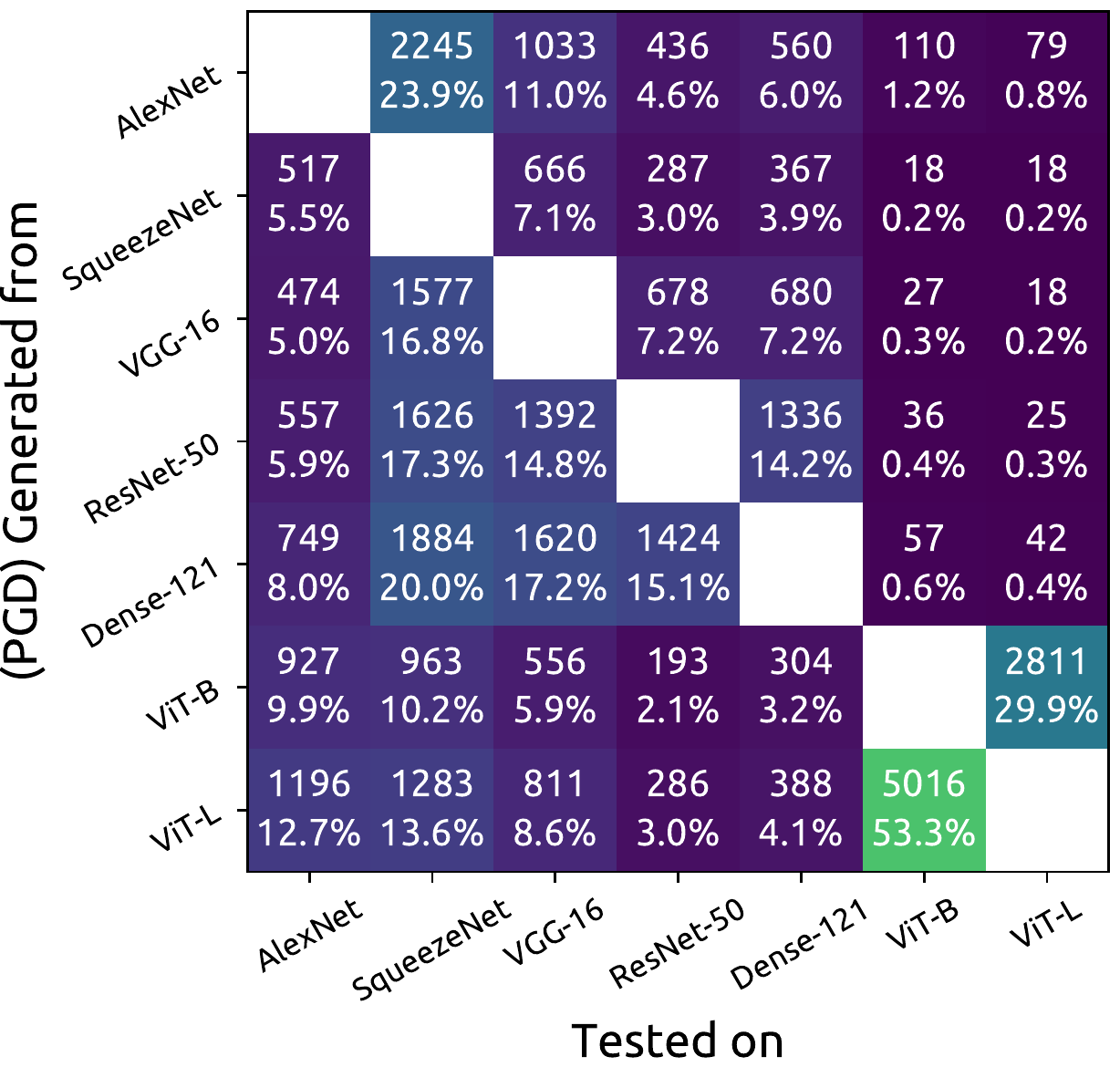}
\caption{Hard images $(\mathbb{S}_h)$}
\end{subfigure}
      \vspace{2em}
      \caption{Number (proportion) of source images that achieved (untargeted) adversarial transferability through the usage of \textbf{PGD} for source images taken from (left) \underline{$\mathbb{S}_f$}, and (right) \underline{$\mathbb{S}_h$}. Adversarial examples are generated from the models listed on the $y$-axis and are tested on the models listed on the $x$-axis.}
      \label{fig:transferability_matrix_PGD}
\end{figure*}

\begin{figure*}[t!]
    \centering
    \begin{subfigure}{0.48\textwidth}
\includegraphics[width=\linewidth]{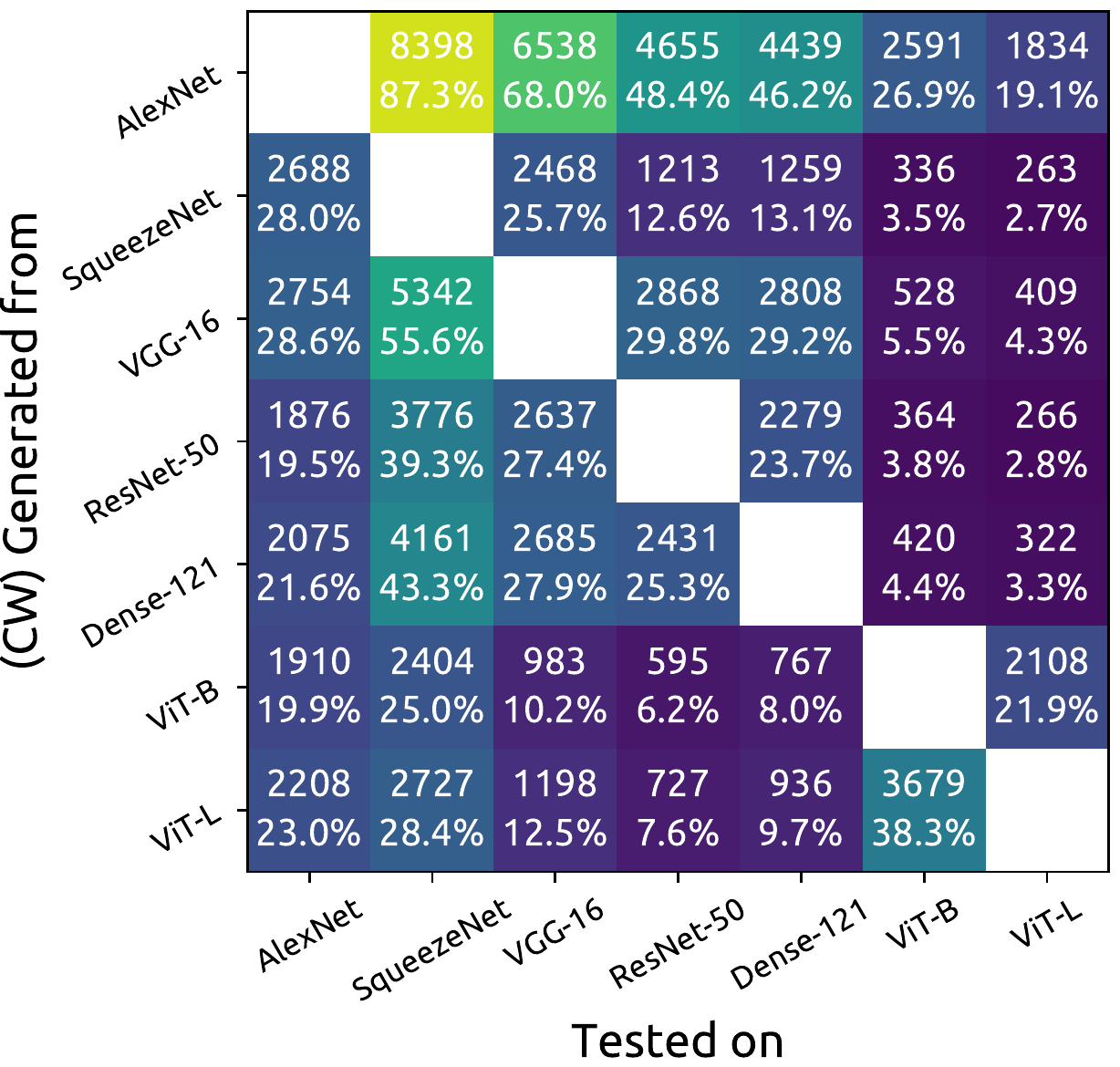}
\caption{Fragile images $(\mathbb{S}_f)$}
\end{subfigure}
    \begin{subfigure}{0.48\textwidth}
\includegraphics[width=\linewidth]{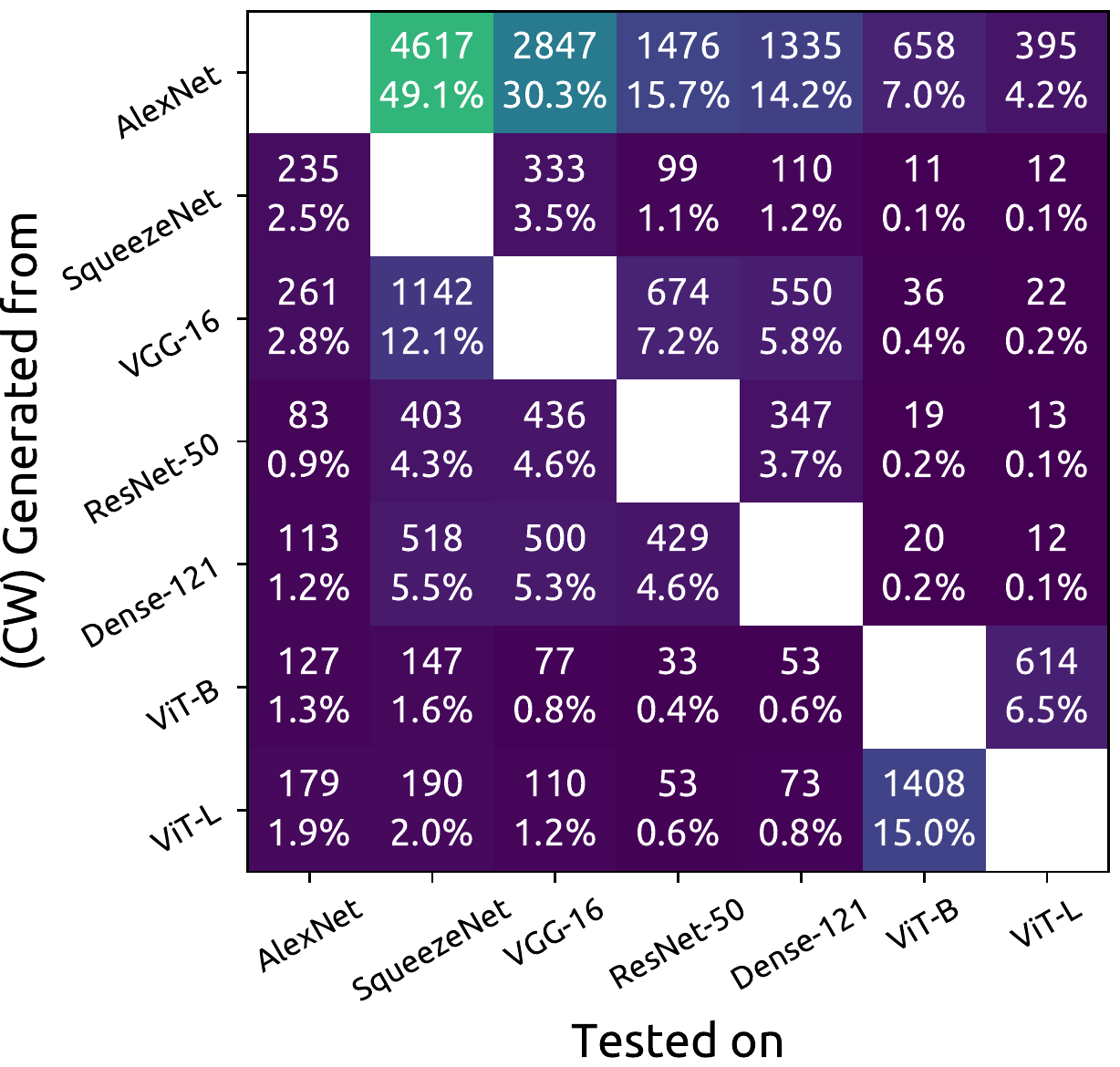}
\caption{Hard images $(\mathbb{S}_h)$}
\end{subfigure}
      \vspace{2em}
      \caption{Number (proportion) of source images that achieved (untargeted) adversarial transferability through the usage of \textbf{CW} for source images taken from (left) \underline{$\mathbb{S}_f$}, and (right) \underline{$\mathbb{S}_h$}. Adversarial examples are generated from the models listed on the $y$-axis and are tested on the models listed on the $x$-axis.}
      \label{fig:transferability_matrix_CW}
\end{figure*}

\begin{figure*}[t!]
    \centering
    \begin{subfigure}{0.48\textwidth}
\includegraphics[width=\linewidth]{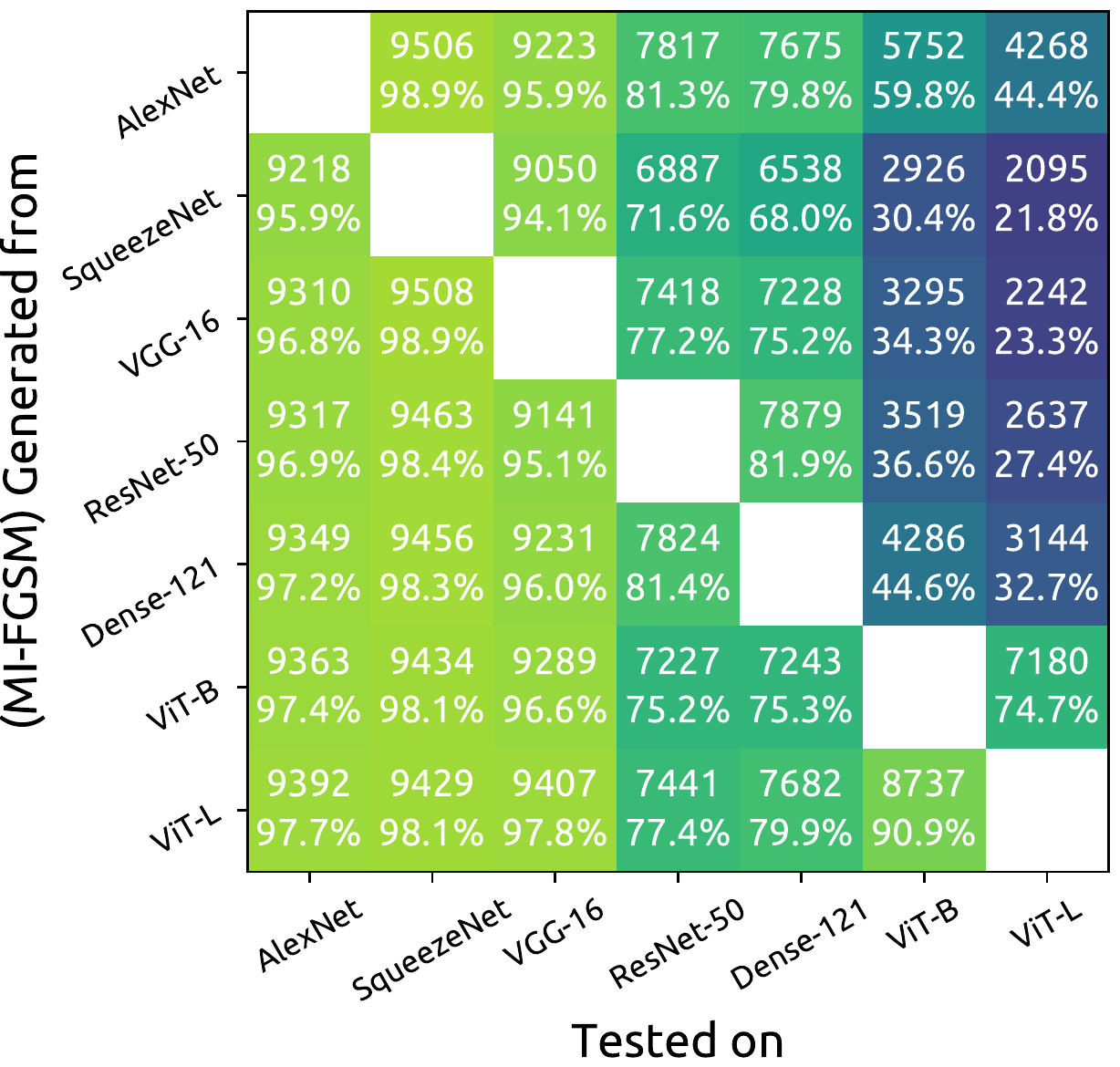}
\caption{Fragile images $(\mathbb{S}_f)$}
\end{subfigure}
    \begin{subfigure}{0.48\textwidth}
\includegraphics[width=\linewidth]{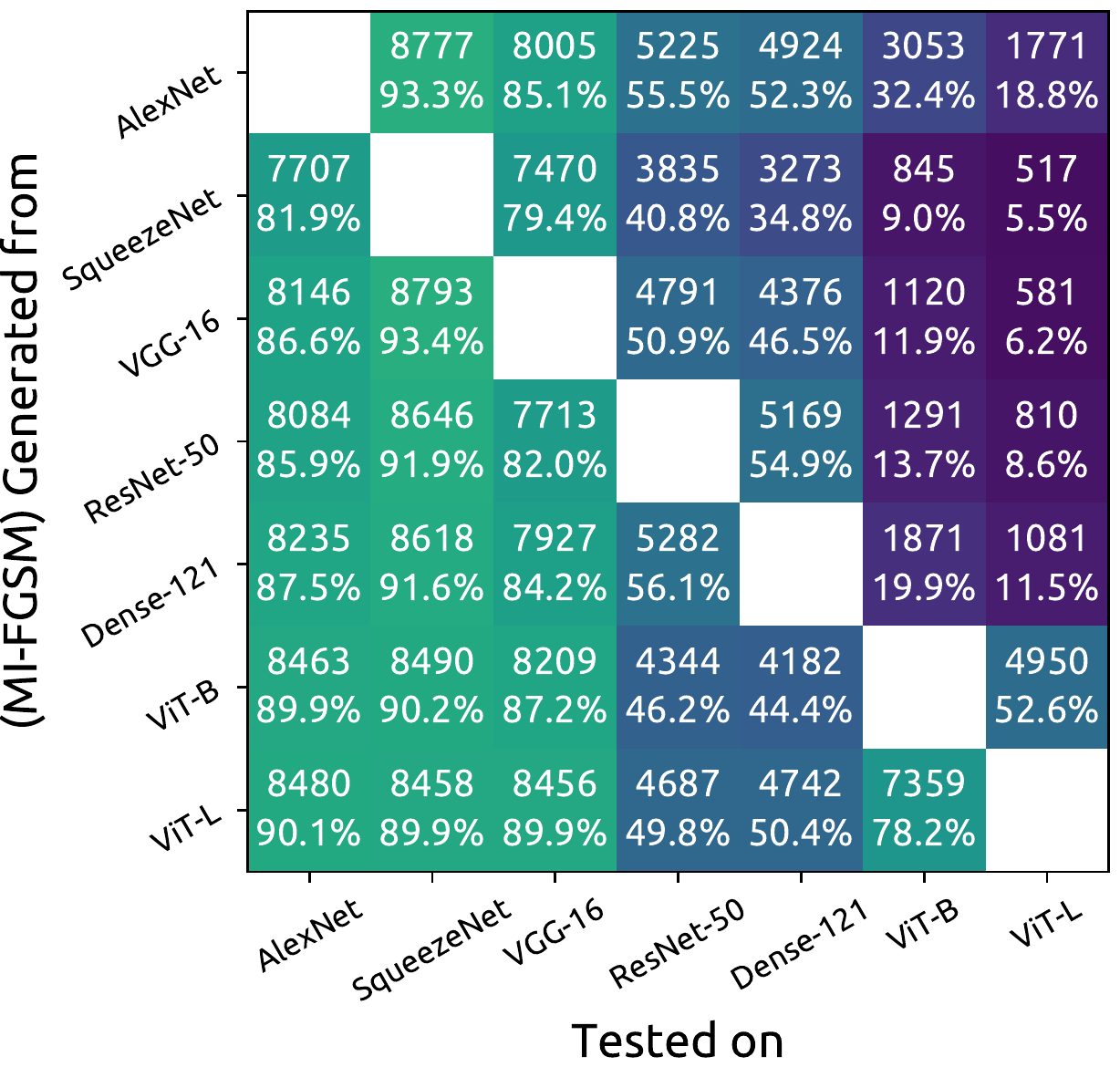}
\caption{Hard images $(\mathbb{S}_h)$}
\end{subfigure}
      \vspace{2em}
      \caption{Number (proportion) of source images that achieved adversarial transferability through the usage of \textbf{MI-FGSM} for source images taken from (left) \underline{$\mathbb{S}_f$}, and (right) \underline{$\mathbb{S}_h$}. Adversarial examples are generated from the models listed on the $y$-axis and are tested on the models listed on the $x$-axis.}
      \label{fig:transferability_matrix_MIFGSM}
\end{figure*}

\clearpage
\begin{figure}[t!]
\centering
\begin{subfigure}{0.51\textwidth}
\includegraphics[width=\textwidth]{bmvc_trans_cnt/im_trans_cnt.pdf}
\caption{All adversarial examples}
\label{fig:tr1}
\end{subfigure}
\begin{subfigure}{0.51\textwidth}
\includegraphics[width=\textwidth]{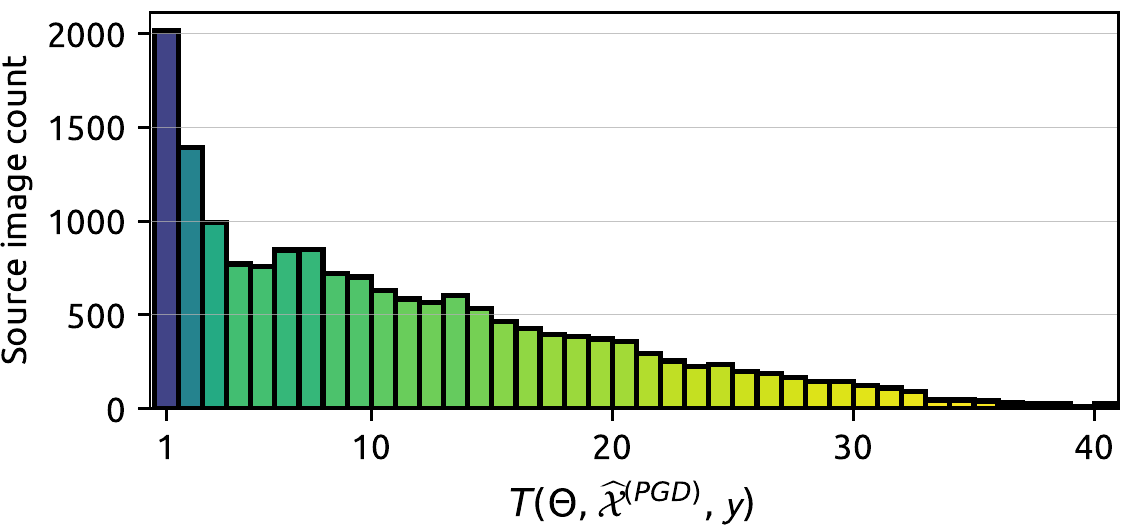}
\caption{All adversarial examples produced with PGD}
\label{fig:tr2}
\end{subfigure}
\begin{subfigure}{0.51\textwidth}
\includegraphics[width=\textwidth]{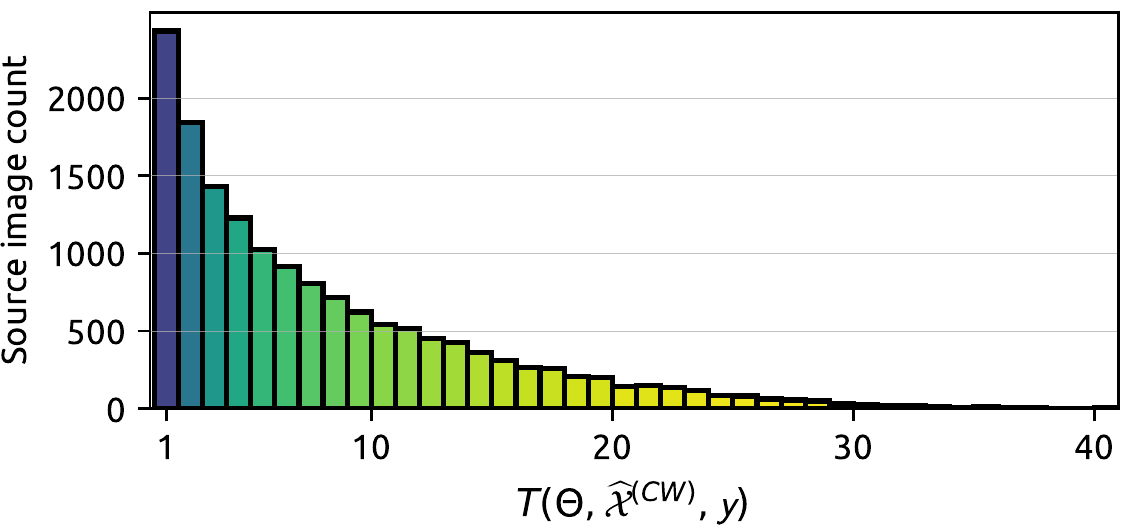}
\caption{All adversarial examples produced with CW}
\label{fig:tr3}
\end{subfigure}
\begin{subfigure}{0.51\textwidth}
\includegraphics[width=\textwidth]{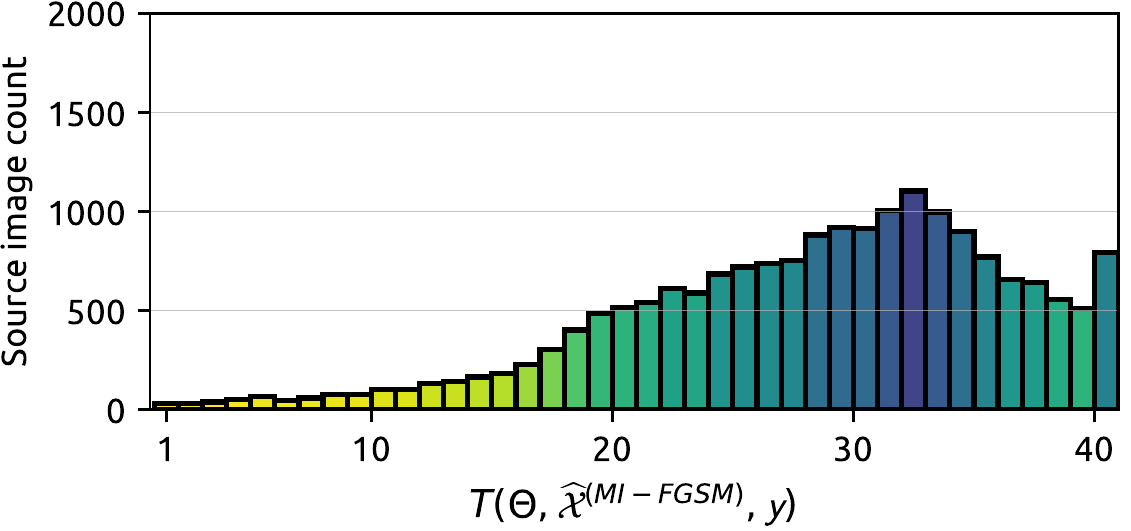}
\caption{All adversarial examples produced with \mbox{MI-FGSM}}
\label{fig:tr4}
\end{subfigure}
\vspace{1em}
\caption{Histogram of source images and their transferability count according to $T(\Theta, \widehat{\mathcal{X}}^{(\text{A})}, \bm{y})$, calculated with (top) all adversarial examples and (bottom three) individual attacks.}
\label{fig:all_hists}
\end{figure}

\clearpage
\begin{figure}[t!]
\centering
\includegraphics[width=0.5\textwidth]{bmvc_pert_vs_trans/A_min_transferability_l2_perturbation.pdf}
\includegraphics[width=0.35\textwidth]{bmvc_pert_vs_trans/A_min_transferability_linf_perturbation.pdf}
\vspace{1em}
\caption*{\qquad\,\,\,\,\, Correlation: $-0.67$ \qquad\qquad\,\,\,\,\,\,\,\,\,\,\,\,\,\,\, Correlation: $-0.61$}
\vspace{1em}
\includegraphics[width=0.5\textwidth]{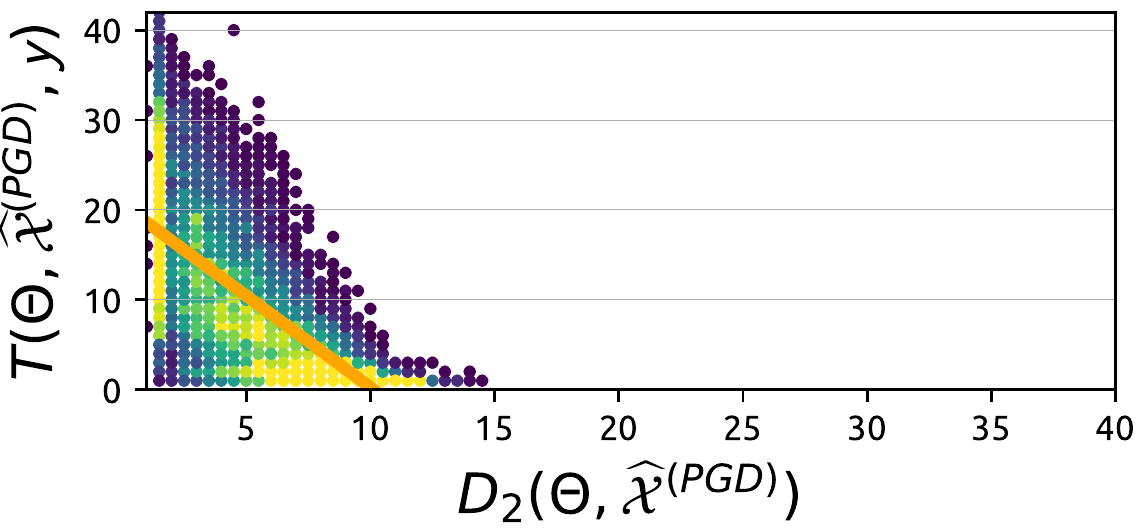}
\includegraphics[width=0.35\textwidth]{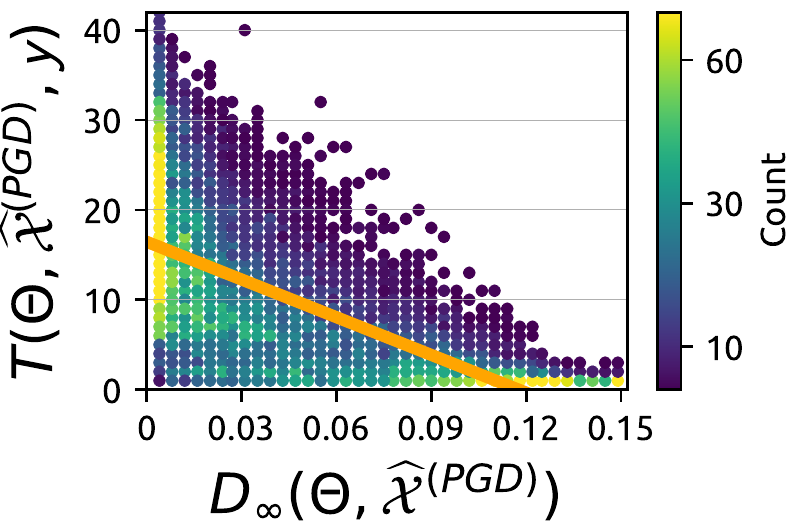}
\vspace{1em}
\caption*{\qquad\,\,\,\,\, Correlation: $-0.61$ \qquad\qquad\,\,\,\,\,\,\,\,\,\,\,\,\,\,\, Correlation: $-0.65$}
\vspace{1em}
\includegraphics[width=0.5\textwidth]{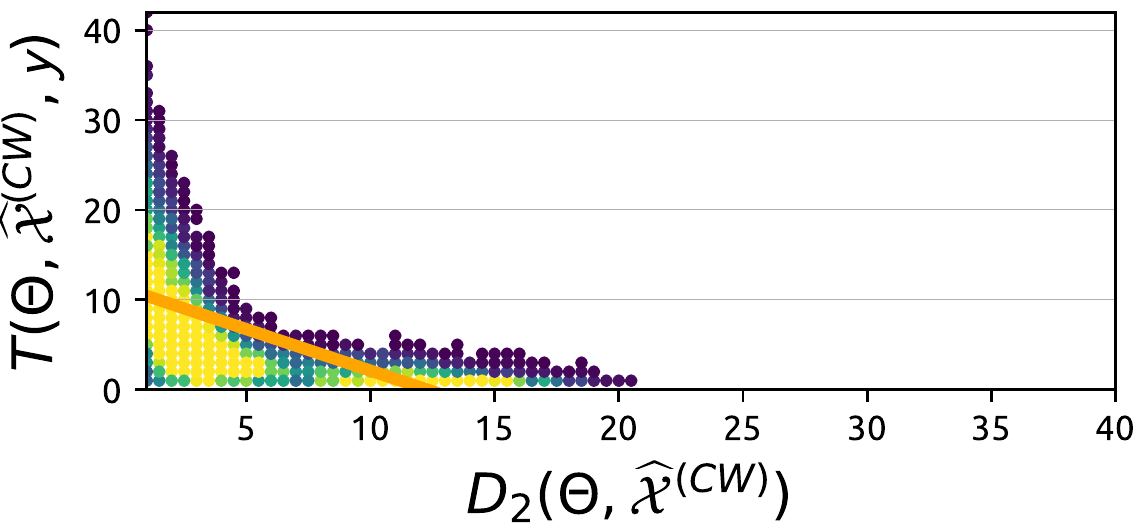}
\includegraphics[width=0.35\textwidth]{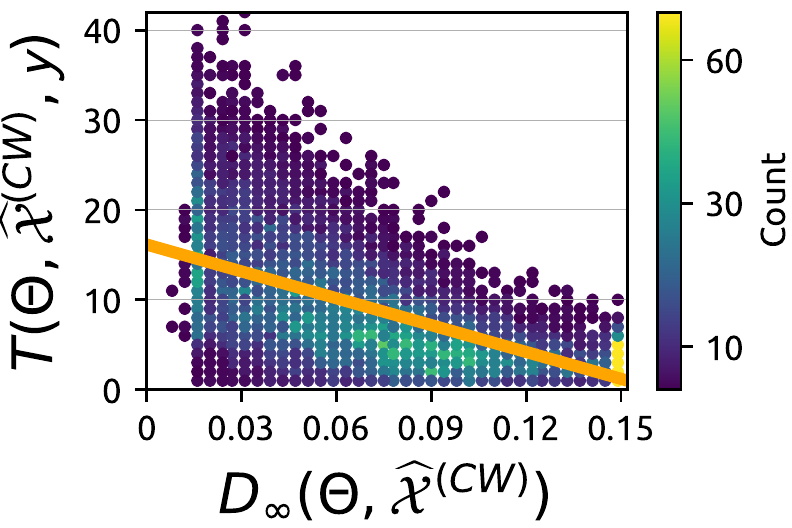}
\vspace{1em}
\caption*{\qquad\,\,\,\,\, Correlation: $-0.70$ \qquad\qquad\,\,\,\,\,\,\,\,\,\,\,\,\,\,\, Correlation: $-0.60$}
\vspace{1em}
\includegraphics[width=0.5\textwidth]{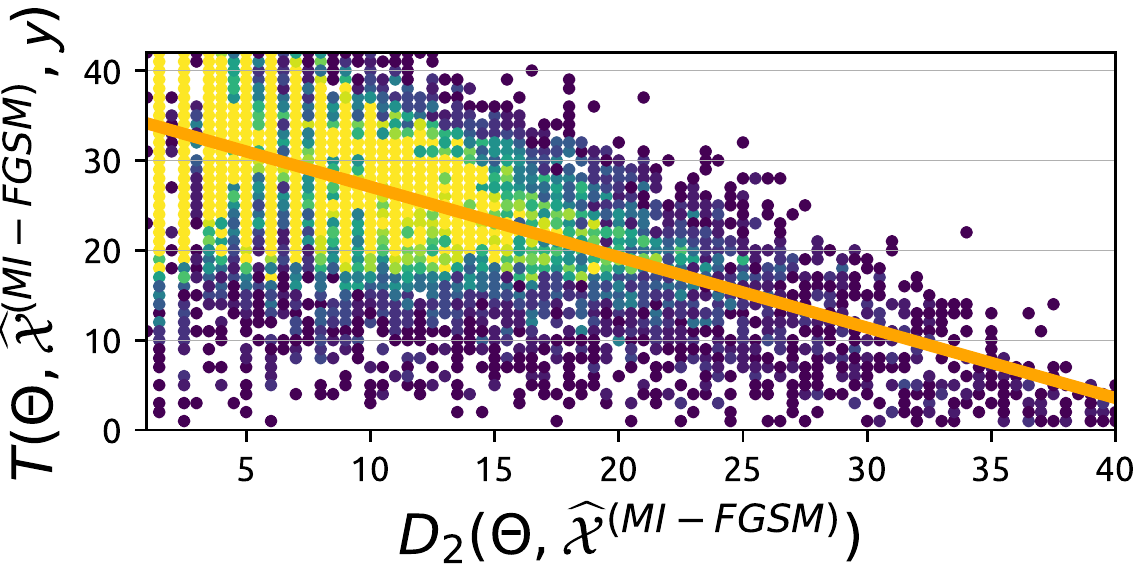}
\includegraphics[width=0.35\textwidth]{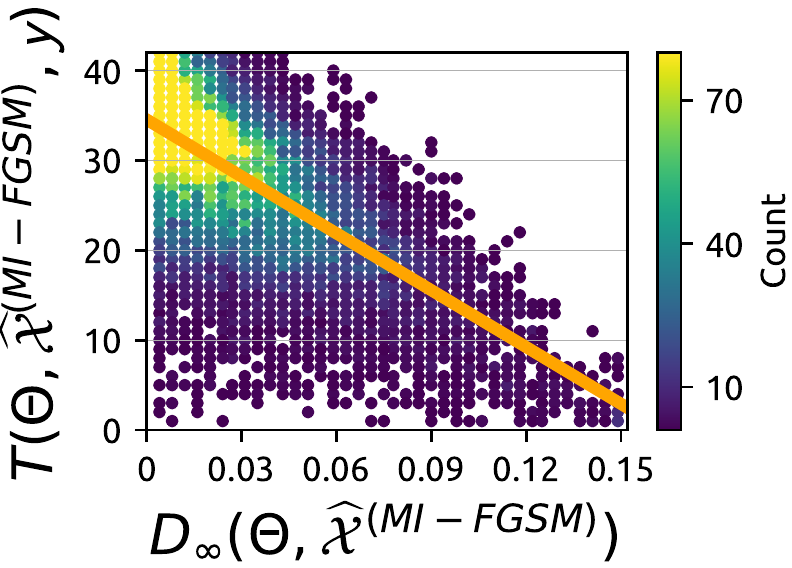}
\vspace{1em}
\caption*{\qquad\,\,\,\,\, Correlation: $-0.64$ \qquad\qquad\,\,\,\,\,\,\,\,\,\,\,\,\,\,\, Correlation: $-0.64$}
\vspace{1em}
\caption{Scatter plot of $D_p(\Theta, \widehat{\mathcal{X}}^{(A)})$, the minimum amount of perturbation required for each source image, against adversarial transferability count $T(\Theta, \widehat{\mathcal{X}}^{\text{(A)}}, \bm{y})$, for $p=2$ (left) and $p=\infty$ (right). The top graph shows the results for all adversarial examples, whereas the following ones present results for individual attacks. The regression line is shown in orange.}
\label{fig:transferability_perturbation}
\end{figure}

\clearpage

\begin{figure*}[hbtp!]
\centering
\rotatebox[origin=l]{90}{\phantom{---}\scriptsize\underline{$T(\cdot)\geq1$}}\hspace{0.5em}
\includegraphics[width=0.4\linewidth]{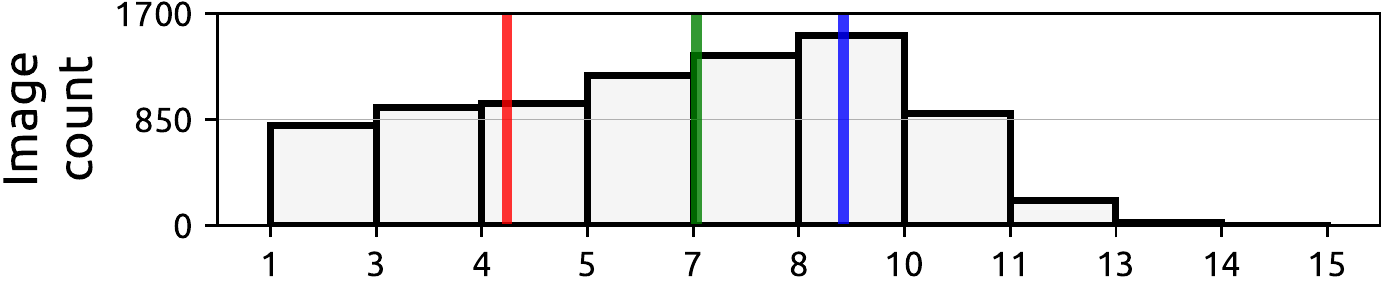}\hspace{1em}
\includegraphics[width=0.4\linewidth]{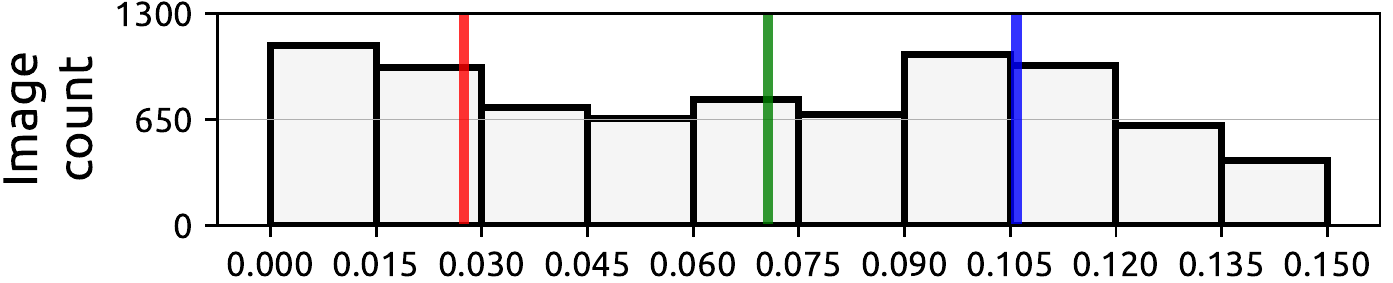}
\rotatebox[origin=l]{180}{\phantom{------}}
\\
\vspace{0.2em}
\rotatebox[origin=l]{90}{\phantom{--}\scriptsize\underline{$T(\cdot)\geq20$}}\hspace{0.5em}
\includegraphics[width=0.4\linewidth]{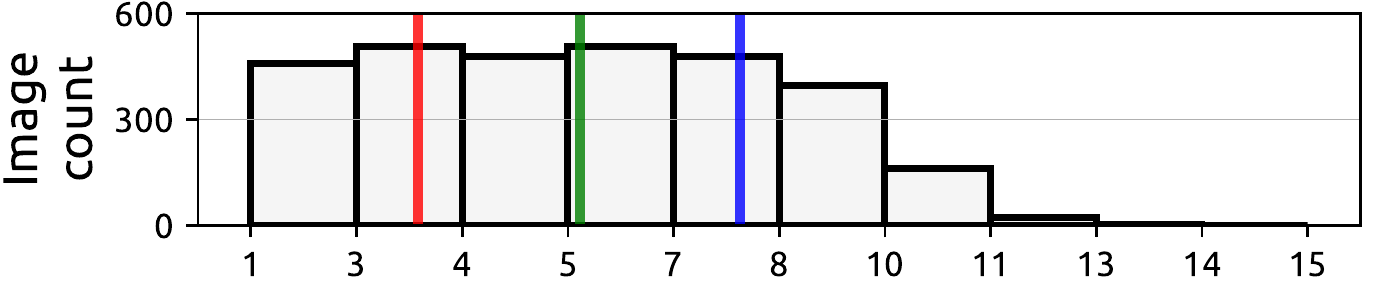}\hspace{1em}
\includegraphics[width=0.4\linewidth]{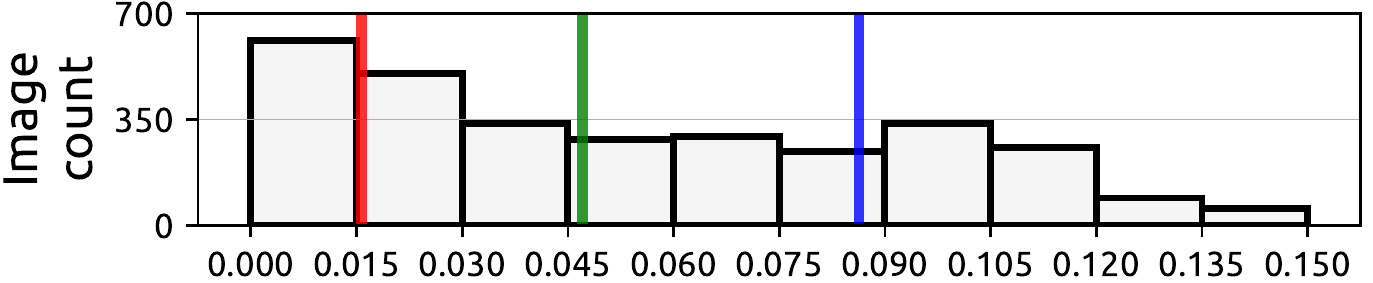}
\rotatebox[origin=l]{180}{\phantom{------}}
\\
\vspace{0.2em}
\rotatebox[origin=l]{90}{\phantom{----}\scriptsize\underline{$T(\cdot)\geq30$}}\hspace{0.5em}
\includegraphics[width=0.4\linewidth]{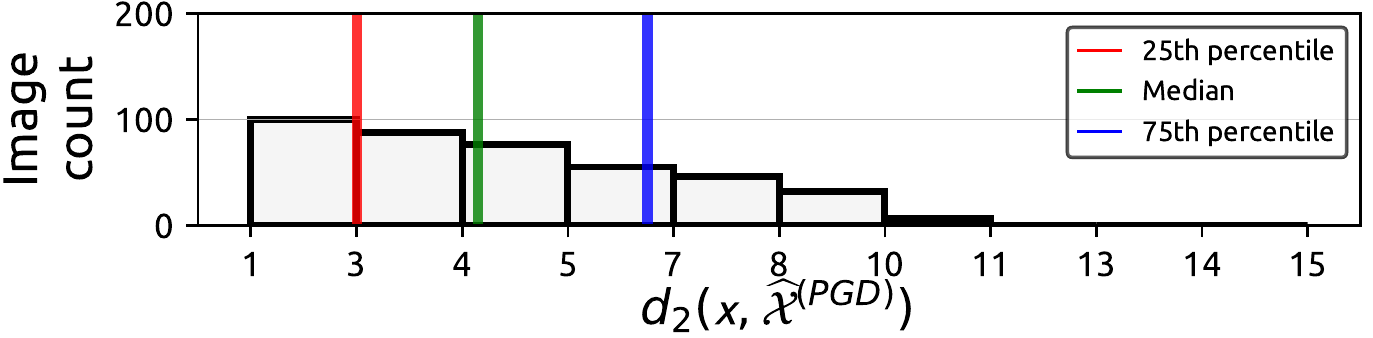}\hspace{1em}
\includegraphics[width=0.4\linewidth]{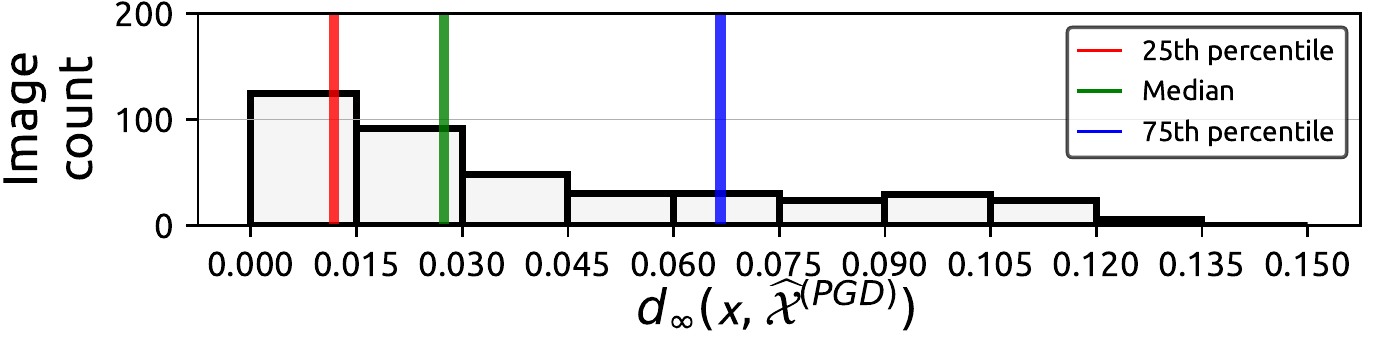}
\rotatebox[origin=l]{180}{\phantom{------}}
\\
\hspace{5.5em}
\\
(a) Adversarial examples transferred to \textbf{AlexNet} with \textbf{PGD}.
\vspace{1em}
\\
\rotatebox[origin=l]{90}{\phantom{---}\scriptsize\underline{$T(\cdot)\geq1$}}\hspace{0.5em}
\includegraphics[width=0.4\linewidth]{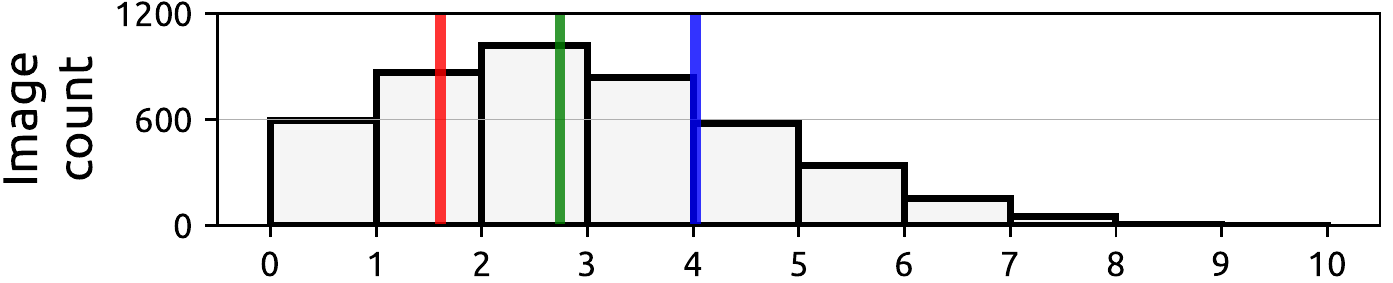}\hspace{1em}
\includegraphics[width=0.4\linewidth]{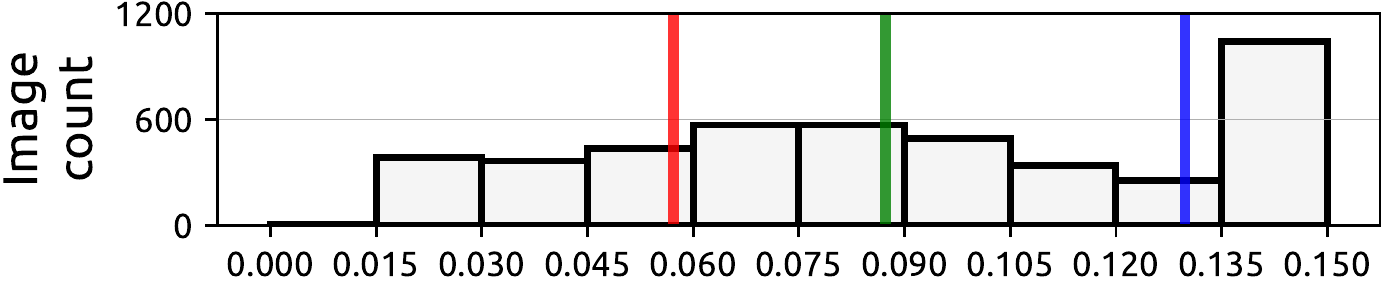}
\rotatebox[origin=l]{180}{\phantom{------}}
\\
\vspace{0.2em}
\rotatebox[origin=l]{90}{\phantom{--}\scriptsize\underline{$T(\cdot)\geq20$}}\hspace{0.5em}
\includegraphics[width=0.4\linewidth]{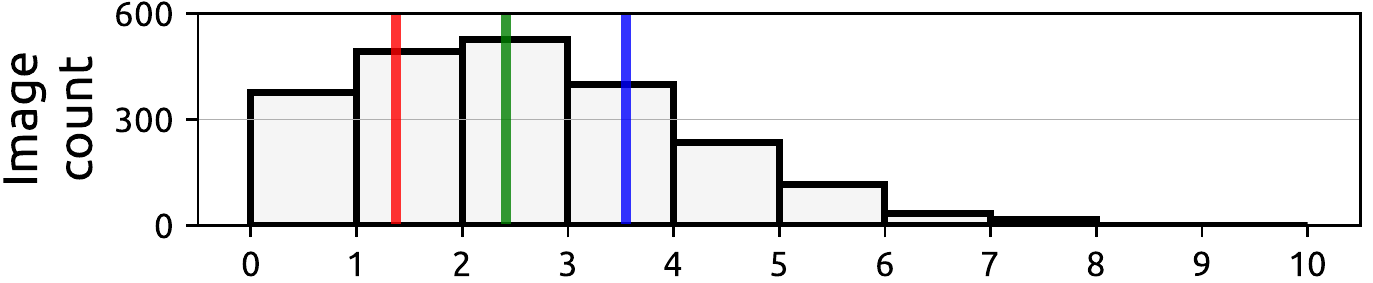}\hspace{1em}
\includegraphics[width=0.4\linewidth]{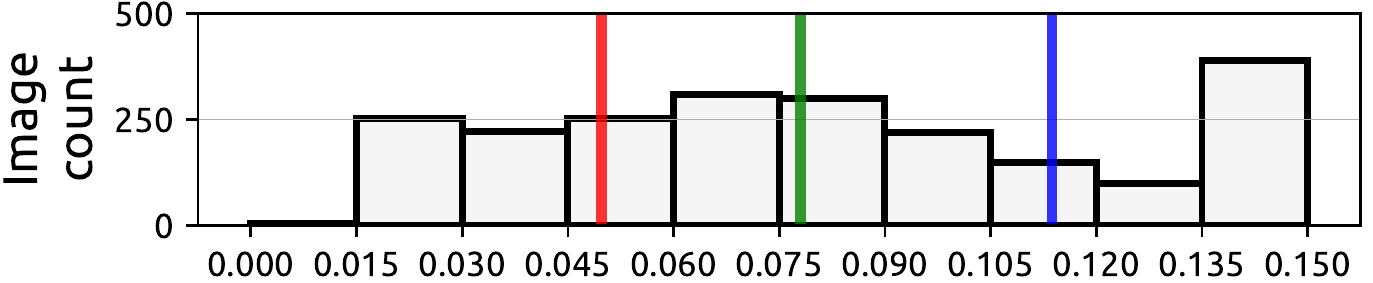}
\rotatebox[origin=l]{180}{\phantom{------}}
\\
\vspace{0.2em}
\rotatebox[origin=l]{90}{\phantom{----}\scriptsize\underline{$T(\cdot)\geq30$}}\hspace{0.5em}
\includegraphics[width=0.4\linewidth]{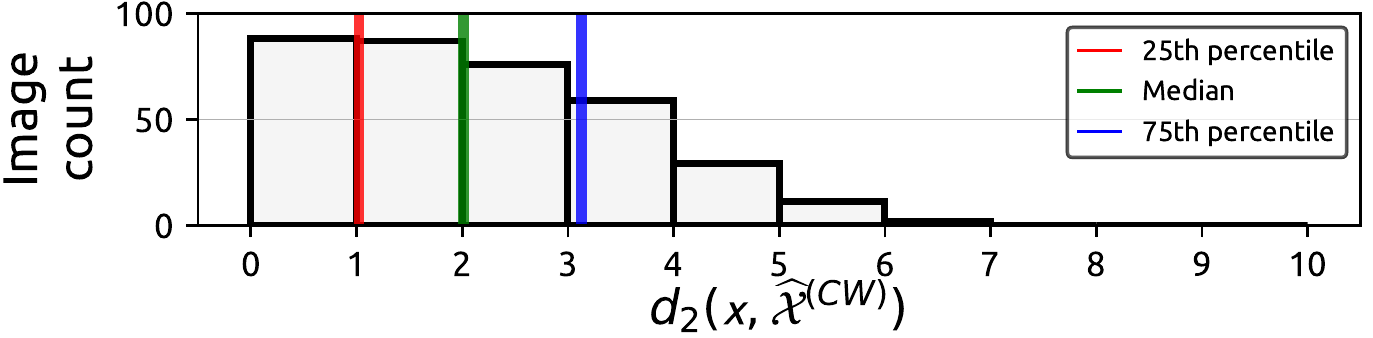}\hspace{1em}
\includegraphics[width=0.4\linewidth]{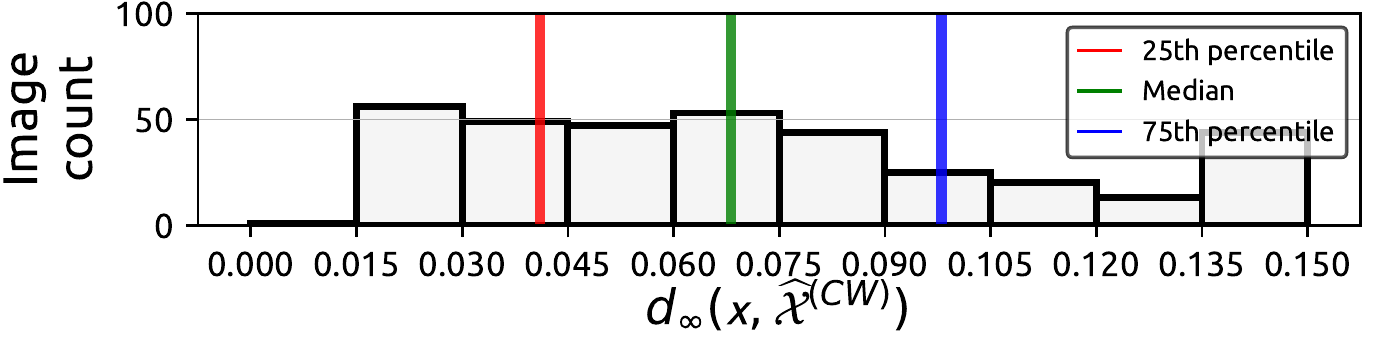}
\rotatebox[origin=l]{180}{\phantom{------}}
\\
\hspace{5.5em}
\\
(b) Adversarial examples transferred to \textbf{AlexNet} with \textbf{CW}.
\vspace{1em}
\\
\rotatebox[origin=l]{90}{\phantom{---}\scriptsize\underline{$T(\cdot)\geq1$}}\hspace{0.5em}
\includegraphics[width=0.4\linewidth]{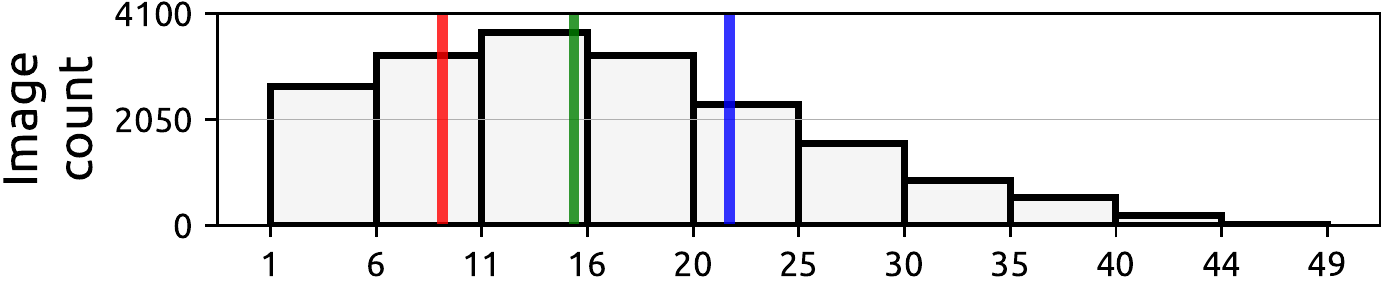}\hspace{1em}
\includegraphics[width=0.4\linewidth]{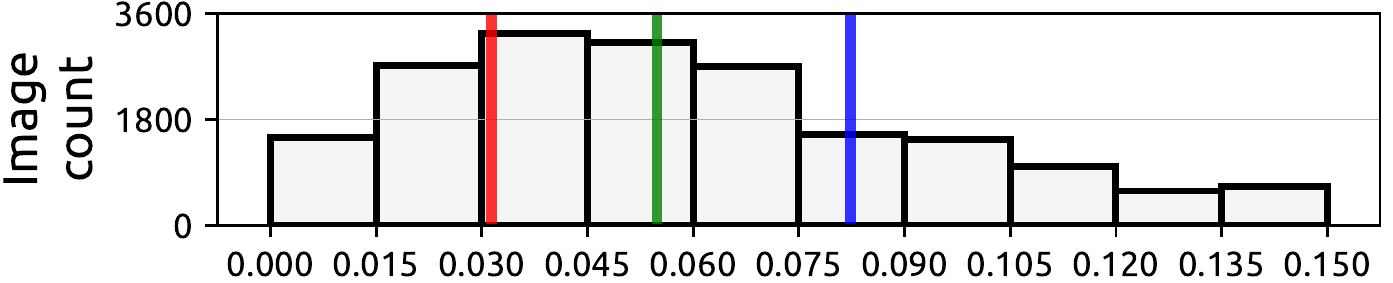}
\rotatebox[origin=l]{180}{\phantom{------}}
\\
\vspace{0.2em}
\rotatebox[origin=l]{90}{\phantom{--}\scriptsize\underline{$T(\cdot)\geq20$}}\hspace{0.5em}
\includegraphics[width=0.4\linewidth]{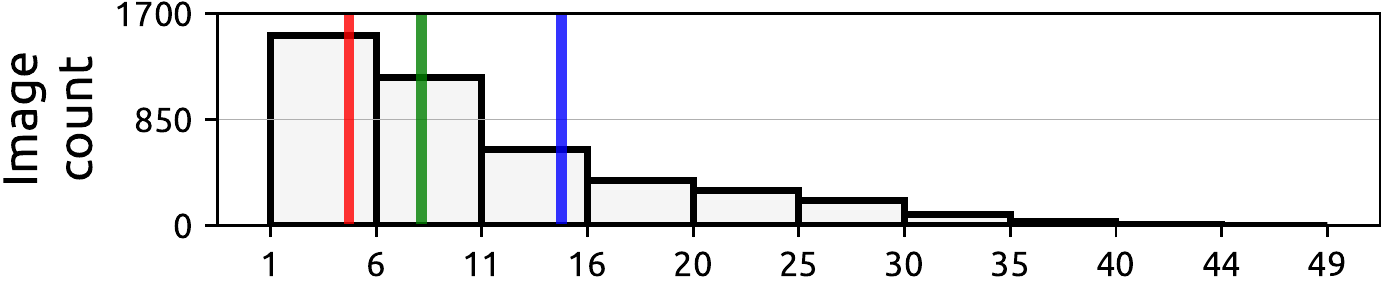}\hspace{1em}
\includegraphics[width=0.4\linewidth]{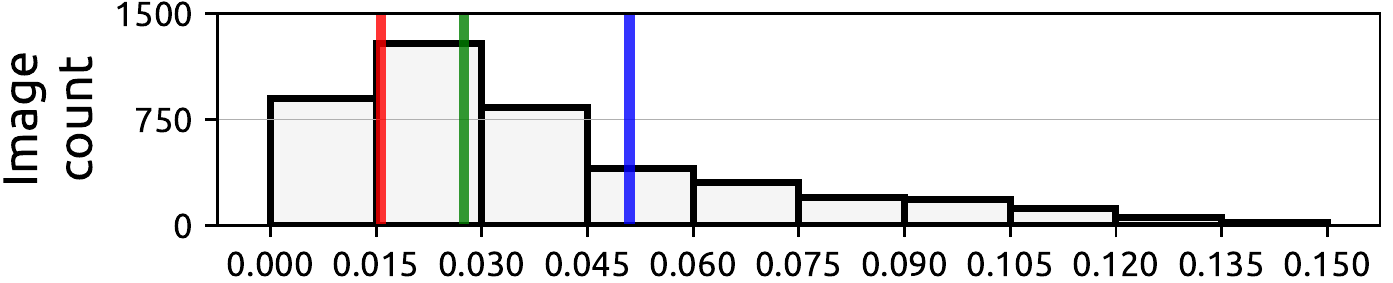}
\rotatebox[origin=l]{180}{\phantom{------}}
\\
\vspace{0.2em}
\rotatebox[origin=l]{90}{\phantom{----}\scriptsize\underline{$T(\cdot)\geq30$}}\hspace{0.5em}
\includegraphics[width=0.4\linewidth]{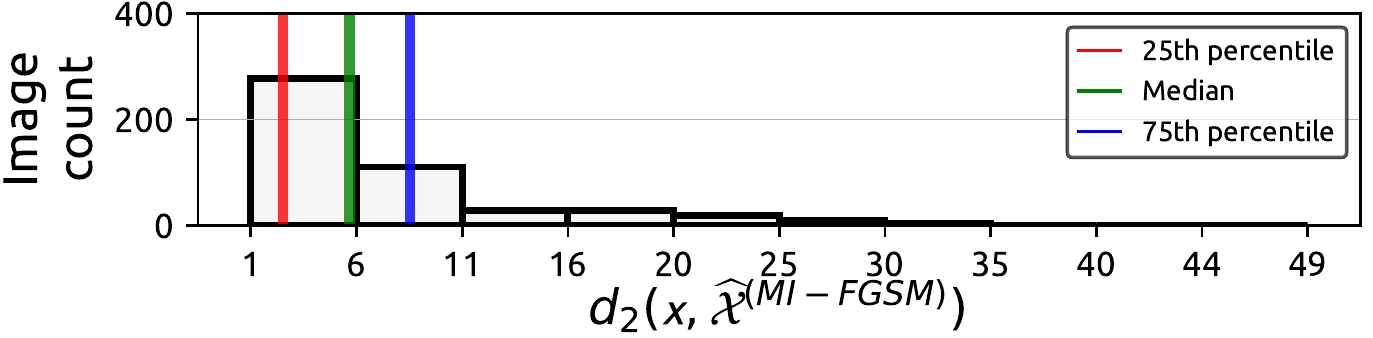}\hspace{1em}
\includegraphics[width=0.4\linewidth]{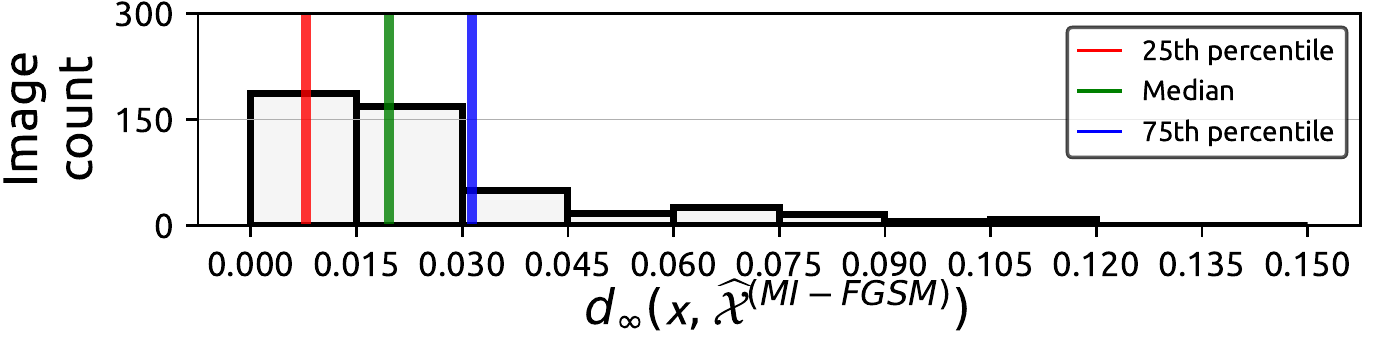}
\rotatebox[origin=l]{180}{\phantom{------}}
\\
\hspace{5.5em}
\\
(c) Adversarial examples transferred to \textbf{AlexNet} with \textbf{MI-FGSM}.
\vspace{2.5em}
\caption{Source images that achieved adversarial transferability to \textbf{AlexNet} are selected based on transferability count, with $T(\Theta, \widehat{\mathcal{X}}^{\text{(A)}}, \bm{y}) \geq \{1, 20, 30\}$. The minimum amount of perturbation required for creating adversarial examples from these source images is histogrammed, measuring the perturbation using $d_p(\bm{x}, \widehat{\mathcal{X}}^{\text{(A)}})$, with $p\in \{2,\infty\}$. The median perturbation, as well as the $25$th and the $75$th percentile, are provided in order to improve interpretability.}
\label{fig:pert_norm_alexnet}
\end{figure*}

\clearpage

\begin{figure*}[hbtp!]
\centering
\rotatebox[origin=l]{90}{\phantom{---}\scriptsize\underline{$T(\cdot)\geq1$}}\hspace{0.5em}
\includegraphics[width=0.4\linewidth]{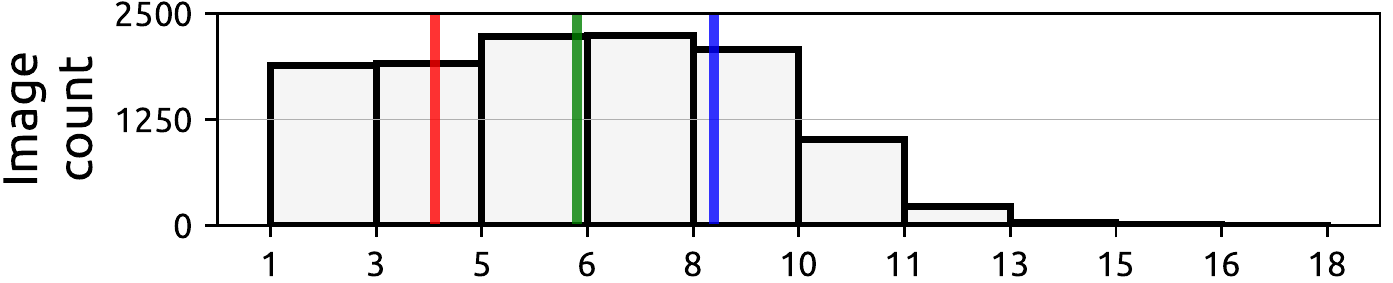}\hspace{1em}
\includegraphics[width=0.4\linewidth]{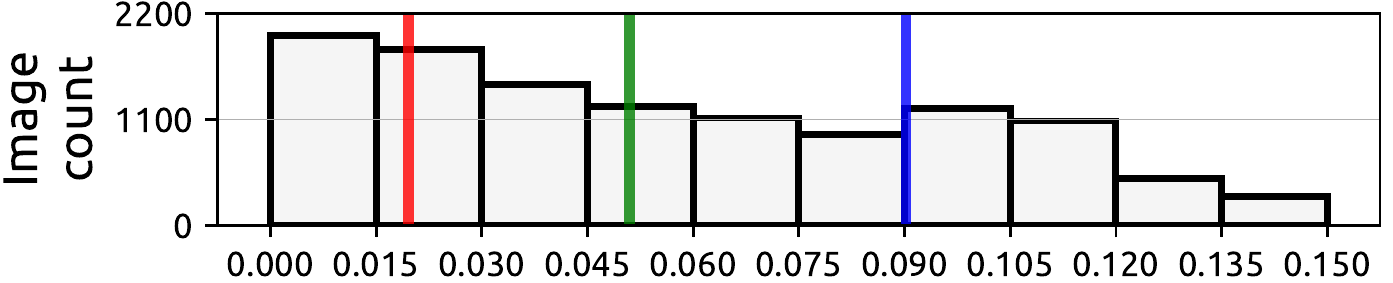}
\rotatebox[origin=l]{180}{\phantom{------}}
\\
\vspace{0.2em}
\rotatebox[origin=l]{90}{\phantom{--}\scriptsize\underline{$T(\cdot)\geq20$}}\hspace{0.5em}
\includegraphics[width=0.4\linewidth]{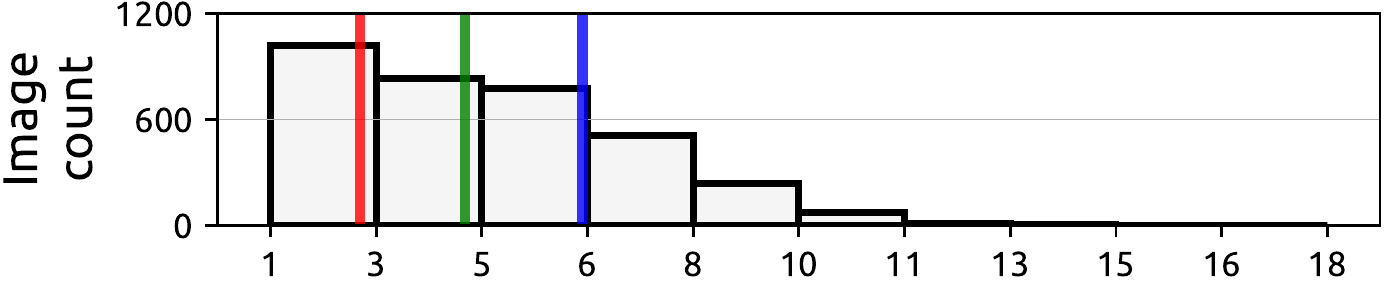}\hspace{1em}
\includegraphics[width=0.4\linewidth]{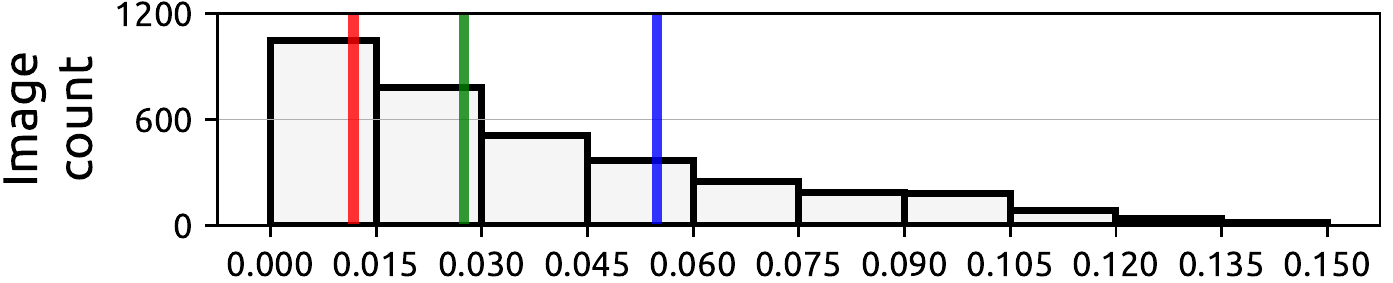}
\rotatebox[origin=l]{180}{\phantom{------}}
\\
\vspace{0.2em}
\rotatebox[origin=l]{90}{\phantom{----}\scriptsize\underline{$T(\cdot)\geq30$}}\hspace{0.5em}
\includegraphics[width=0.4\linewidth]{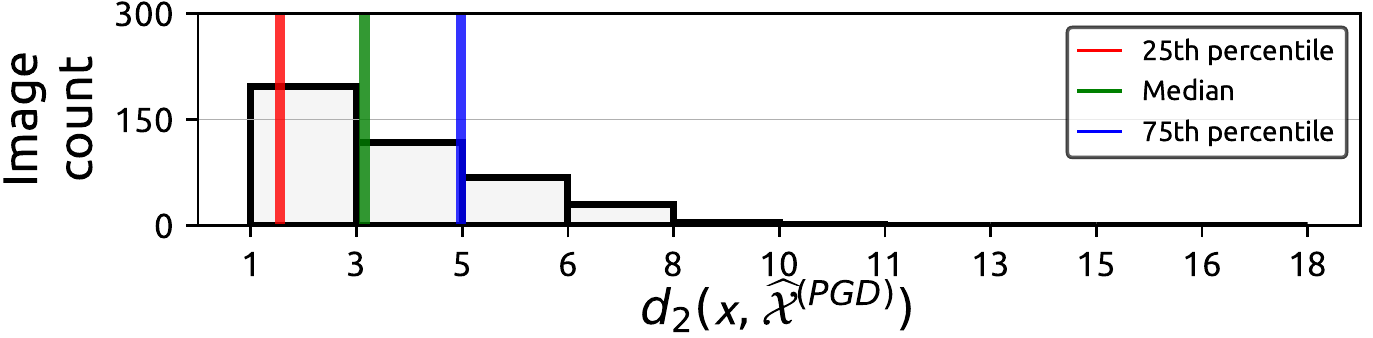}\hspace{1em}
\includegraphics[width=0.4\linewidth]{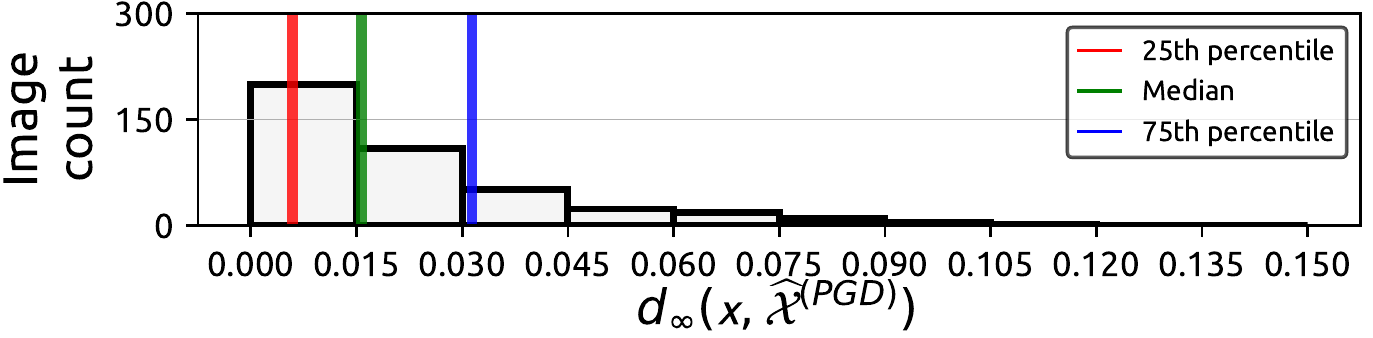}
\rotatebox[origin=l]{180}{\phantom{------}}
\\
\hspace{5.5em}
\\
(a) Adversarial examples transferred to \textbf{SqueezeNet} with \textbf{PGD}.
\vspace{1em}
\\
\rotatebox[origin=l]{90}{\phantom{---}\scriptsize\underline{$T(\cdot)\geq1$}}\hspace{0.5em}
\includegraphics[width=0.4\linewidth]{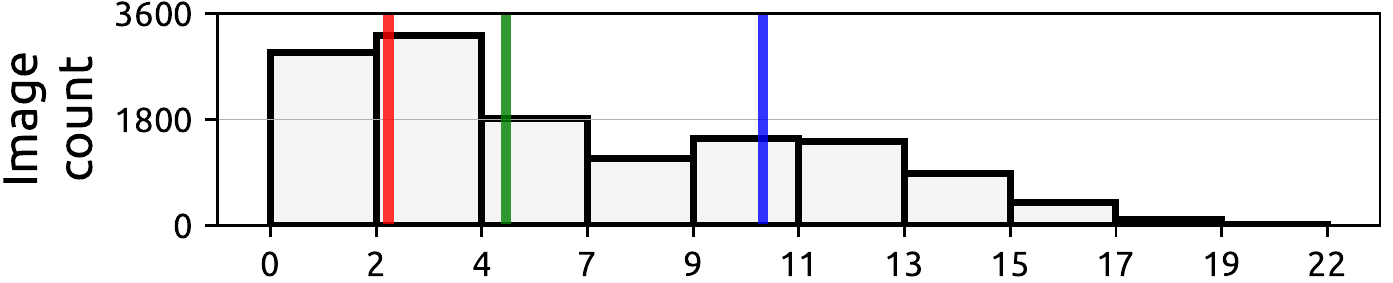}\hspace{1em}
\includegraphics[width=0.4\linewidth]{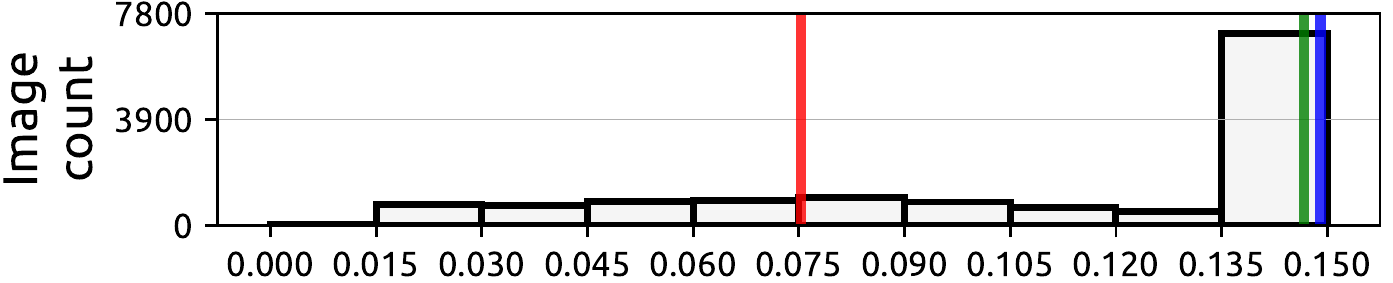}
\rotatebox[origin=l]{180}{\phantom{------}}
\\
\vspace{0.2em}
\rotatebox[origin=l]{90}{\phantom{--}\scriptsize\underline{$T(\cdot)\geq20$}}\hspace{0.5em}
\includegraphics[width=0.4\linewidth]{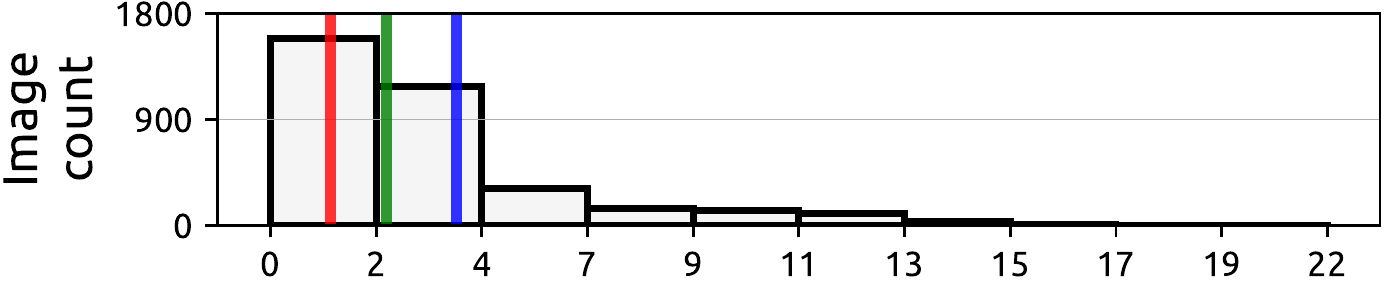}\hspace{1em}
\includegraphics[width=0.4\linewidth]{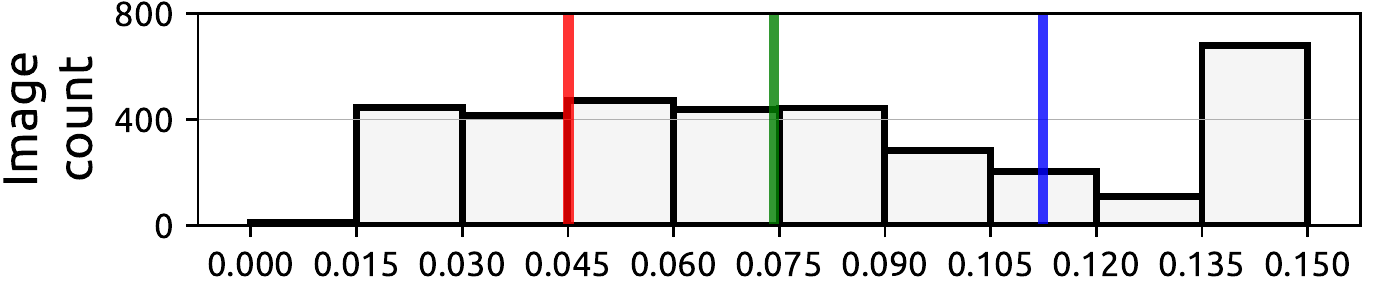}
\rotatebox[origin=l]{180}{\phantom{------}}
\\
\vspace{0.2em}
\rotatebox[origin=l]{90}{\phantom{----}\scriptsize\underline{$T(\cdot)\geq30$}}\hspace{0.5em}
\includegraphics[width=0.4\linewidth]{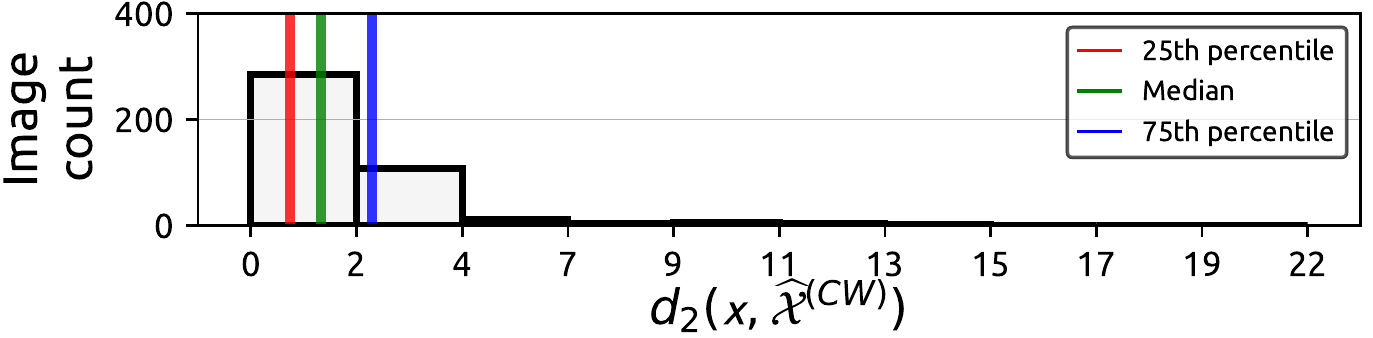}\hspace{1em}
\includegraphics[width=0.4\linewidth]{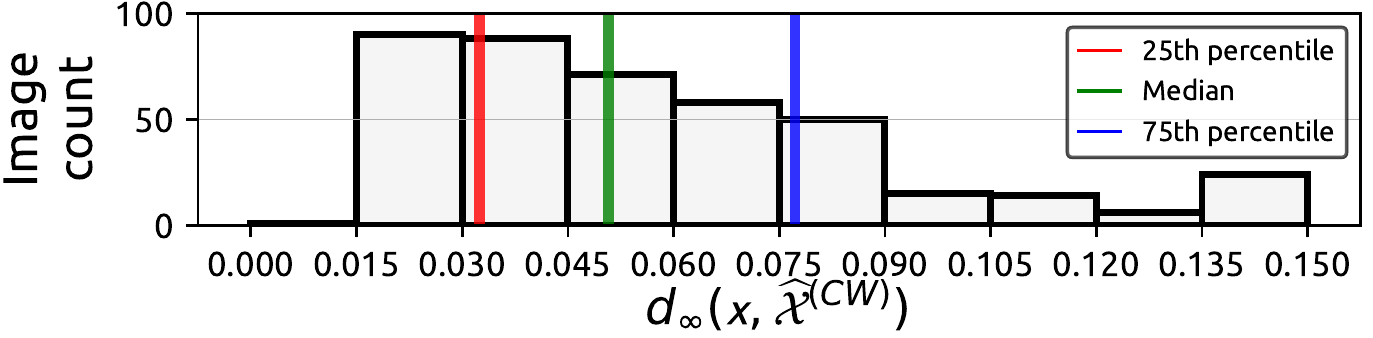}
\rotatebox[origin=l]{180}{\phantom{------}}
\\
\hspace{5.5em}
\\
(b) Adversarial examples transferred to \textbf{SqueezeNet} with \textbf{CW}.
\vspace{1em}
\\
\rotatebox[origin=l]{90}{\phantom{---}\scriptsize\underline{$T(\cdot)\geq1$}}\hspace{0.5em}
\includegraphics[width=0.4\linewidth]{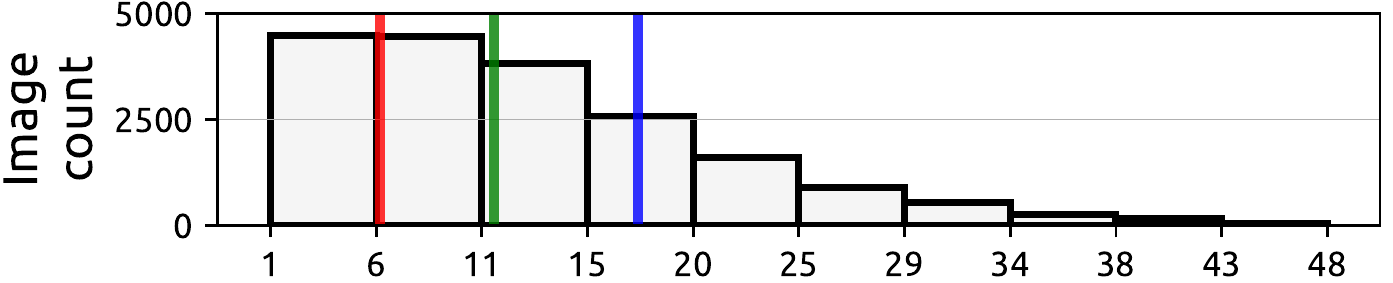}\hspace{1em}
\includegraphics[width=0.4\linewidth]{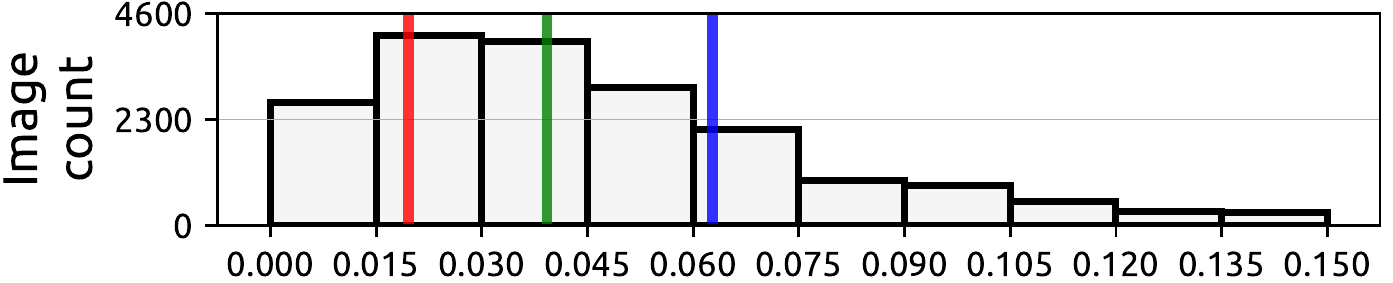}
\rotatebox[origin=l]{180}{\phantom{------}}
\\
\vspace{0.2em}
\rotatebox[origin=l]{90}{\phantom{--}\scriptsize\underline{$T(\cdot)\geq20$}}\hspace{0.5em}
\includegraphics[width=0.4\linewidth]{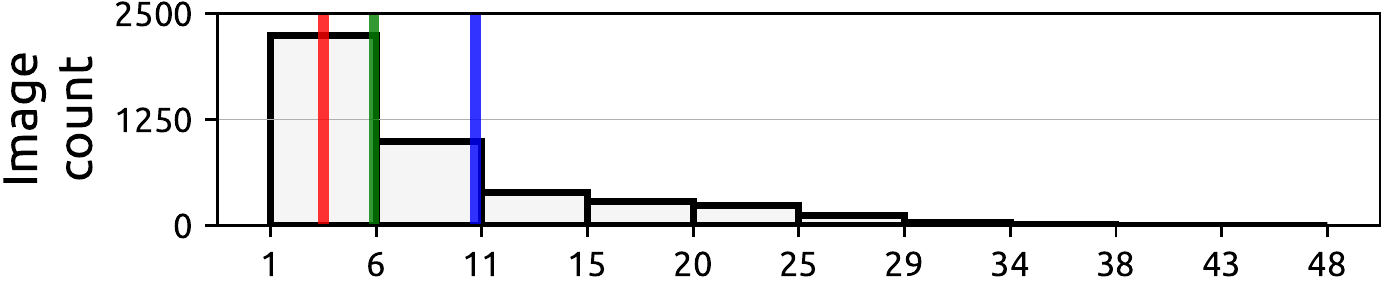}\hspace{1em}
\includegraphics[width=0.4\linewidth]{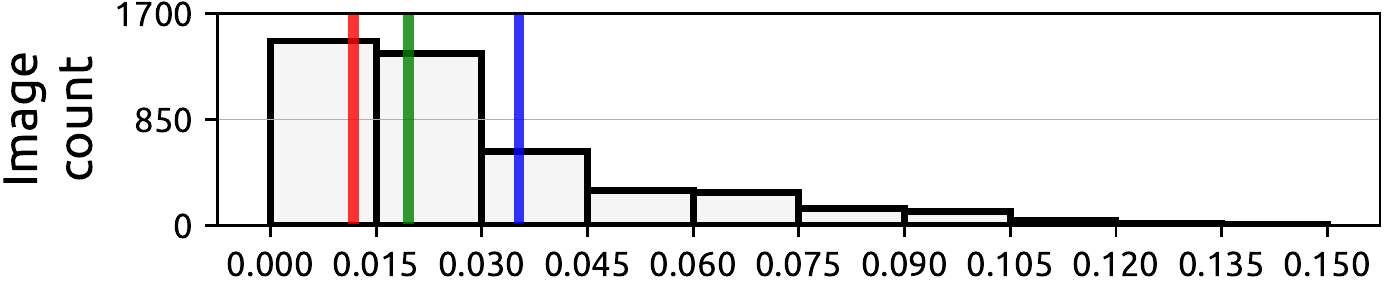}
\rotatebox[origin=l]{180}{\phantom{------}}
\\
\vspace{0.2em}
\rotatebox[origin=l]{90}{\phantom{----}\scriptsize\underline{$T(\cdot)\geq30$}}\hspace{0.5em}
\includegraphics[width=0.4\linewidth]{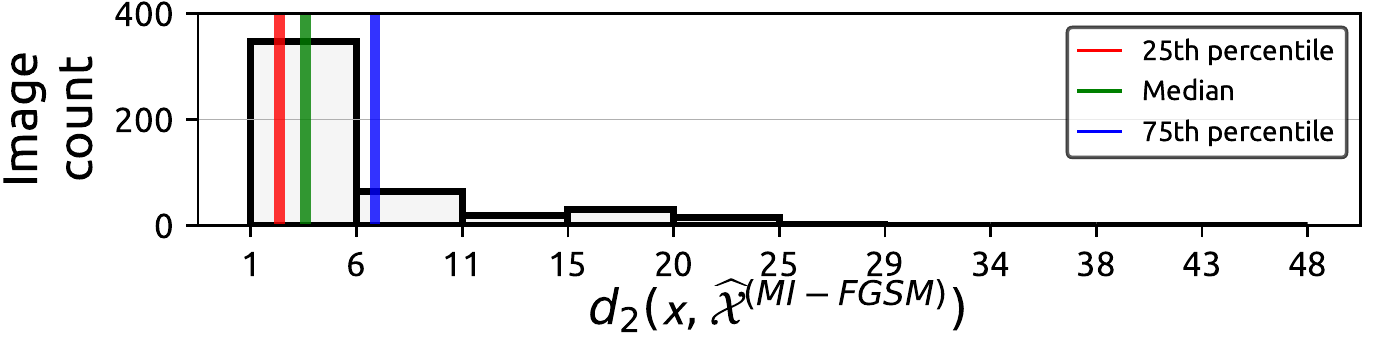}\hspace{1em}
\includegraphics[width=0.4\linewidth]{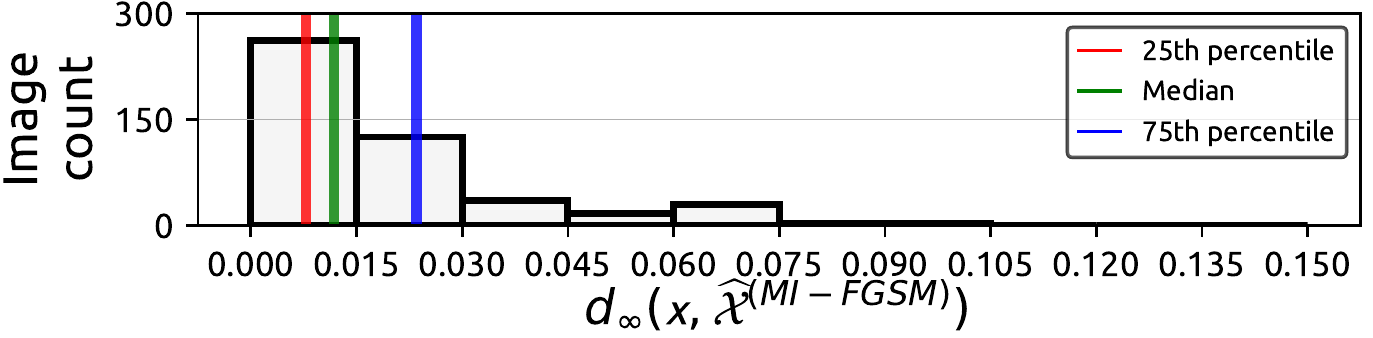}
\rotatebox[origin=l]{180}{\phantom{------}}
\\
\hspace{5.5em}
\\
(c) Adversarial examples transferred to \textbf{SqueezeNet} with \textbf{MI-FGSM}.
\vspace{2.5em}
\caption{Source images that achieved adversarial transferability to \textbf{SqueezeNet} are selected based on transferability count, with $T(\Theta, \widehat{\mathcal{X}}^{\text{(A)}}, \bm{y}) \geq \{1, 20, 30\}$. The minimum amount of perturbation required for creating adversarial examples from these source images is histogrammed, measuring the perturbation using $d_p(\bm{x}, \widehat{\mathcal{X}}^{\text{(A)}})$, with $p\in \{2,\infty\}$. The median perturbation, as well as the $25$th and the $75$th percentile, are provided in order to improve interpretability.}
\label{fig:pert_norm_squeezenet}
\end{figure*}

\clearpage

\begin{figure*}[hbtp!]
\centering
\rotatebox[origin=l]{90}{\phantom{---}\scriptsize\underline{$T(\cdot)\geq1$}}\hspace{0.5em}
\includegraphics[width=0.4\linewidth]{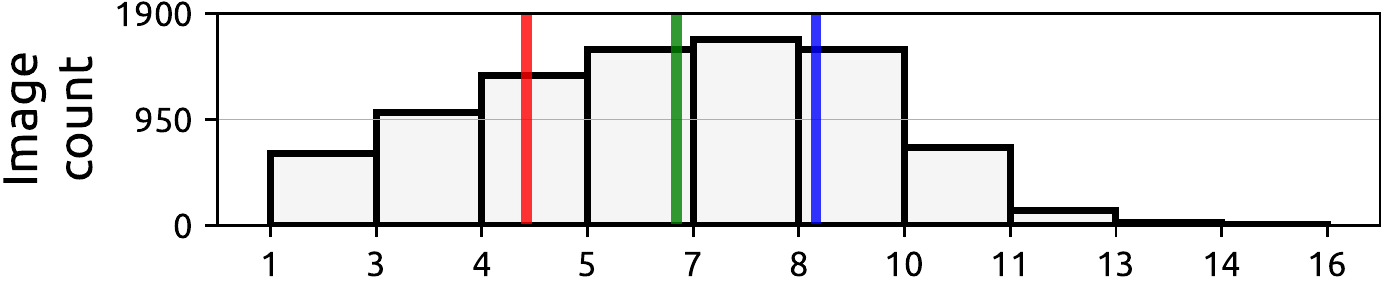}\hspace{1em}
\includegraphics[width=0.4\linewidth]{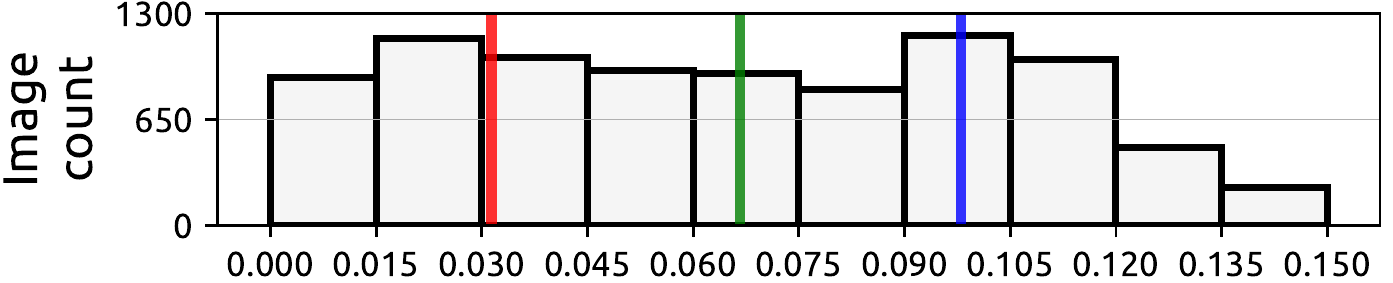}
\rotatebox[origin=l]{180}{\phantom{------}}
\\
\vspace{0.2em}
\rotatebox[origin=l]{90}{\phantom{--}\scriptsize\underline{$T(\cdot)\geq20$}}\hspace{0.5em}
\includegraphics[width=0.4\linewidth]{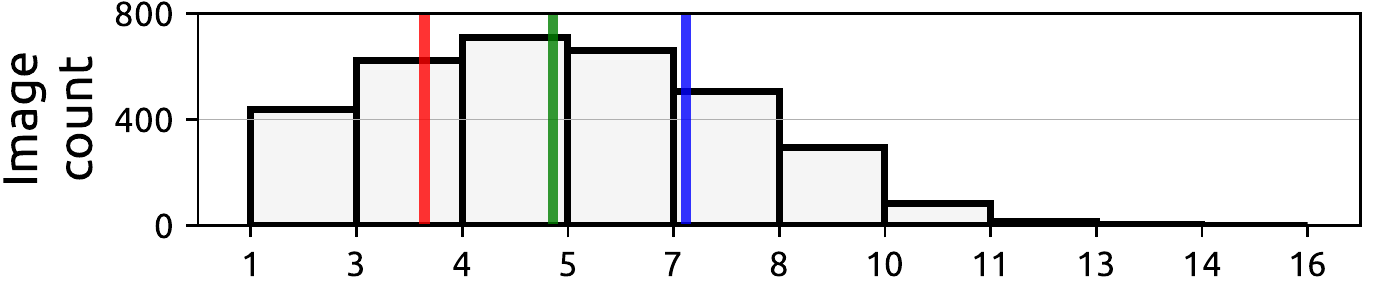}\hspace{1em}
\includegraphics[width=0.4\linewidth]{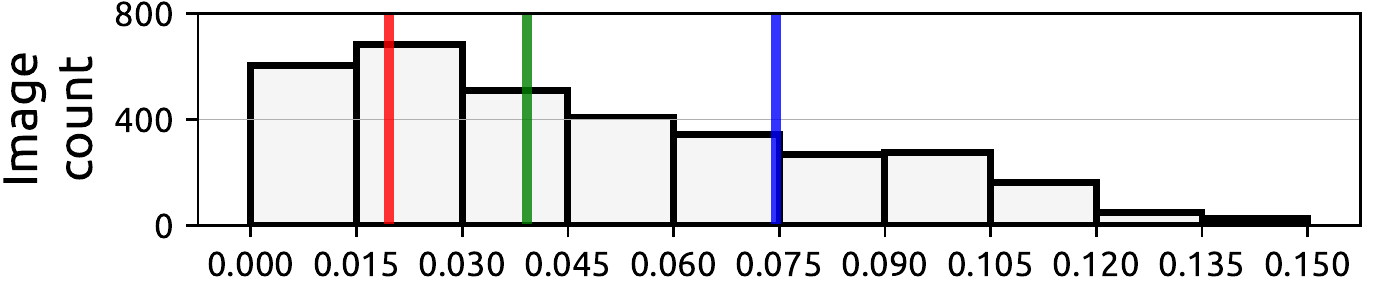}
\rotatebox[origin=l]{180}{\phantom{------}}
\\
\vspace{0.2em}
\rotatebox[origin=l]{90}{\phantom{----}\scriptsize\underline{$T(\cdot)\geq30$}}\hspace{0.5em}
\includegraphics[width=0.4\linewidth]{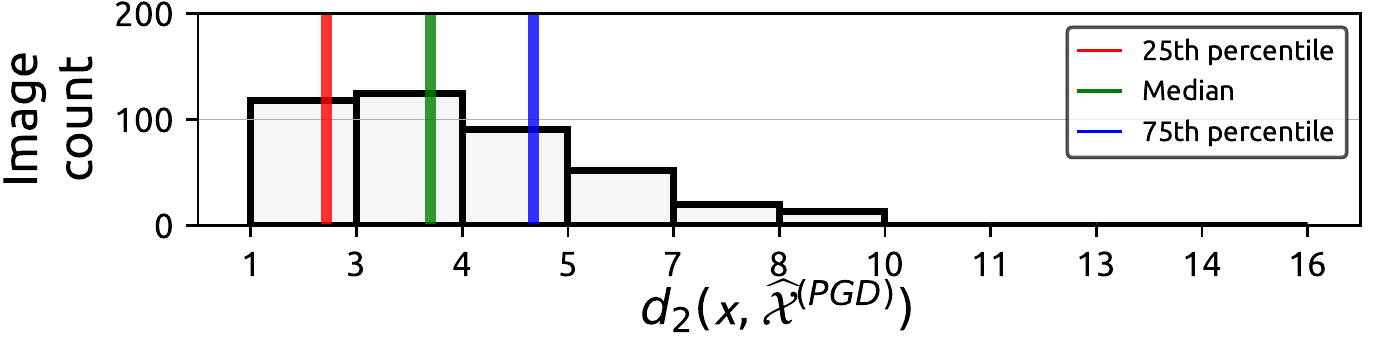}\hspace{1em}
\includegraphics[width=0.4\linewidth]{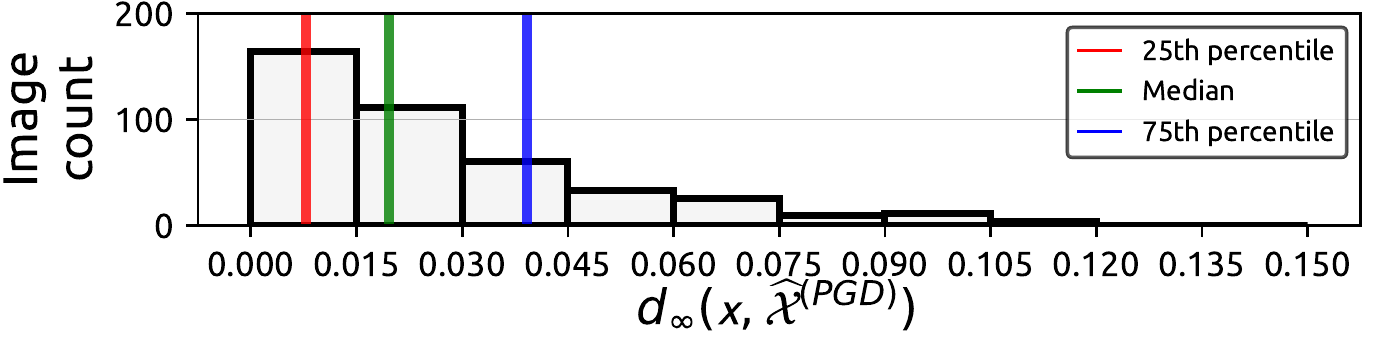}
\rotatebox[origin=l]{180}{\phantom{------}}
\\
\hspace{5.5em}
\\
(a) Adversarial examples transferred to \textbf{VGG-16} with \textbf{PGD}.
\vspace{1em}
\\
\rotatebox[origin=l]{90}{\phantom{---}\scriptsize\underline{$T(\cdot)\geq1$}}\hspace{0.5em}
\includegraphics[width=0.4\linewidth]{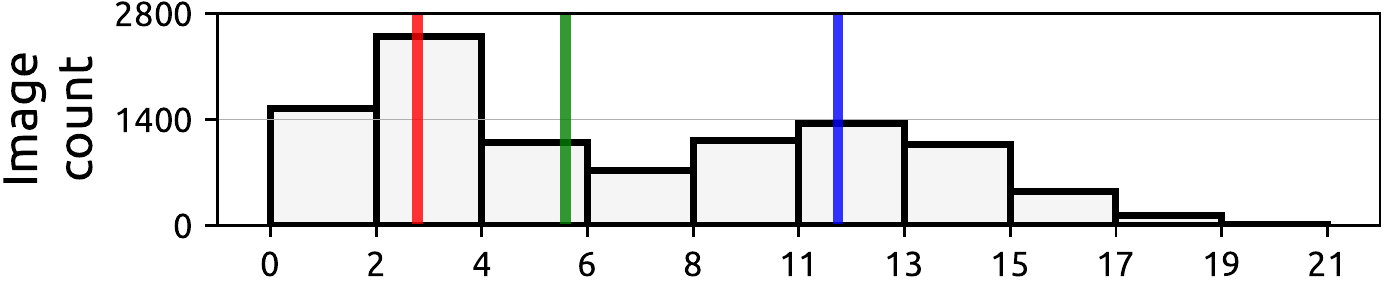}\hspace{1em}
\includegraphics[width=0.4\linewidth]{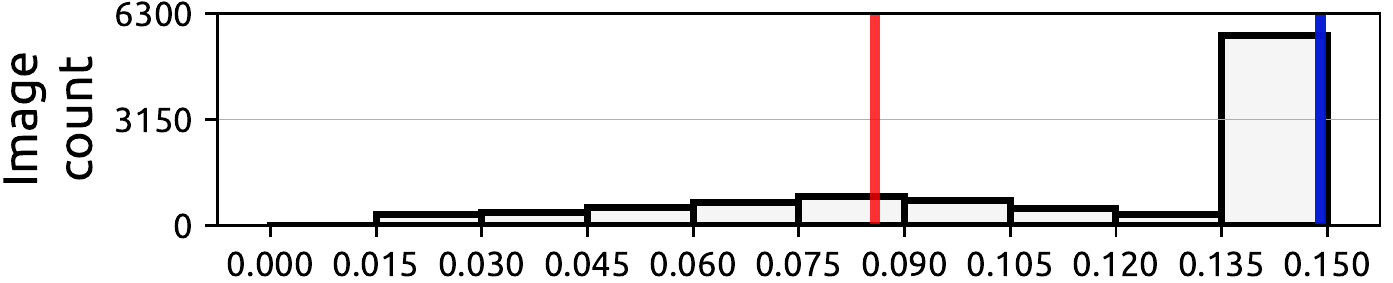}
\rotatebox[origin=l]{180}{\phantom{------}}
\\
\vspace{0.2em}
\rotatebox[origin=l]{90}{\phantom{--}\scriptsize\underline{$T(\cdot)\geq20$}}\hspace{0.5em}
\includegraphics[width=0.4\linewidth]{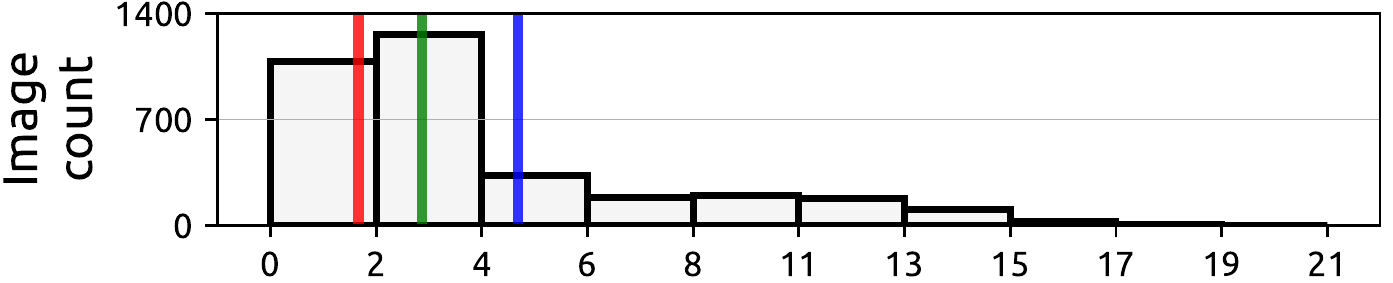}\hspace{1em}
\includegraphics[width=0.4\linewidth]{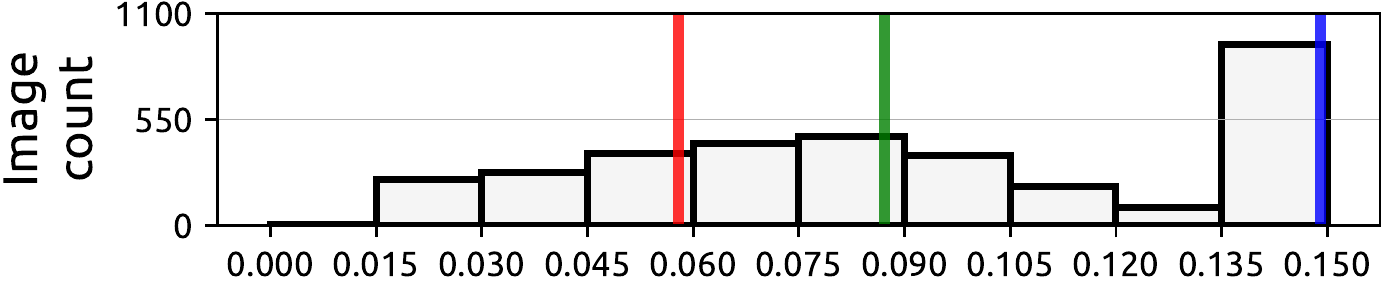}
\rotatebox[origin=l]{180}{\phantom{------}}
\\
\vspace{0.2em}
\rotatebox[origin=l]{90}{\phantom{----}\scriptsize\underline{$T(\cdot)\geq30$}}\hspace{0.5em}
\includegraphics[width=0.4\linewidth]{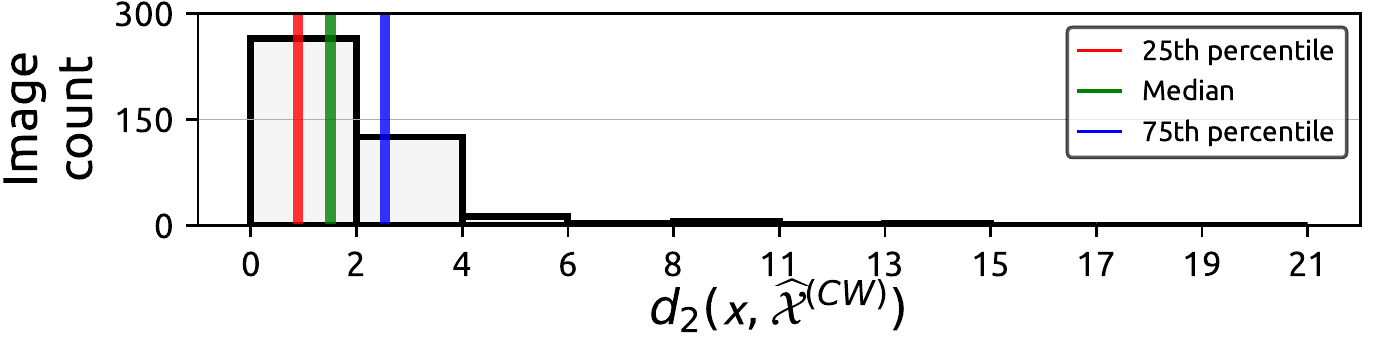}\hspace{1em}
\includegraphics[width=0.4\linewidth]{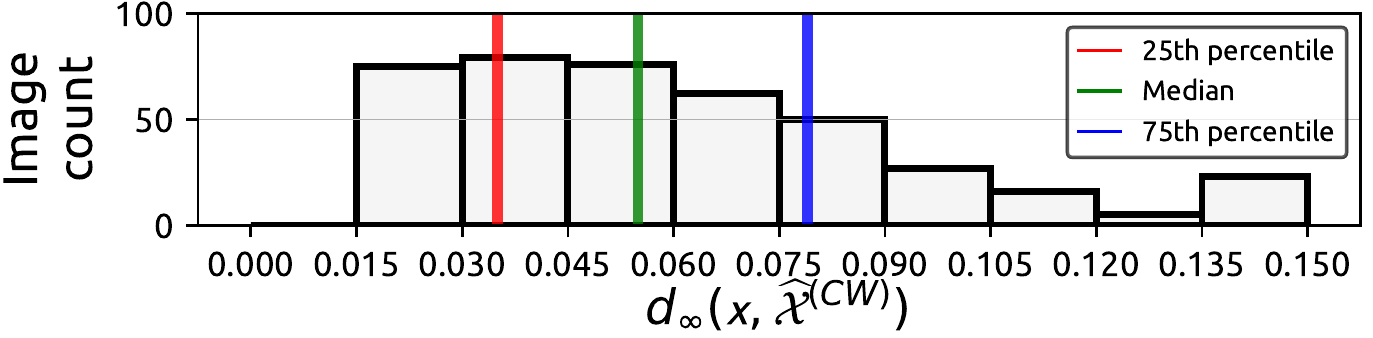}
\rotatebox[origin=l]{180}{\phantom{------}}
\\
\hspace{5.5em}
\\
(b) Adversarial examples transferred to \textbf{VGG-16} with \textbf{CW}.
\vspace{1em}
\\
\rotatebox[origin=l]{90}{\phantom{---}\scriptsize\underline{$T(\cdot)\geq1$}}\hspace{0.5em}
\includegraphics[width=0.4\linewidth]{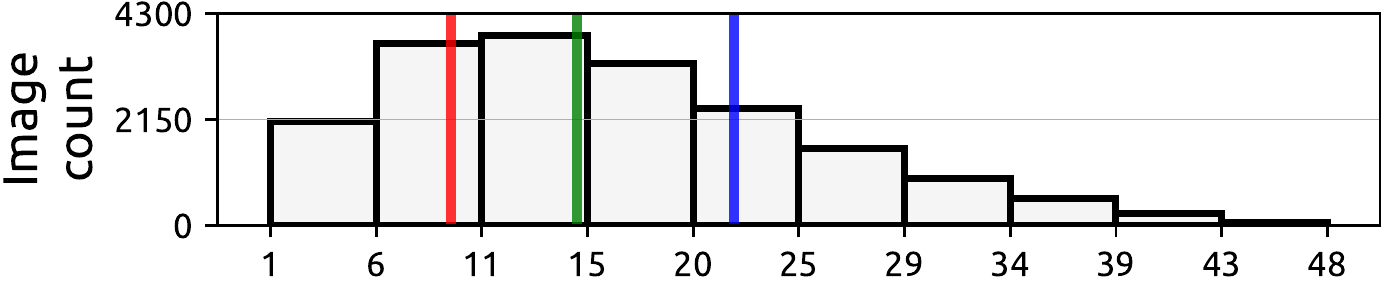}\hspace{1em}
\includegraphics[width=0.4\linewidth]{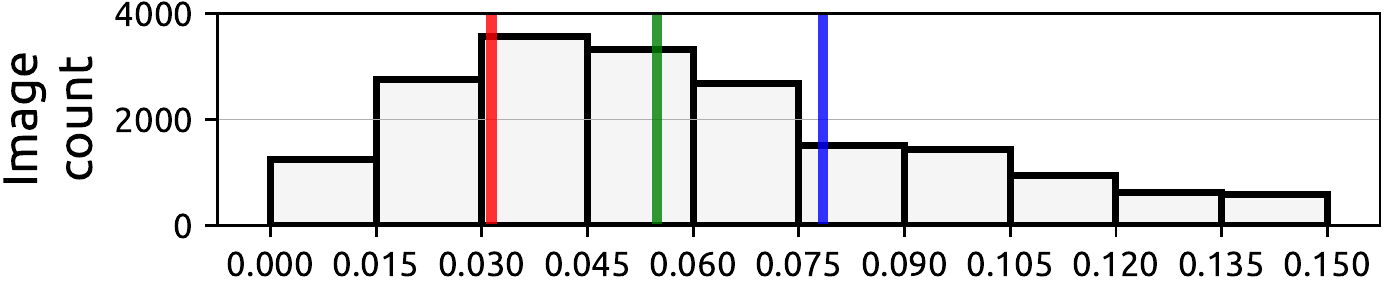}
\rotatebox[origin=l]{180}{\phantom{------}}
\\
\vspace{0.2em}
\rotatebox[origin=l]{90}{\phantom{--}\scriptsize\underline{$T(\cdot)\geq20$}}\hspace{0.5em}
\includegraphics[width=0.4\linewidth]{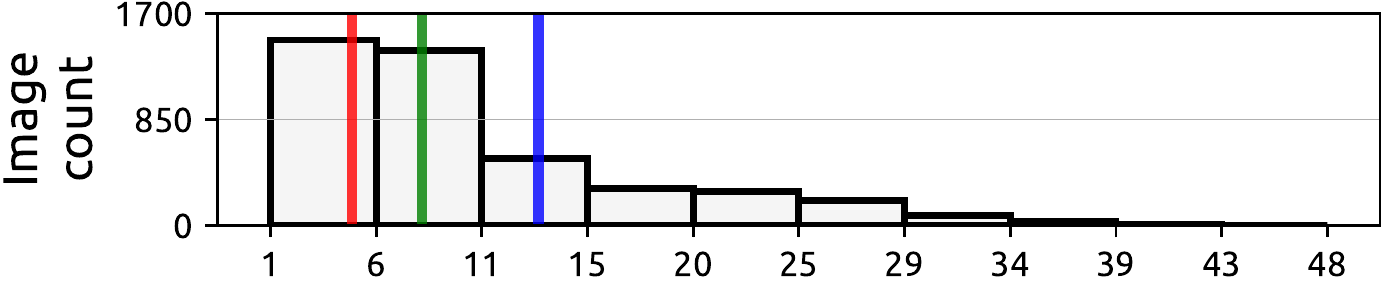}\hspace{1em}
\includegraphics[width=0.4\linewidth]{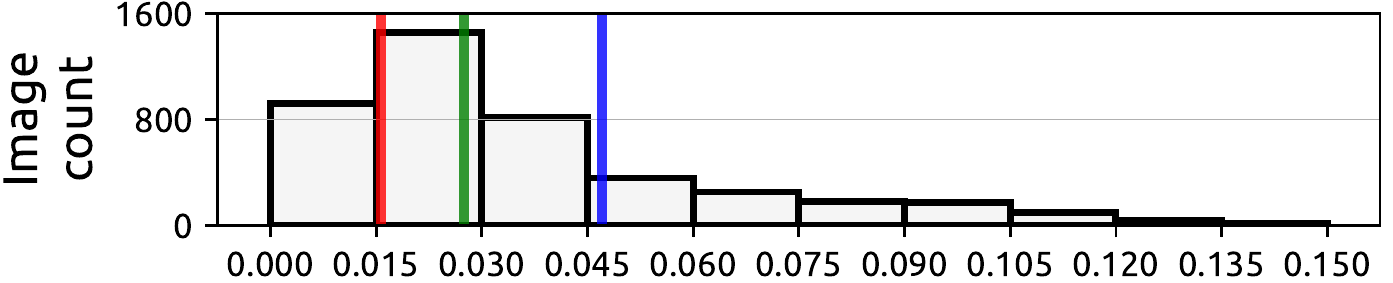}
\rotatebox[origin=l]{180}{\phantom{------}}
\\
\vspace{0.2em}
\rotatebox[origin=l]{90}{\phantom{----}\scriptsize\underline{$T(\cdot)\geq30$}}\hspace{0.5em}
\includegraphics[width=0.4\linewidth]{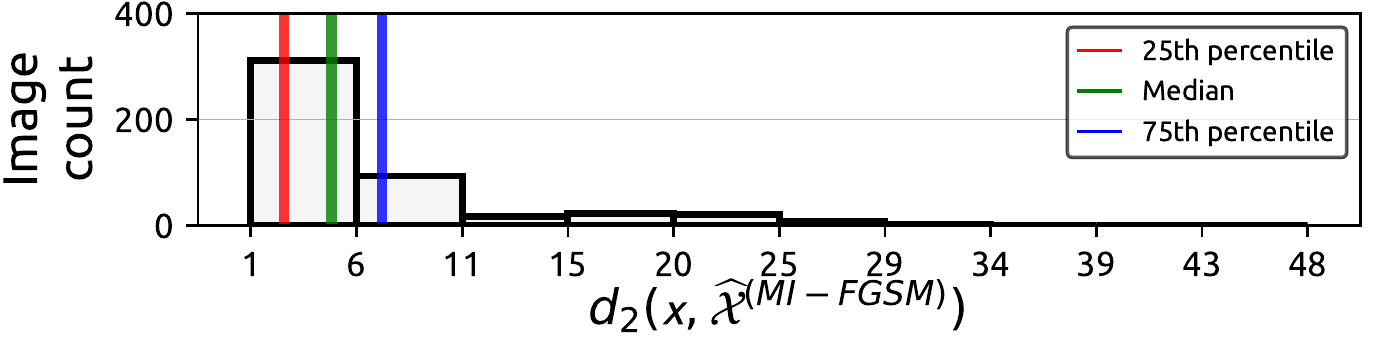}\hspace{1em}
\includegraphics[width=0.4\linewidth]{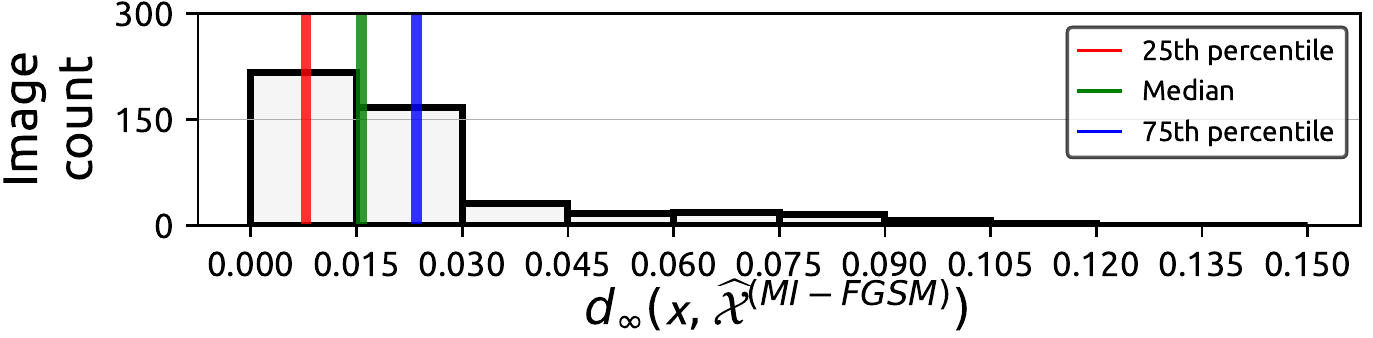}
\rotatebox[origin=l]{180}{\phantom{------}}
\\
\hspace{5.5em}
\\
(c) Adversarial examples transferred to \textbf{VGG-16} with \textbf{MI-FGSM}.
\vspace{2.5em}
\caption{Source images that achieved adversarial transferability to \textbf{VGG-16} are selected based on transferability count, with $T(\Theta, \widehat{\mathcal{X}}^{\text{(A)}}, \bm{y}) \geq \{1, 20, 30\}$. The minimum amount of perturbation required for creating adversarial examples from these source images is histogrammed, measuring the perturbation using $d_p(\bm{x}, \widehat{\mathcal{X}}^{\text{(A)}})$, with $p\in \{2,\infty\}$. The median perturbation, as well as the $25$th and the $75$th percentile, are provided in order to improve interpretability.}
\label{fig:pert_norm_vgg16}
\end{figure*}

\clearpage

\begin{figure*}[hbtp!]
\centering
\rotatebox[origin=l]{90}{\phantom{---}\scriptsize\underline{$T(\cdot)\geq1$}}\hspace{0.5em}
\includegraphics[width=0.4\linewidth]{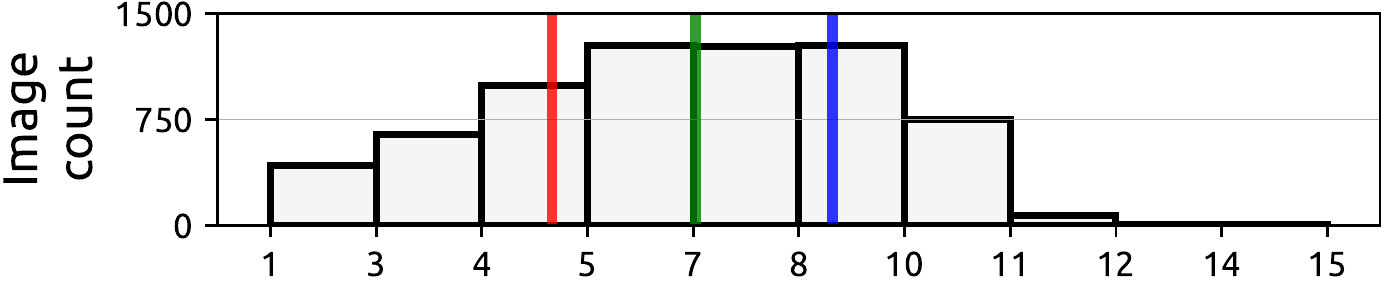}\hspace{1em}
\includegraphics[width=0.4\linewidth]{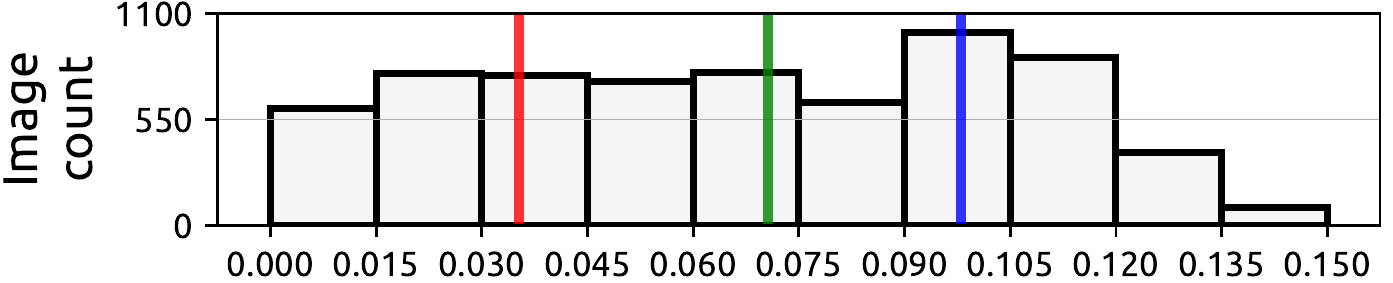}
\rotatebox[origin=l]{180}{\phantom{------}}
\\
\vspace{0.2em}
\rotatebox[origin=l]{90}{\phantom{--}\scriptsize\underline{$T(\cdot)\geq20$}}\hspace{0.5em}
\includegraphics[width=0.4\linewidth]{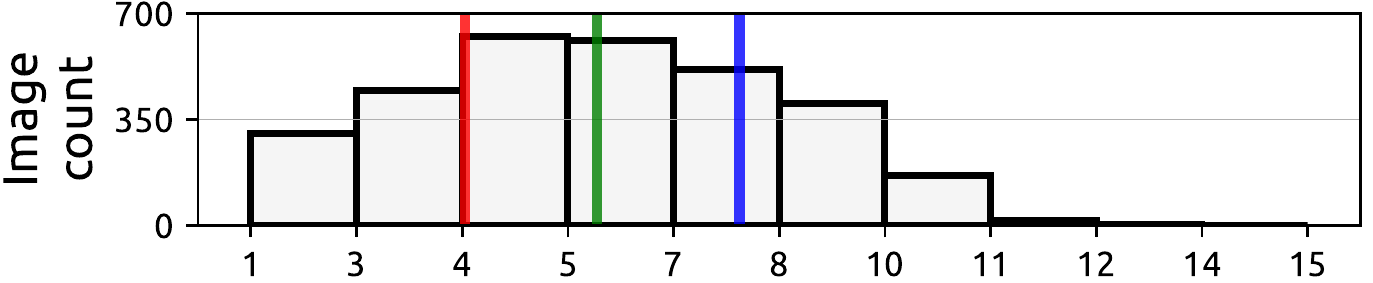}\hspace{1em}
\includegraphics[width=0.4\linewidth]{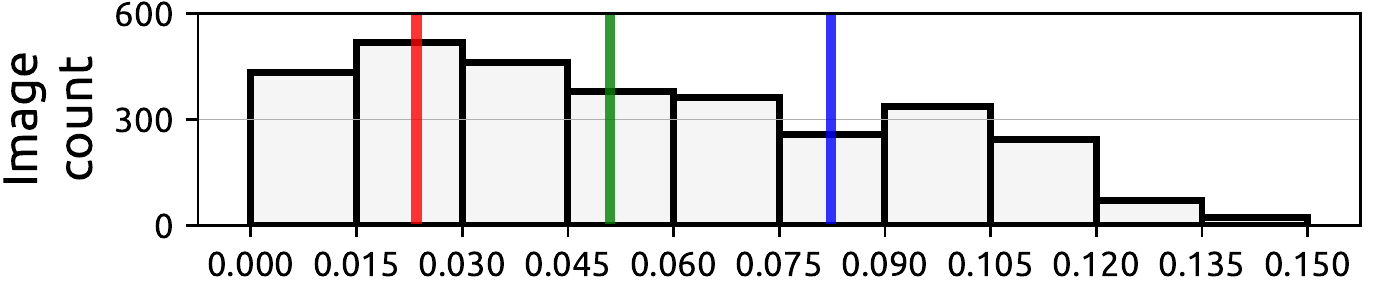}
\rotatebox[origin=l]{180}{\phantom{------}}
\\
\vspace{0.2em}
\rotatebox[origin=l]{90}{\phantom{----}\scriptsize\underline{$T(\cdot)\geq30$}}\hspace{0.5em}
\includegraphics[width=0.4\linewidth]{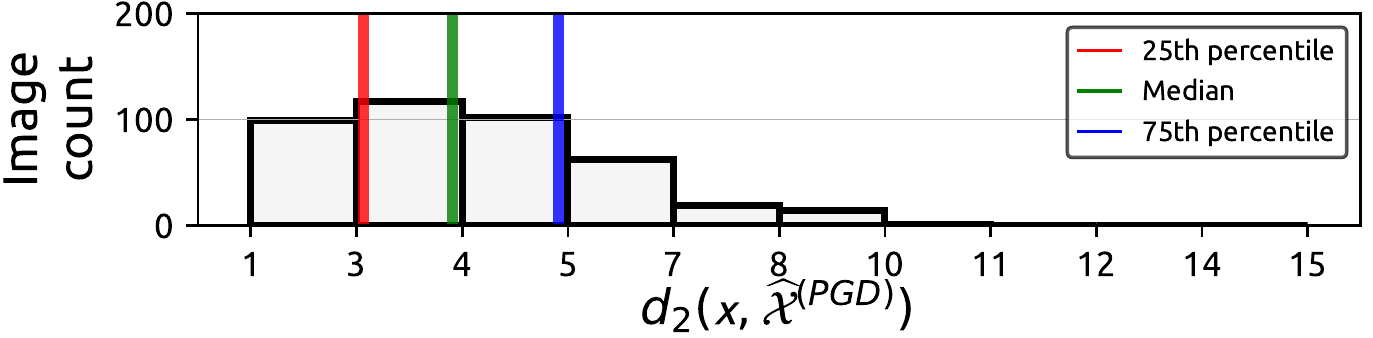}\hspace{1em}
\includegraphics[width=0.4\linewidth]{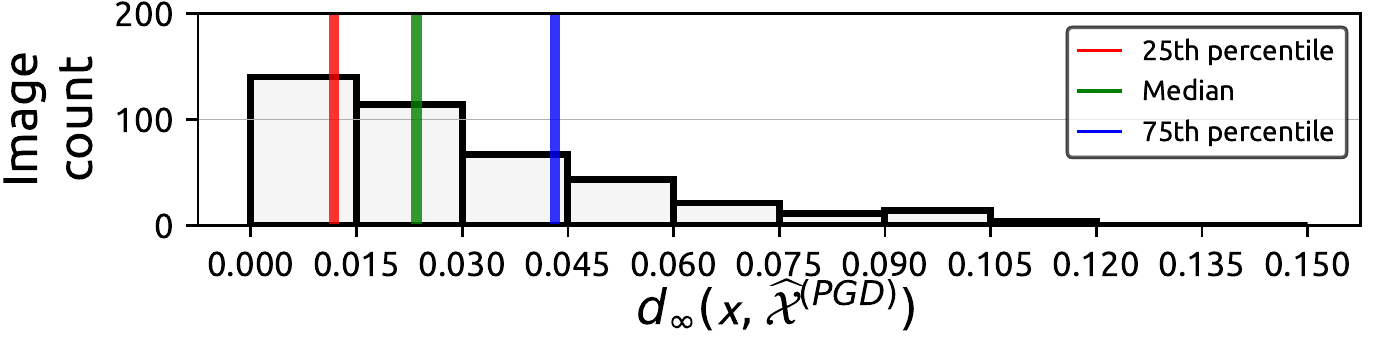}
\rotatebox[origin=l]{180}{\phantom{------}}
\\
\hspace{5.5em}
\\
(a) Adversarial examples transferred to \textbf{ResNet-50} with \textbf{PGD}.
\vspace{1em}
\\
\rotatebox[origin=l]{90}{\phantom{---}\scriptsize\underline{$T(\cdot)\geq1$}}\hspace{0.5em}
\includegraphics[width=0.4\linewidth]{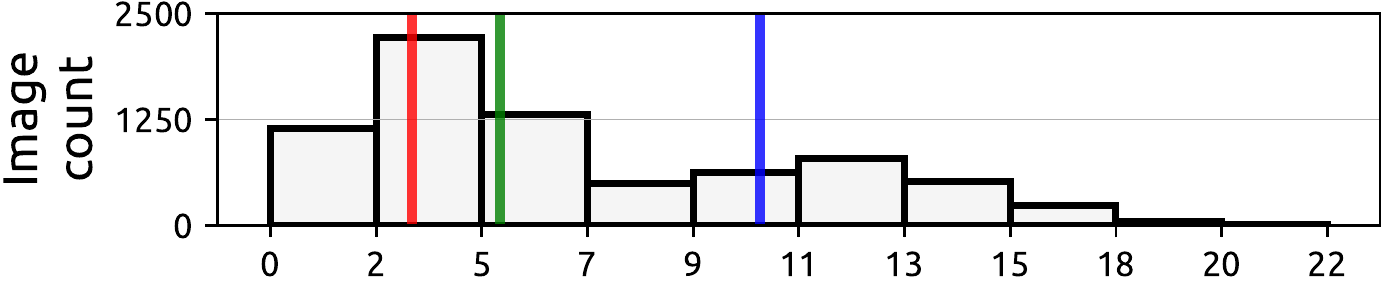}\hspace{1em}
\includegraphics[width=0.4\linewidth]{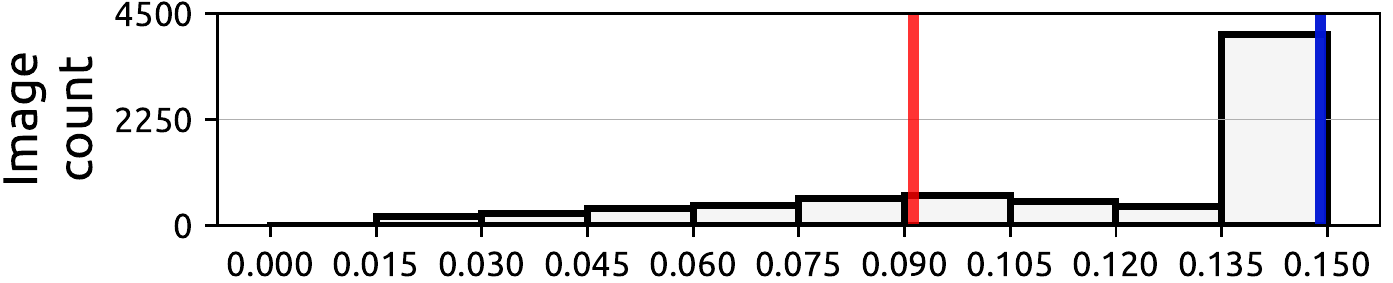}
\rotatebox[origin=l]{180}{\phantom{------}}
\\
\vspace{0.2em}
\rotatebox[origin=l]{90}{\phantom{--}\scriptsize\underline{$T(\cdot)\geq20$}}\hspace{0.5em}
\includegraphics[width=0.4\linewidth]{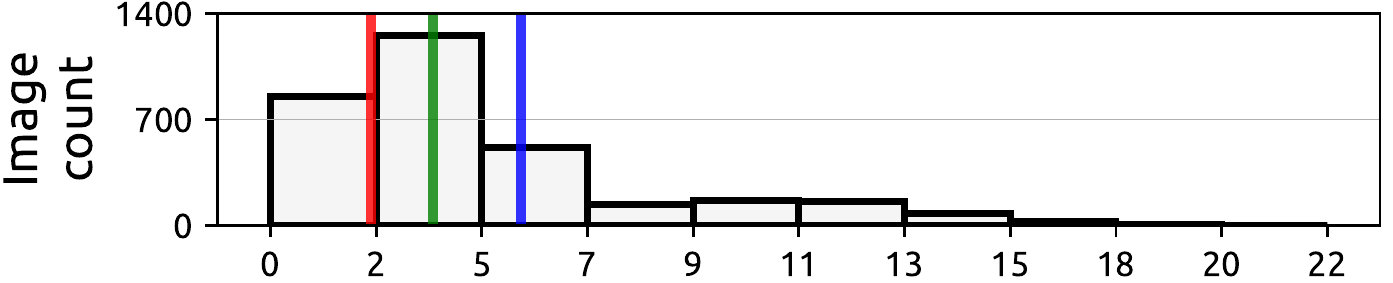}\hspace{1em}
\includegraphics[width=0.4\linewidth]{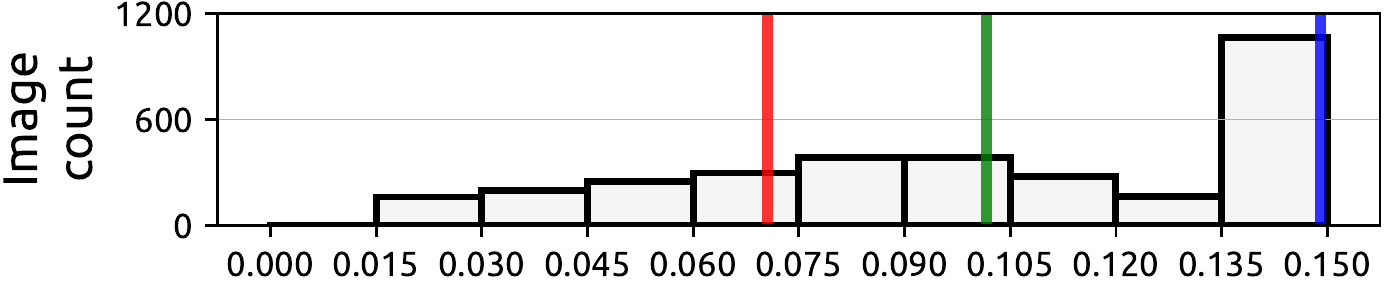}
\rotatebox[origin=l]{180}{\phantom{------}}
\\
\vspace{0.2em}
\rotatebox[origin=l]{90}{\phantom{----}\scriptsize\underline{$T(\cdot)\geq30$}}\hspace{0.5em}
\includegraphics[width=0.4\linewidth]{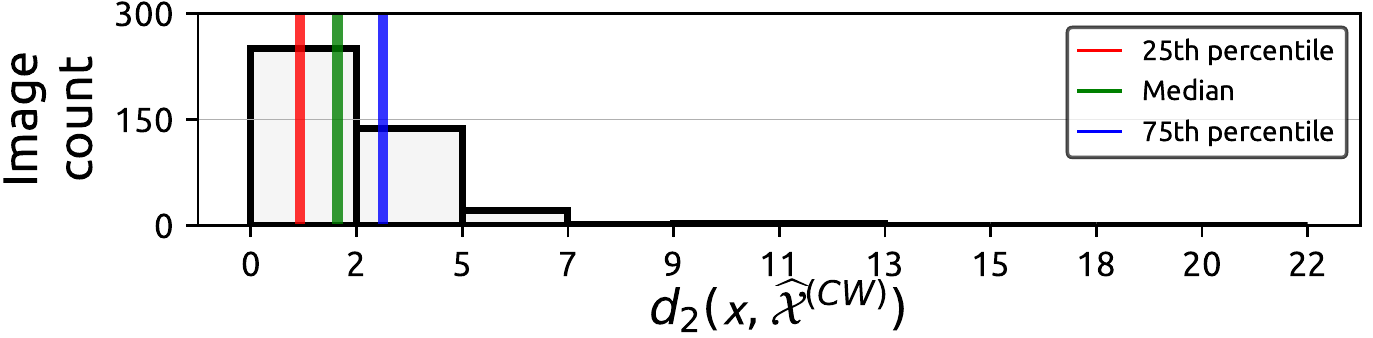}\hspace{1em}
\includegraphics[width=0.4\linewidth]{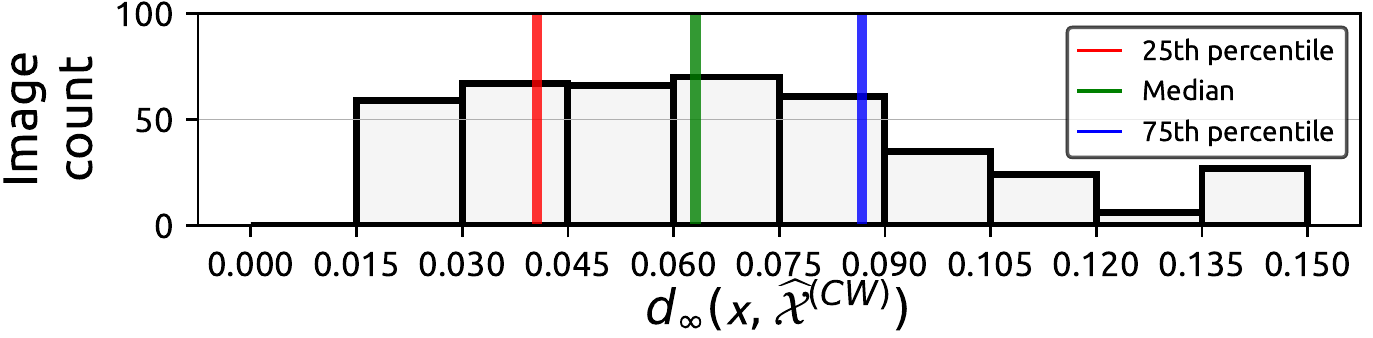}
\rotatebox[origin=l]{180}{\phantom{------}}
\\
\hspace{5.5em}
\\
(b) Adversarial examples transferred to \textbf{ResNet-50} with \textbf{CW}.
\vspace{1em}
\\
\rotatebox[origin=l]{90}{\phantom{---}\scriptsize\underline{$T(\cdot)\geq1$}}\hspace{0.5em}
\includegraphics[width=0.4\linewidth]{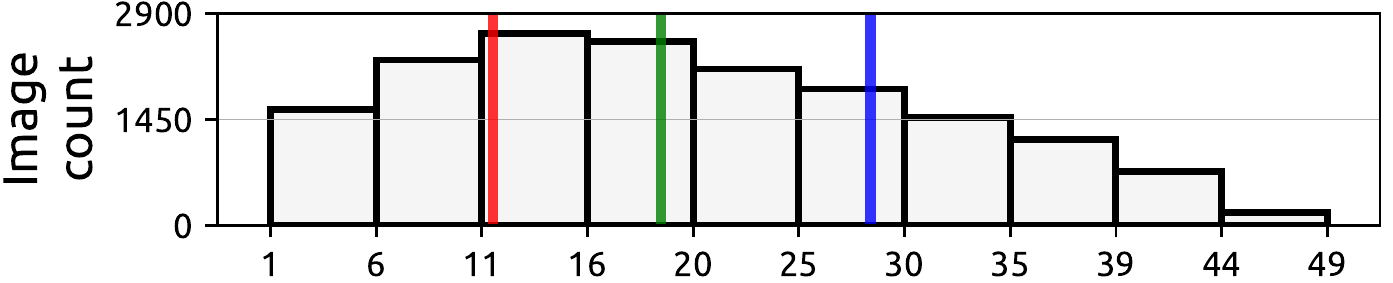}\hspace{1em}
\includegraphics[width=0.4\linewidth]{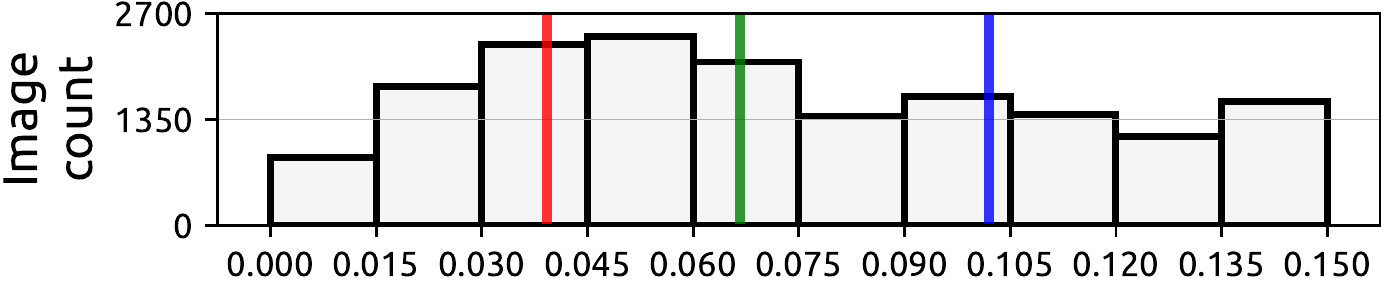}
\rotatebox[origin=l]{180}{\phantom{------}}
\\
\vspace{0.2em}
\rotatebox[origin=l]{90}{\phantom{--}\scriptsize\underline{$T(\cdot)\geq20$}}\hspace{0.5em}
\includegraphics[width=0.4\linewidth]{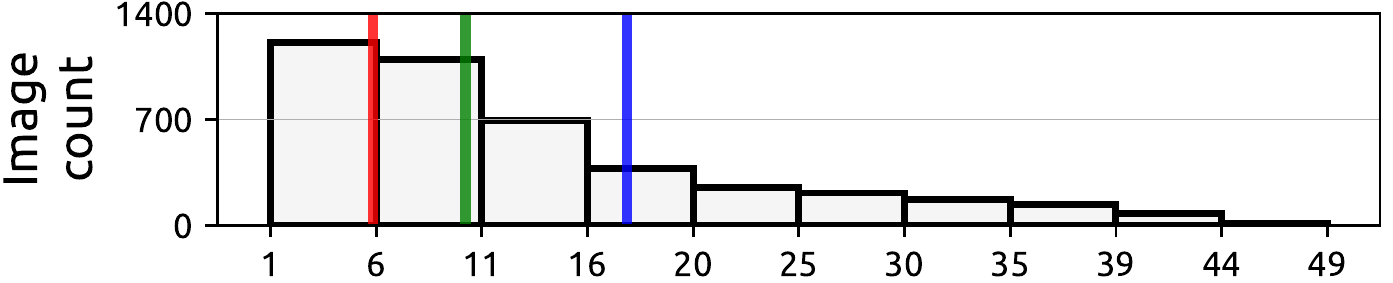}\hspace{1em}
\includegraphics[width=0.4\linewidth]{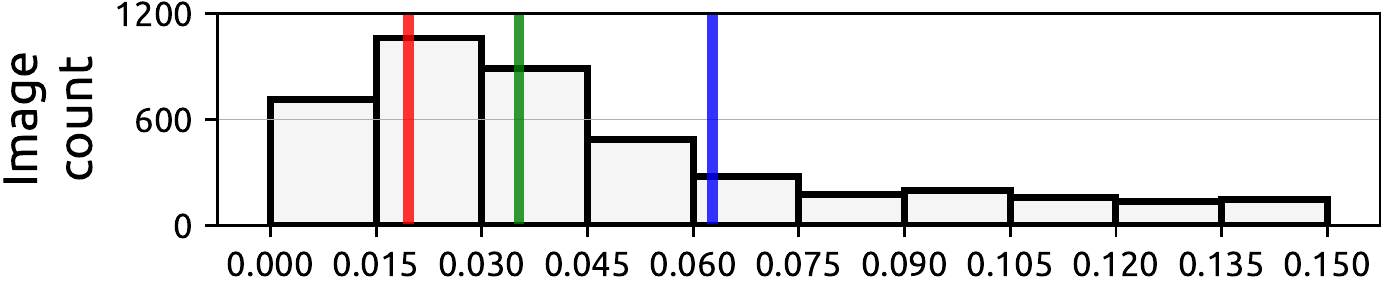}
\rotatebox[origin=l]{180}{\phantom{------}}
\\
\vspace{0.2em}
\rotatebox[origin=l]{90}{\phantom{----}\scriptsize\underline{$T(\cdot)\geq30$}}\hspace{0.5em}
\includegraphics[width=0.4\linewidth]{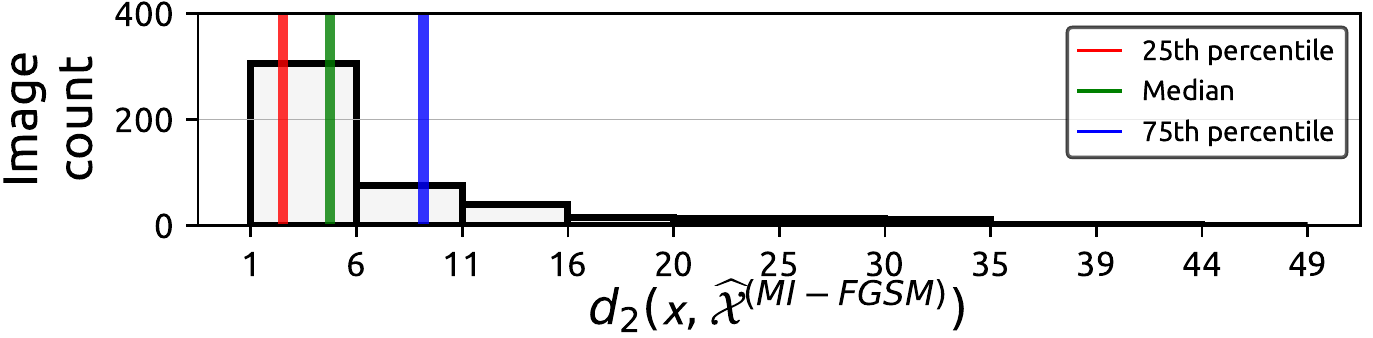}\hspace{1em}
\includegraphics[width=0.4\linewidth]{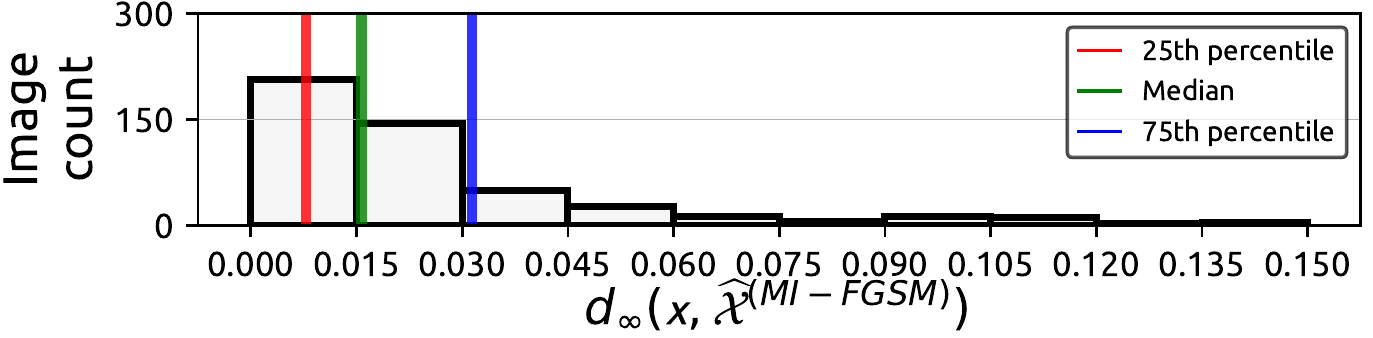}
\rotatebox[origin=l]{180}{\phantom{------}}
\\
\hspace{5.5em}
\\
(c) Adversarial examples transferred to \textbf{ResNet-50} with \textbf{MI-FGSM}.
\vspace{2.5em}
\caption{Source images that achieved adversarial transferability to \textbf{ResNet-50} are selected based on transferability count, with $T(\Theta, \widehat{\mathcal{X}}^{\text{(A)}}, \bm{y}) \geq \{1, 20, 30\}$. The minimum amount of perturbation required for creating adversarial examples from these source images is histogrammed, measuring the perturbation using $d_p(\bm{x}, \widehat{\mathcal{X}}^{\text{(A)}})$, with $p\in \{2,\infty\}$. The median perturbation, as well as the $25$th and the $75$th percentile, are provided in order to improve interpretability.}
\label{fig:pert_norm_resnet50}
\end{figure*}

\clearpage

\begin{figure*}[hbtp!]
\centering
\rotatebox[origin=l]{90}{\phantom{---}\scriptsize\underline{$T(\cdot)\geq1$}}\hspace{0.5em}
\includegraphics[width=0.4\linewidth]{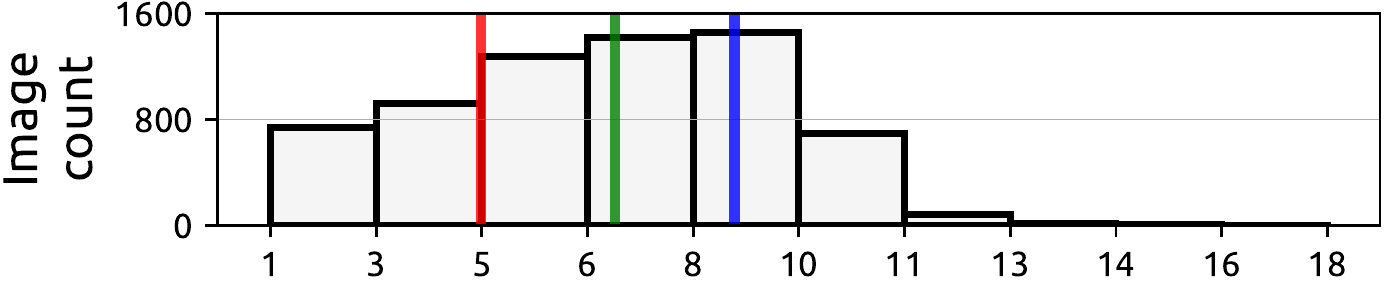}\hspace{1em}
\includegraphics[width=0.4\linewidth]{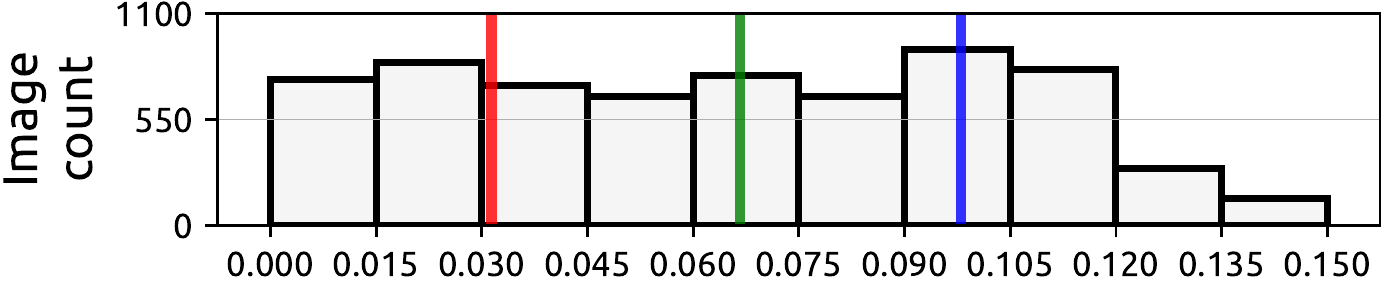}
\rotatebox[origin=l]{180}{\phantom{------}}
\\
\vspace{0.2em}
\rotatebox[origin=l]{90}{\phantom{--}\scriptsize\underline{$T(\cdot)\geq20$}}\hspace{0.5em}
\includegraphics[width=0.4\linewidth]{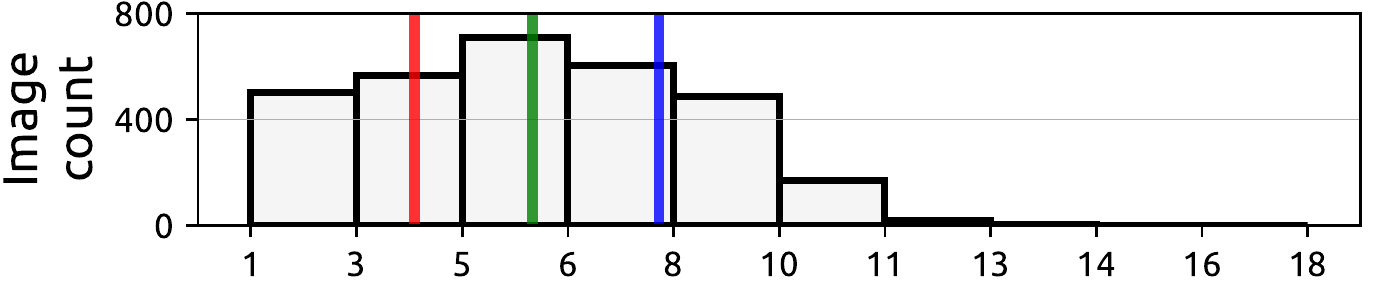}\hspace{1em}
\includegraphics[width=0.4\linewidth]{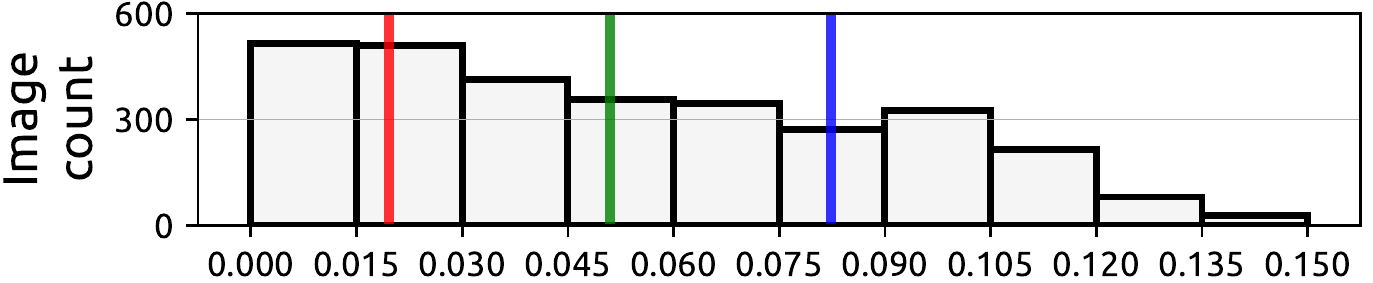}
\rotatebox[origin=l]{180}{\phantom{------}}
\\
\vspace{0.2em}
\rotatebox[origin=l]{90}{\phantom{----}\scriptsize\underline{$T(\cdot)\geq30$}}\hspace{0.5em}
\includegraphics[width=0.4\linewidth]{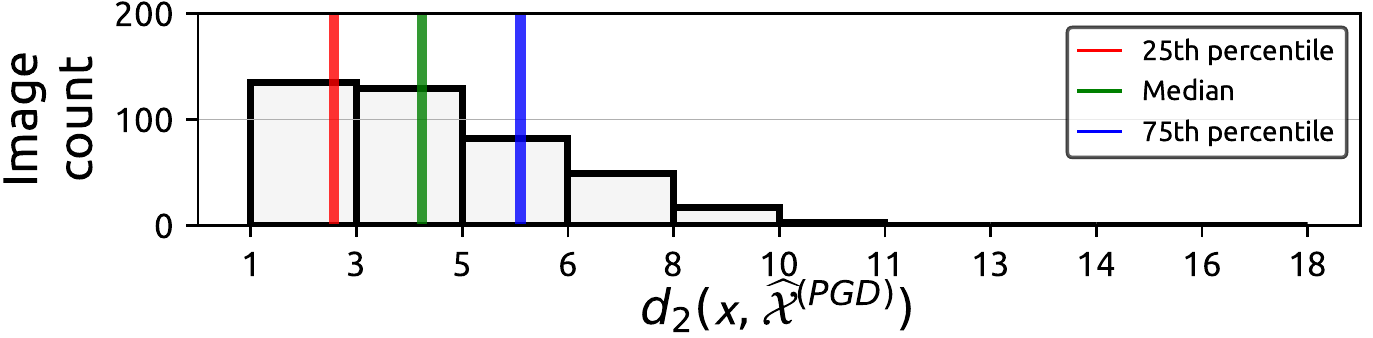}\hspace{1em}
\includegraphics[width=0.4\linewidth]{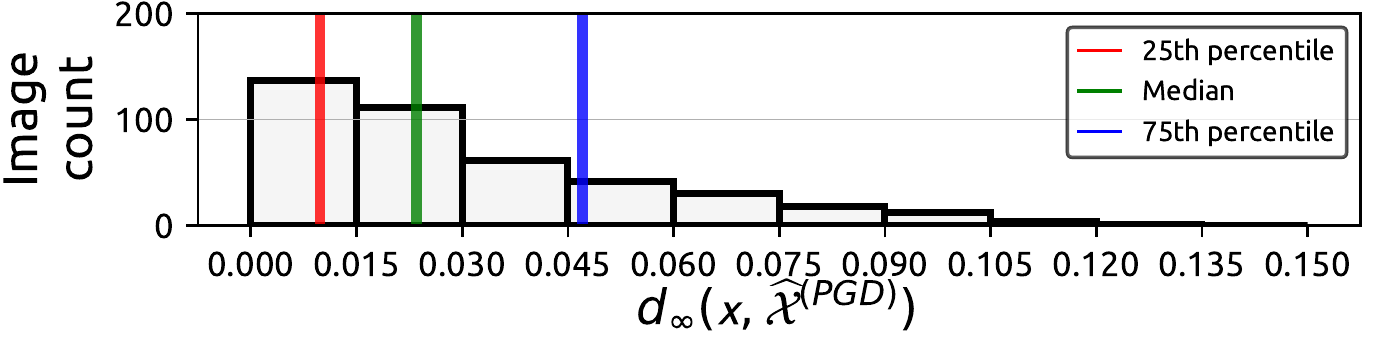}
\rotatebox[origin=l]{180}{\phantom{------}}
\\
\hspace{5.5em}
\\
(a) Adversarial examples transferred to \textbf{DenseNet-121} with \textbf{PGD}.
\vspace{1em}
\\
\rotatebox[origin=l]{90}{\phantom{---}\scriptsize\underline{$T(\cdot)\geq1$}}\hspace{0.5em}
\includegraphics[width=0.4\linewidth]{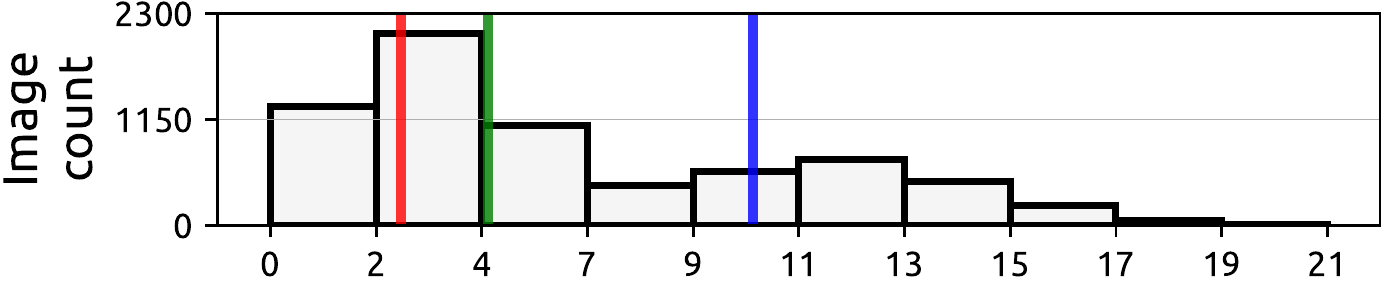}\hspace{1em}
\includegraphics[width=0.4\linewidth]{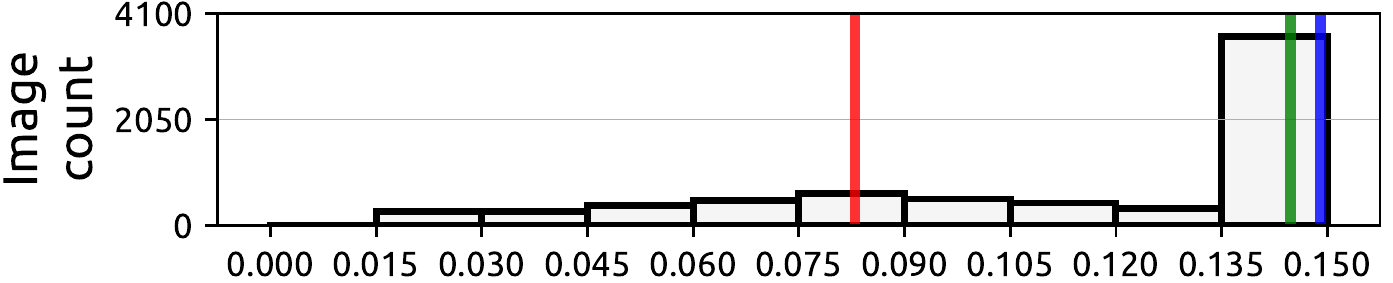}
\rotatebox[origin=l]{180}{\phantom{------}}
\\
\vspace{0.2em}
\rotatebox[origin=l]{90}{\phantom{--}\scriptsize\underline{$T(\cdot)\geq20$}}\hspace{0.5em}
\includegraphics[width=0.4\linewidth]{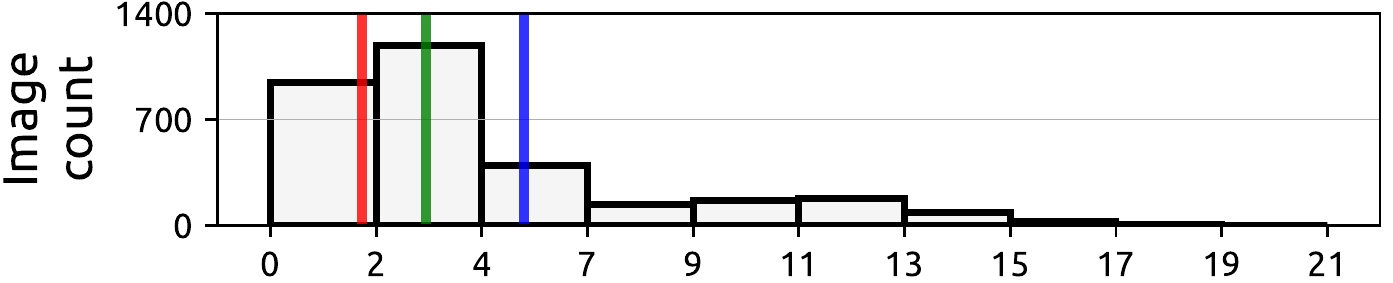}\hspace{1em}
\includegraphics[width=0.4\linewidth]{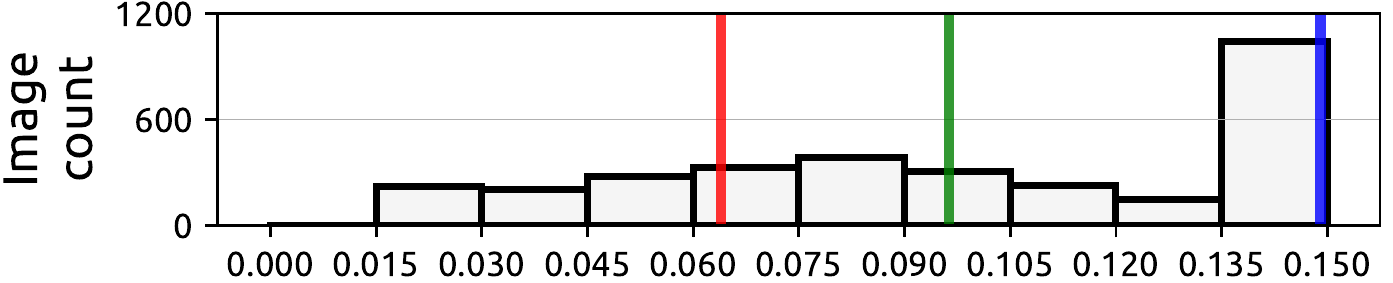}
\rotatebox[origin=l]{180}{\phantom{------}}
\\
\vspace{0.2em}
\rotatebox[origin=l]{90}{\phantom{----}\scriptsize\underline{$T(\cdot)\geq30$}}\hspace{0.5em}
\includegraphics[width=0.4\linewidth]{bmvc_hist/resnet50_transfer_30_CW_pert.pdf}\hspace{1em}
\includegraphics[width=0.4\linewidth]{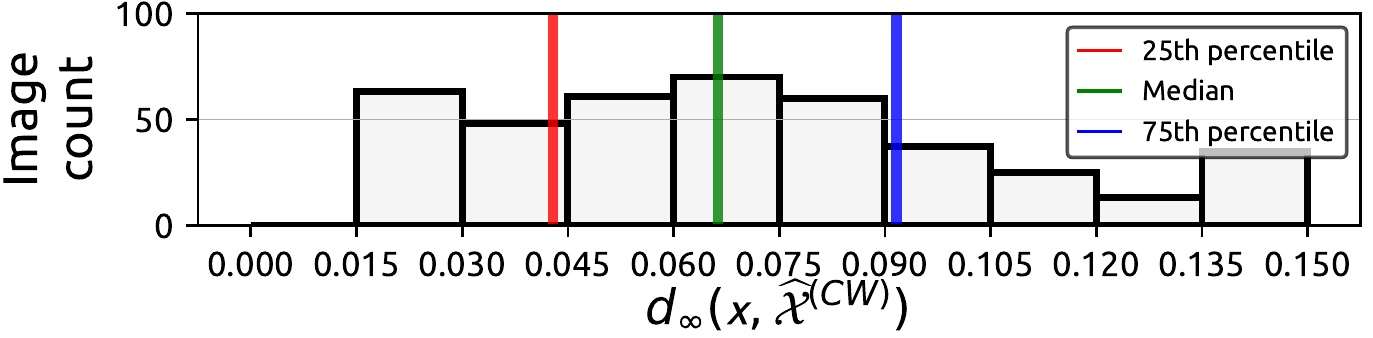}
\rotatebox[origin=l]{180}{\phantom{------}}
\\
\hspace{5.5em}
\\
(b) Adversarial examples transferred to \textbf{DenseNet-121} with \textbf{CW}.
\vspace{1em}
\\
\rotatebox[origin=l]{90}{\phantom{---}\scriptsize\underline{$T(\cdot)\geq1$}}\hspace{0.5em}
\includegraphics[width=0.4\linewidth]{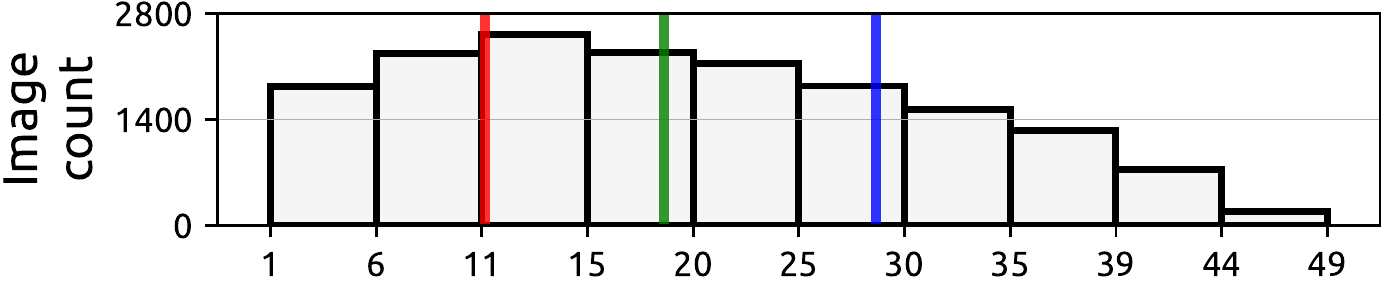}\hspace{1em}
\includegraphics[width=0.4\linewidth]{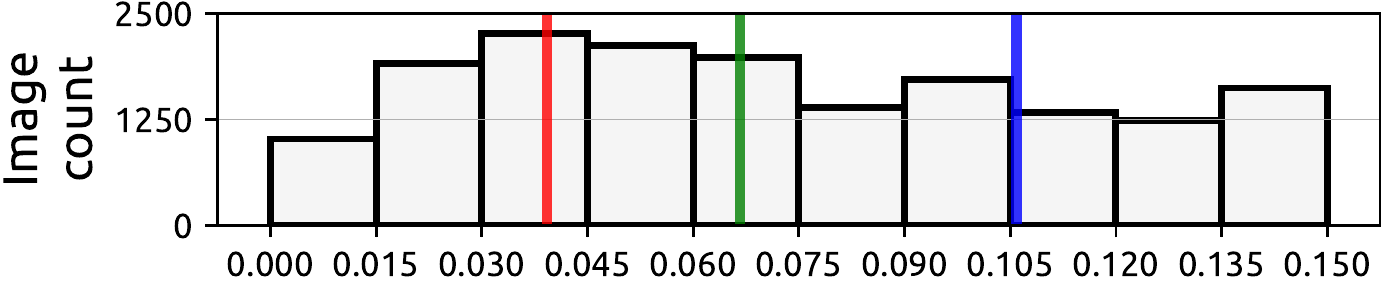}
\rotatebox[origin=l]{180}{\phantom{------}}
\\
\vspace{0.2em}
\rotatebox[origin=l]{90}{\phantom{--}\scriptsize\underline{$T(\cdot)\geq20$}}\hspace{0.5em}
\includegraphics[width=0.4\linewidth]{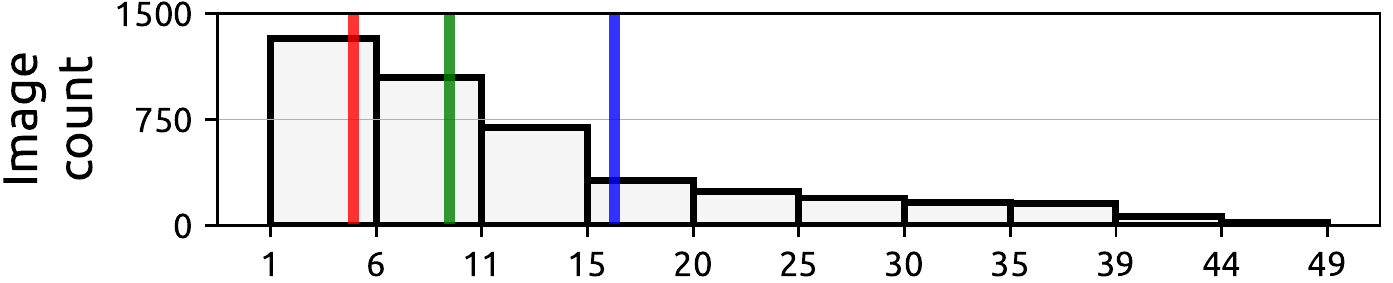}\hspace{1em}
\includegraphics[width=0.4\linewidth]{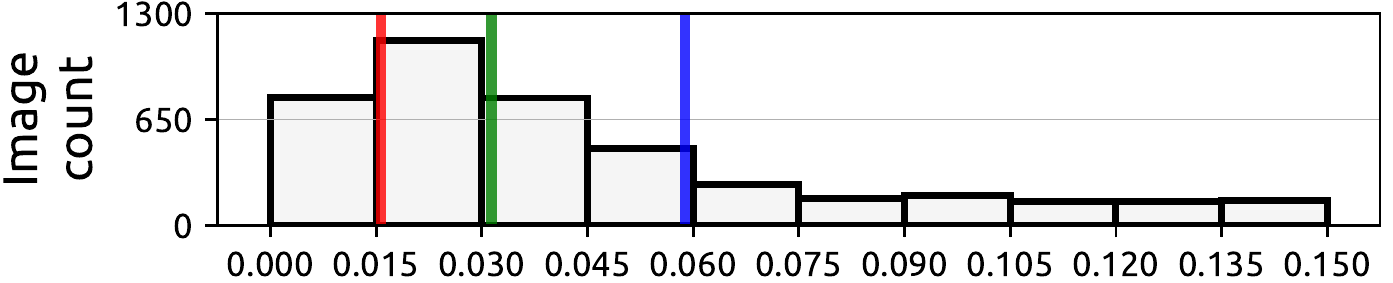}
\rotatebox[origin=l]{180}{\phantom{------}}
\\
\vspace{0.2em}
\rotatebox[origin=l]{90}{\phantom{----}\scriptsize\underline{$T(\cdot)\geq30$}}\hspace{0.5em}
\includegraphics[width=0.4\linewidth]{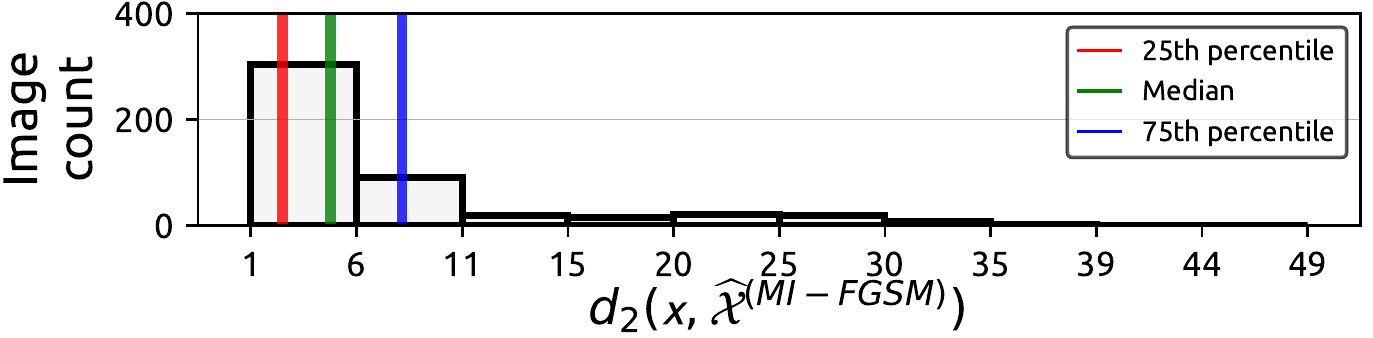}\hspace{1em}
\includegraphics[width=0.4\linewidth]{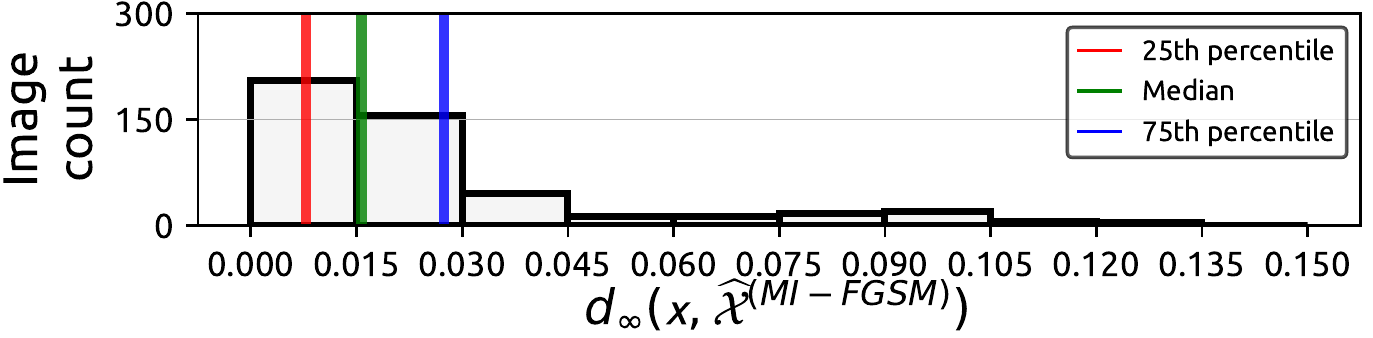}
\rotatebox[origin=l]{180}{\phantom{------}}
\\
\hspace{5.5em}
\\
(c) Adversarial examples transferred to \textbf{DenseNet-121} with \textbf{MI-FGSM}.
\vspace{2.5em}
\caption{Source images that achieved adversarial transferability to \textbf{DenseNet-121} are selected based on transferability count, with $T(\Theta, \widehat{\mathcal{X}}^{\text{(A)}}, \bm{y}) \geq \{1, 20, 30\}$. The minimum amount of perturbation required for creating adversarial examples from these source images is histogrammed, measuring the perturbation using $d_p(\bm{x}, \widehat{\mathcal{X}}^{\text{(A)}})$, with $p\in \{2,\infty\}$. The median perturbation, as well as the $25$th and the $75$th percentile, are provided in order to improve interpretability.}
\label{fig:pert_norm_densenet}
\end{figure*}

\clearpage

\begin{figure*}[hbtp!]
\centering
\rotatebox[origin=l]{90}{\phantom{---}\scriptsize\underline{$T(\cdot)\geq1$}}\hspace{0.5em}
\includegraphics[width=0.4\linewidth]{bmvc_hist/vit_base_PGD_pert.pdf}\hspace{1em}
\includegraphics[width=0.4\linewidth]{bmvc_hist/vit_base_PGD_linf_pert.pdf}
\rotatebox[origin=l]{180}{\phantom{------}}
\\
\vspace{0.2em}
\rotatebox[origin=l]{90}{\phantom{--}\scriptsize\underline{$T(\cdot)\geq20$}}\hspace{0.5em}
\includegraphics[width=0.4\linewidth]{bmvc_hist/vit_base_transfer_20_PGD_pert.pdf}\hspace{1em}
\includegraphics[width=0.4\linewidth]{bmvc_hist/vit_base_transfer_20_PGD_linf_pert.pdf}
\rotatebox[origin=l]{180}{\phantom{------}}
\\
\vspace{0.2em}
\rotatebox[origin=l]{90}{\phantom{----}\scriptsize\underline{$T(\cdot)\geq30$}}\hspace{0.5em}
\includegraphics[width=0.4\linewidth]{bmvc_hist/vit_base_transfer_30_PGD_pert.pdf}\hspace{1em}
\includegraphics[width=0.4\linewidth]{bmvc_hist/vit_base_transfer_30_PGD_linf_pert.pdf}
\rotatebox[origin=l]{180}{\phantom{------}}
\\
\hspace{5.5em}
\\
(a) Adversarial examples transferred to \textbf{ViT-B} with \textbf{PGD}.
\vspace{1em}
\\
\rotatebox[origin=l]{90}{\phantom{---}\scriptsize\underline{$T(\cdot)\geq1$}}\hspace{0.5em}
\includegraphics[width=0.4\linewidth]{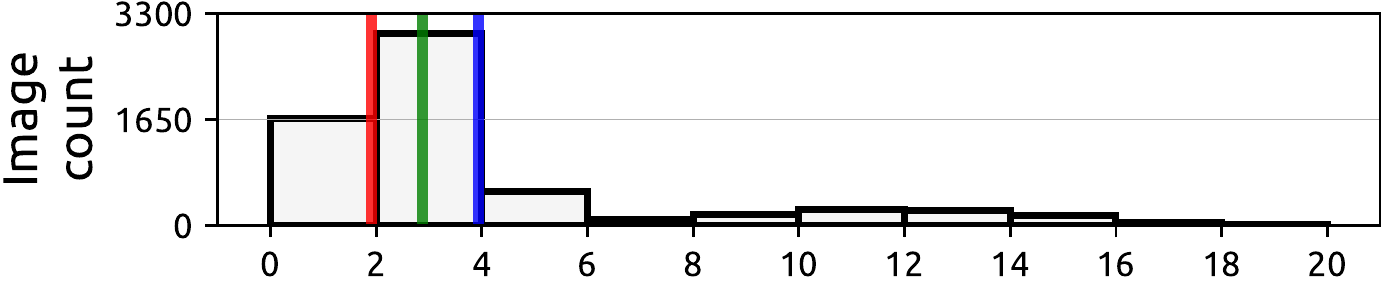}\hspace{1em}
\includegraphics[width=0.4\linewidth]{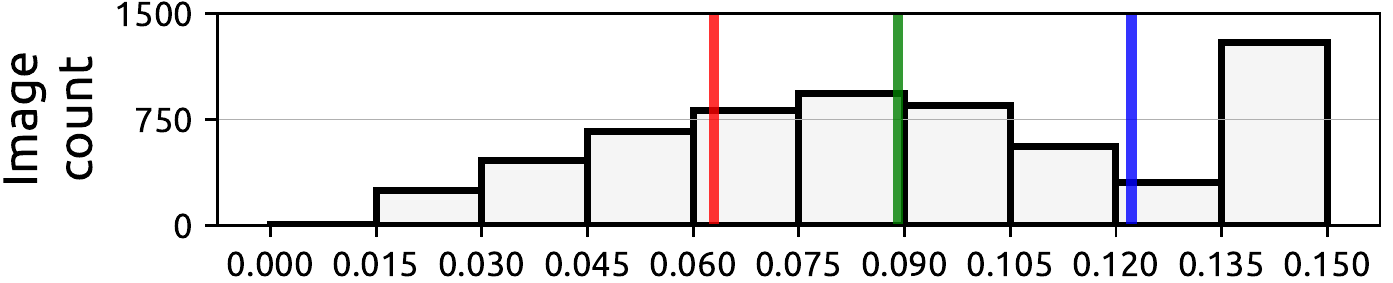}
\rotatebox[origin=l]{180}{\phantom{------}}
\\
\vspace{0.2em}
\rotatebox[origin=l]{90}{\phantom{--}\scriptsize\underline{$T(\cdot)\geq20$}}\hspace{0.5em}
\includegraphics[width=0.4\linewidth]{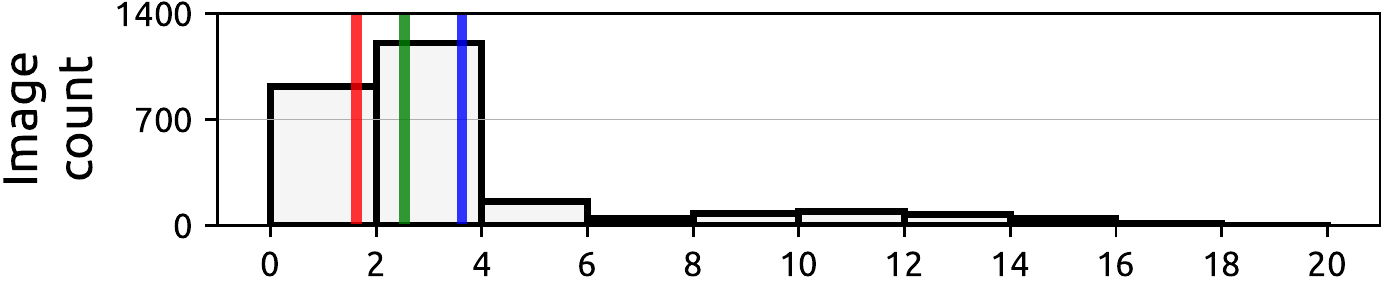}\hspace{1em}
\includegraphics[width=0.4\linewidth]{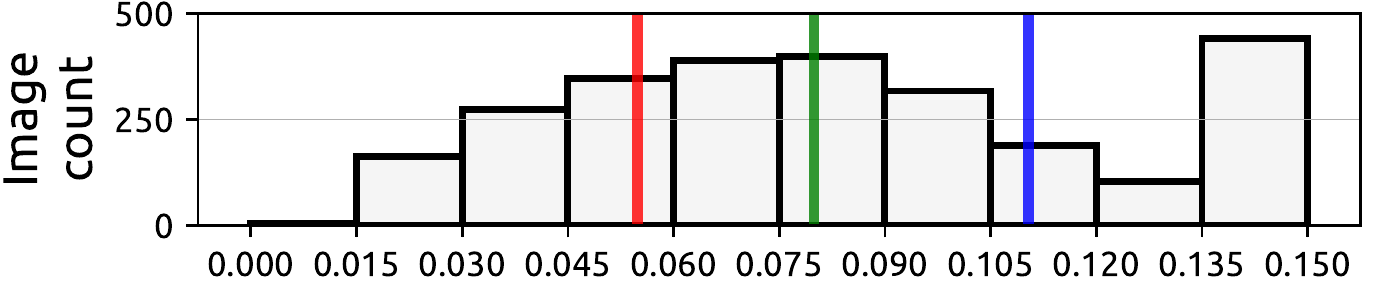}
\rotatebox[origin=l]{180}{\phantom{------}}
\\
\vspace{0.2em}
\rotatebox[origin=l]{90}{\phantom{----}\scriptsize\underline{$T(\cdot)\geq30$}}\hspace{0.5em}
\includegraphics[width=0.4\linewidth]{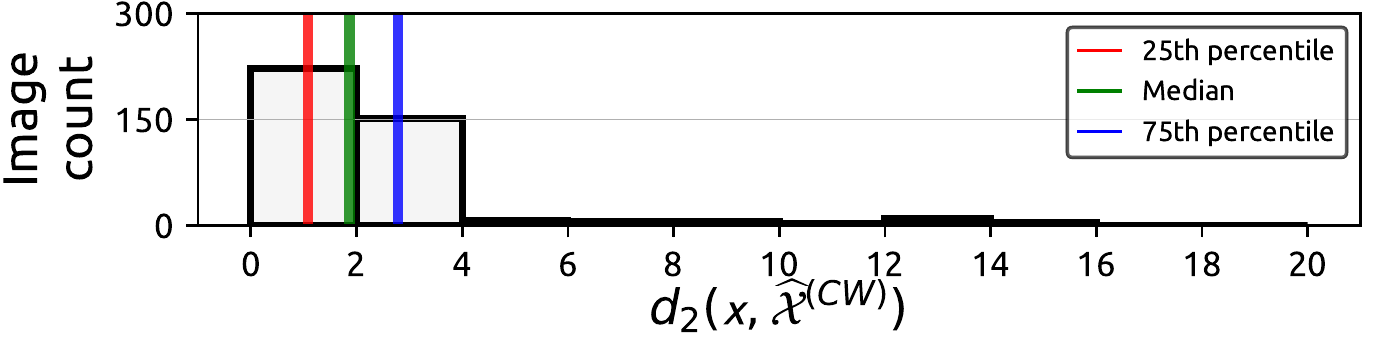}\hspace{1em}
\includegraphics[width=0.4\linewidth]{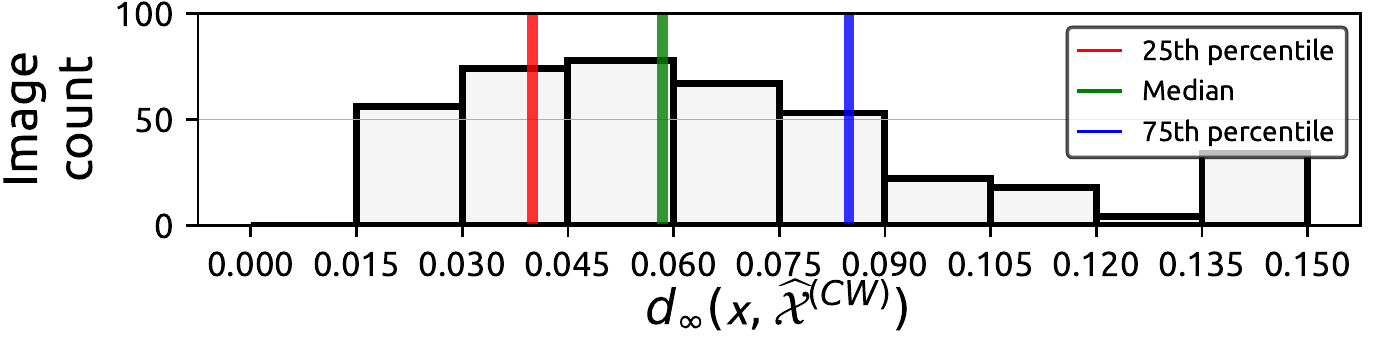}
\rotatebox[origin=l]{180}{\phantom{------}}
\\
\hspace{5.5em}
\\
(b) Adversarial examples transferred to \textbf{ViT-B} with \textbf{CW}.
\vspace{1em}
\\
\rotatebox[origin=l]{90}{\phantom{---}\scriptsize\underline{$T(\cdot)\geq1$}}\hspace{0.5em}
\includegraphics[width=0.4\linewidth]{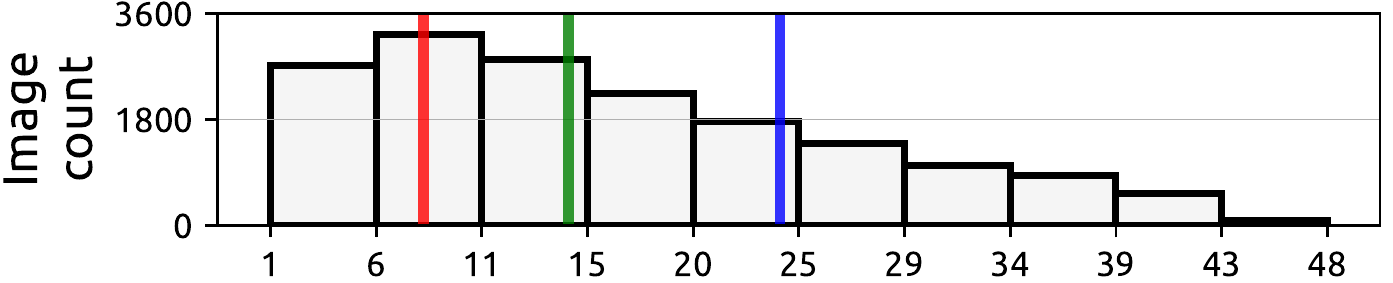}\hspace{1em}
\includegraphics[width=0.4\linewidth]{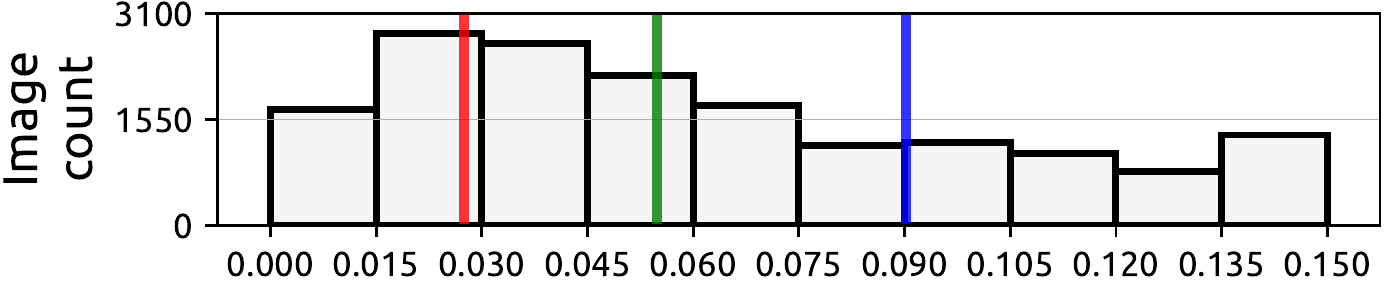}
\rotatebox[origin=l]{180}{\phantom{------}}
\\
\vspace{0.2em}
\rotatebox[origin=l]{90}{\phantom{--}\scriptsize\underline{$T(\cdot)\geq20$}}\hspace{0.5em}
\includegraphics[width=0.4\linewidth]{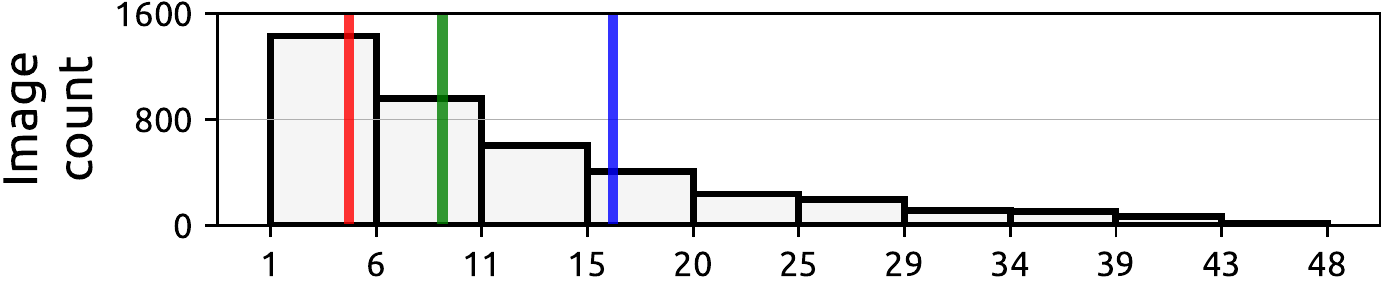}\hspace{1em}
\includegraphics[width=0.4\linewidth]{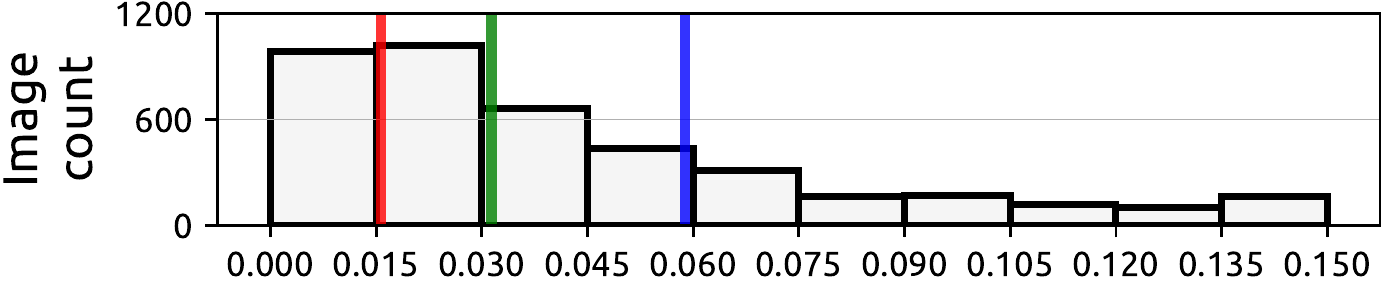}
\rotatebox[origin=l]{180}{\phantom{------}}
\\
\vspace{0.2em}
\rotatebox[origin=l]{90}{\phantom{----}\scriptsize\underline{$T(\cdot)\geq30$}}\hspace{0.5em}
\includegraphics[width=0.4\linewidth]{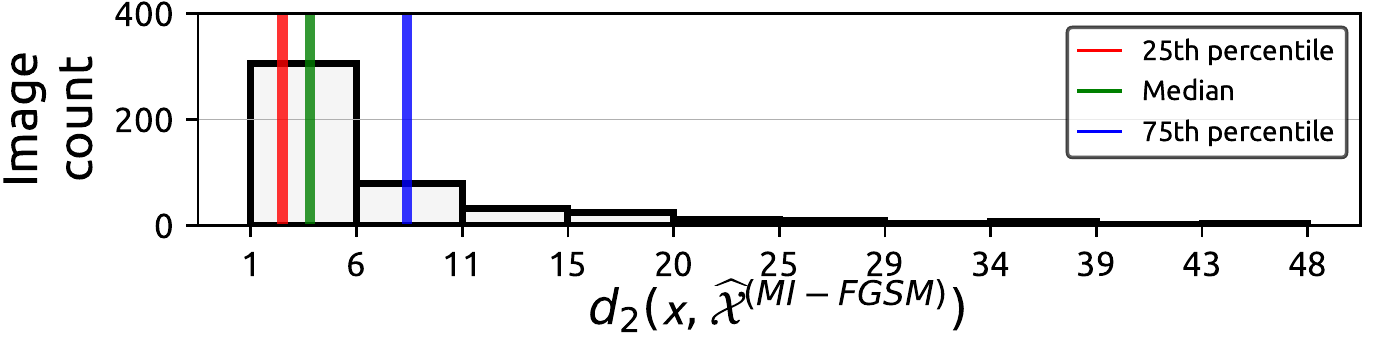}\hspace{1em}
\includegraphics[width=0.4\linewidth]{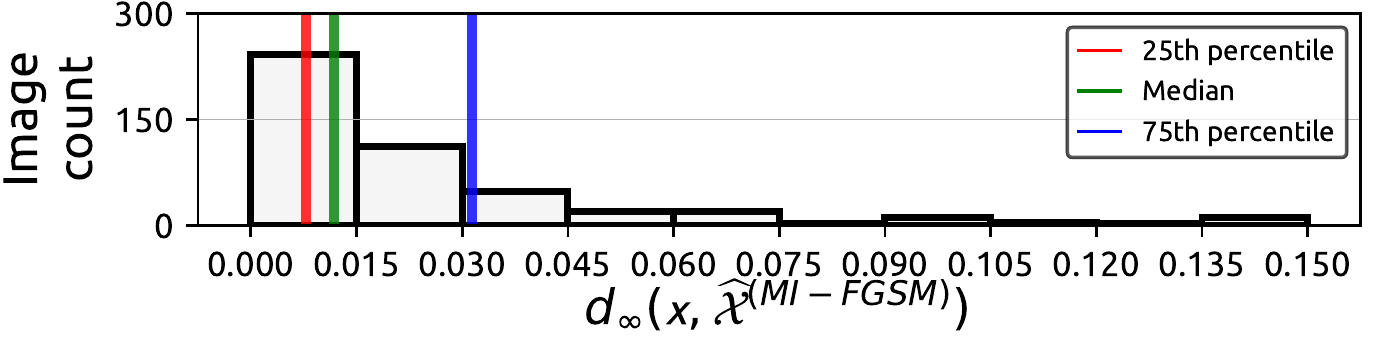}
\rotatebox[origin=l]{180}{\phantom{------}}
\\
\hspace{5.5em}
\\
(c) Adversarial examples transferred to \textbf{ViT-B} with \textbf{MI-FGSM}.
\vspace{2.5em}
\caption{Source images that achieved adversarial transferability to \textbf{ViT-B} are selected based on transferability count, with $T(\Theta, \widehat{\mathcal{X}}^{\text{(A)}}, \bm{y}) \geq \{1, 20, 30\}$. The minimum amount of perturbation required for creating adversarial examples from these source images is histogrammed, measuring the perturbation using $d_p(\bm{x}, \widehat{\mathcal{X}}^{\text{(A)}})$, with $p\in \{2,\infty\}$. The median perturbation, as well as the $25$th and the $75$th percentile, are provided in order to improve interpretability.}
\label{fig:pert_norm_vitb}
\end{figure*}

\clearpage

\begin{figure*}[hbtp!]
\centering
\rotatebox[origin=l]{90}{\phantom{---}\scriptsize\underline{$T(\cdot)\geq1$}}\hspace{0.5em}
\includegraphics[width=0.4\linewidth]{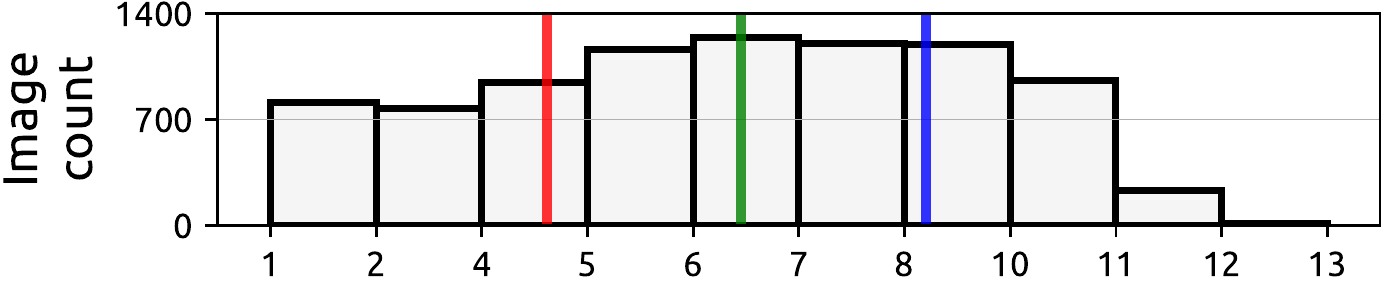}\hspace{1em}
\includegraphics[width=0.4\linewidth]{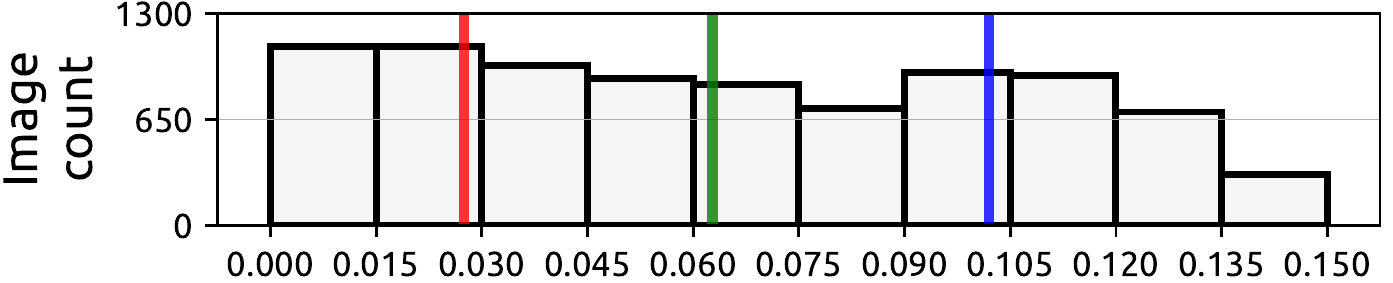}
\rotatebox[origin=l]{180}{\phantom{------}}
\\
\vspace{0.2em}
\rotatebox[origin=l]{90}{\phantom{--}\scriptsize\underline{$T(\cdot)\geq20$}}\hspace{0.5em}
\includegraphics[width=0.4\linewidth]{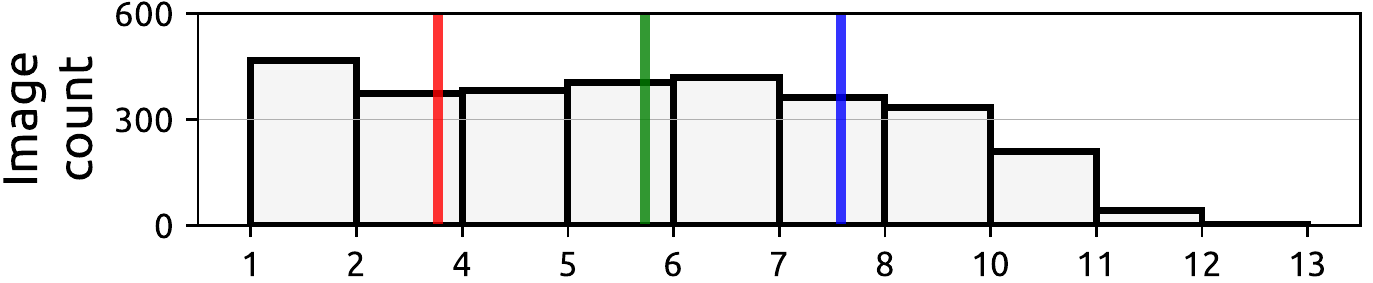}\hspace{1em}
\includegraphics[width=0.4\linewidth]{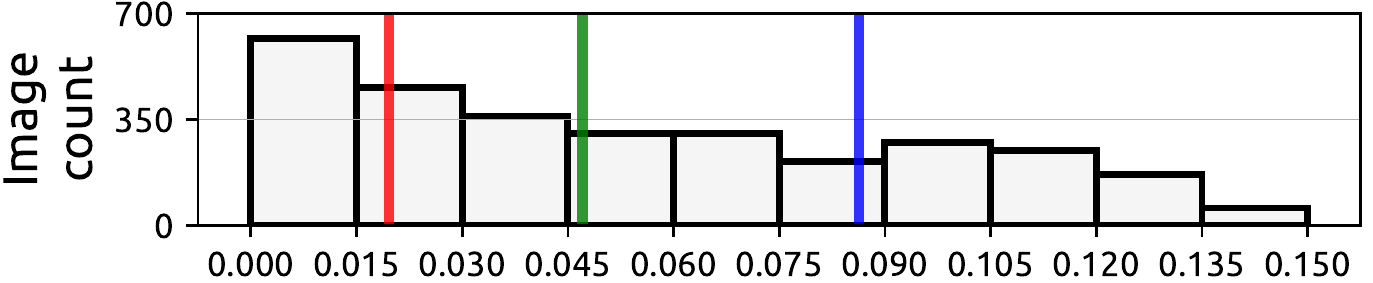}
\rotatebox[origin=l]{180}{\phantom{------}}
\\
\vspace{0.2em}
\rotatebox[origin=l]{90}{\phantom{----}\scriptsize\underline{$T(\cdot)\geq30$}}\hspace{0.5em}
\includegraphics[width=0.4\linewidth]{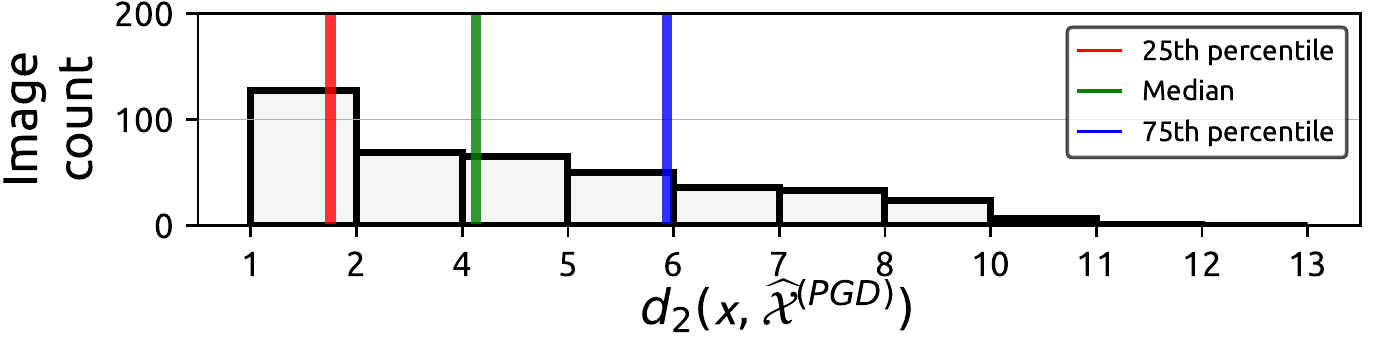}\hspace{1em}
\includegraphics[width=0.4\linewidth]{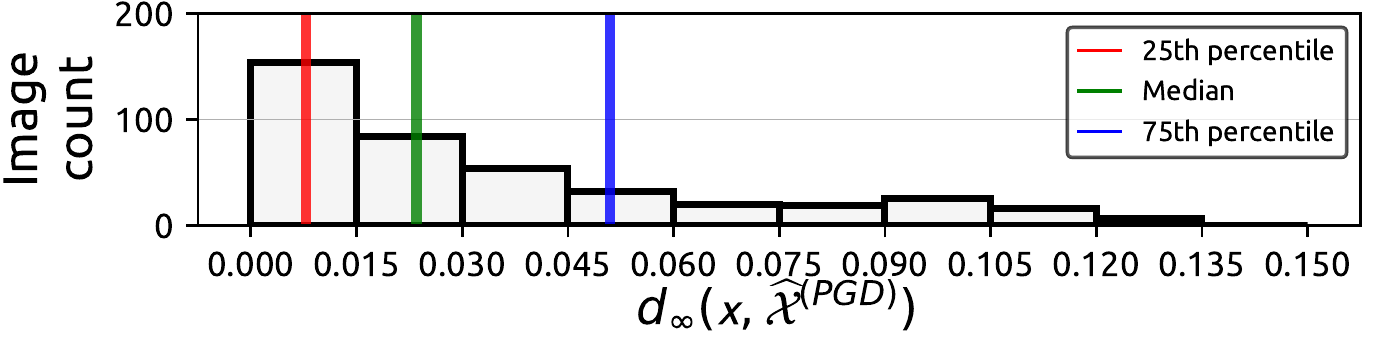}
\rotatebox[origin=l]{180}{\phantom{------}}
\\
\hspace{5.5em}
\\
(a) Adversarial examples transferred to \textbf{ViT-L} with \textbf{PGD}.
\vspace{1em}
\\
\rotatebox[origin=l]{90}{\phantom{---}\scriptsize\underline{$T(\cdot)\geq1$}}\hspace{0.5em}
\includegraphics[width=0.4\linewidth]{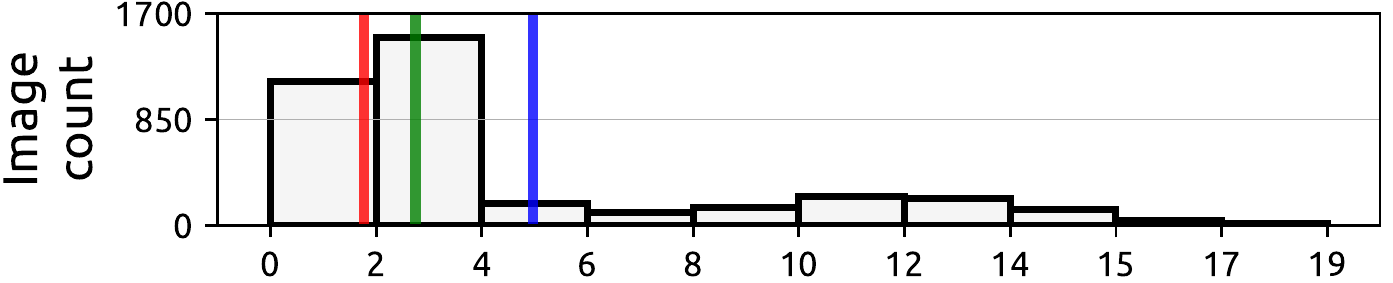}\hspace{1em}
\includegraphics[width=0.4\linewidth]{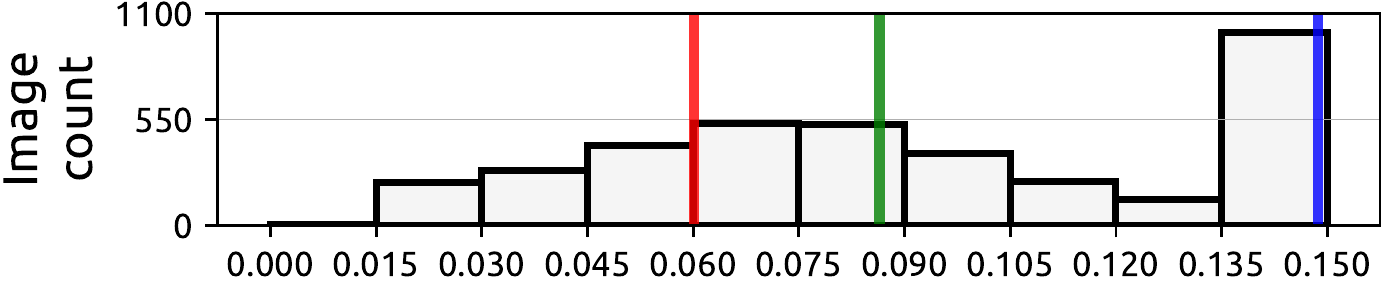}
\rotatebox[origin=l]{180}{\phantom{------}}
\\
\vspace{0.2em}
\rotatebox[origin=l]{90}{\phantom{--}\scriptsize\underline{$T(\cdot)\geq20$}}\hspace{0.5em}
\includegraphics[width=0.4\linewidth]{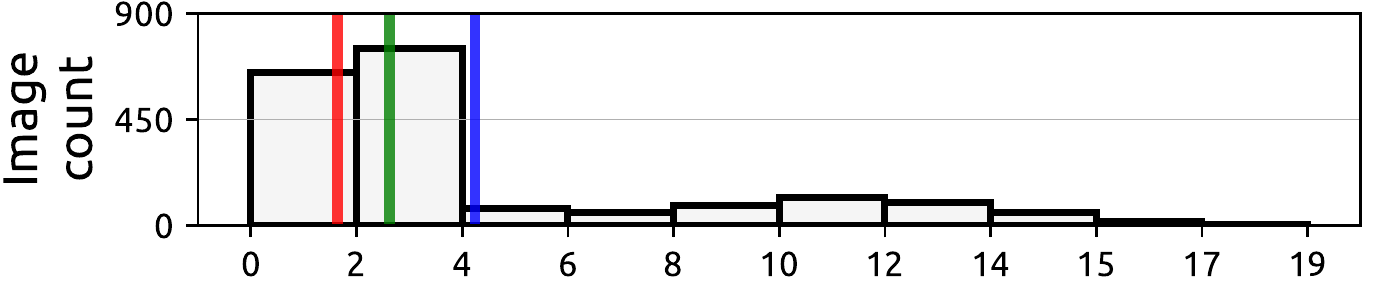}\hspace{1em}
\includegraphics[width=0.4\linewidth]{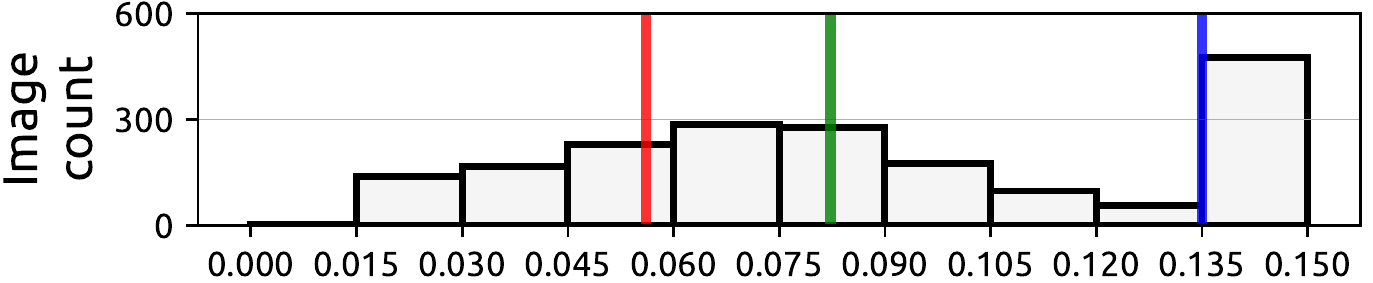}
\rotatebox[origin=l]{180}{\phantom{------}}
\\
\vspace{0.2em}
\rotatebox[origin=l]{90}{\phantom{----}\scriptsize\underline{$T(\cdot)\geq30$}}\hspace{0.5em}
\includegraphics[width=0.4\linewidth]{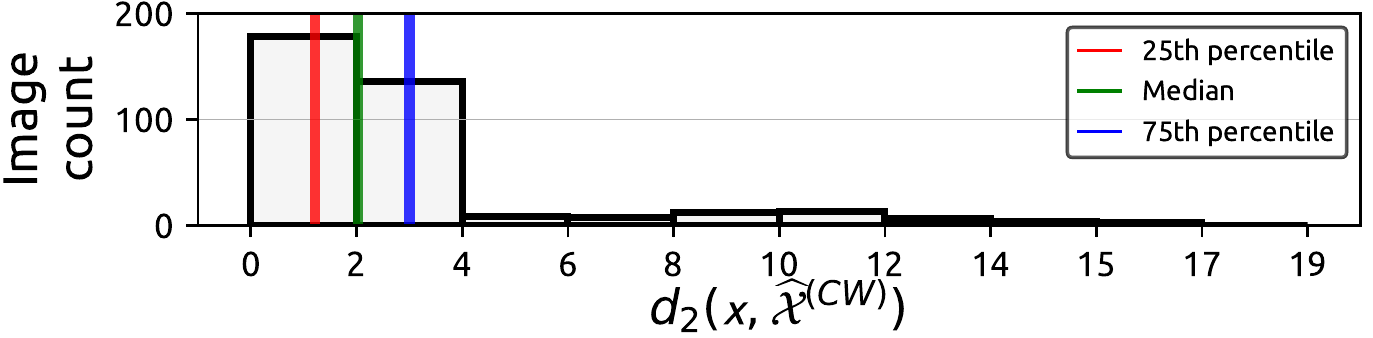}\hspace{1em}
\includegraphics[width=0.4\linewidth]{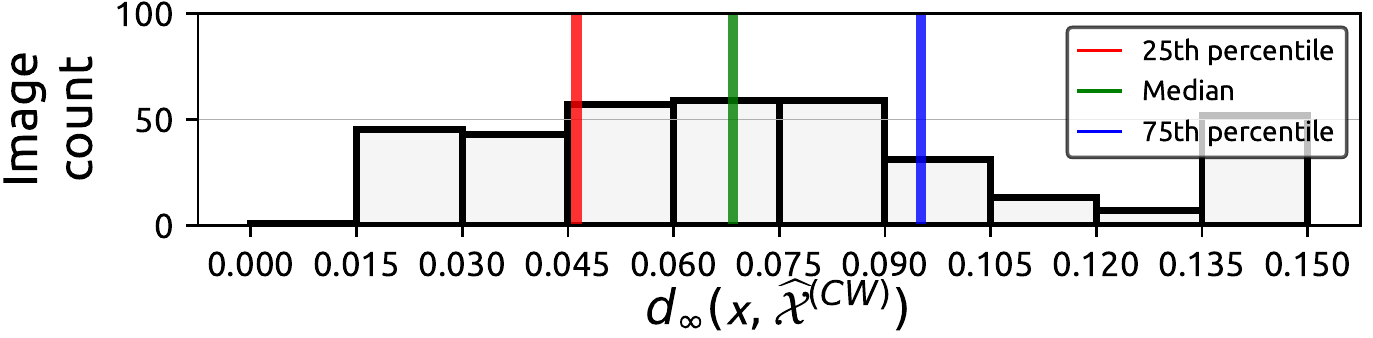}
\rotatebox[origin=l]{180}{\phantom{------}}
\\
\hspace{5.5em}
\\
(b) Adversarial examples transferred to \textbf{ViT-L} with \textbf{CW}.
\vspace{1em}
\\
\rotatebox[origin=l]{90}{\phantom{---}\scriptsize\underline{$T(\cdot)\geq1$}}\hspace{0.5em}
\includegraphics[width=0.4\linewidth]{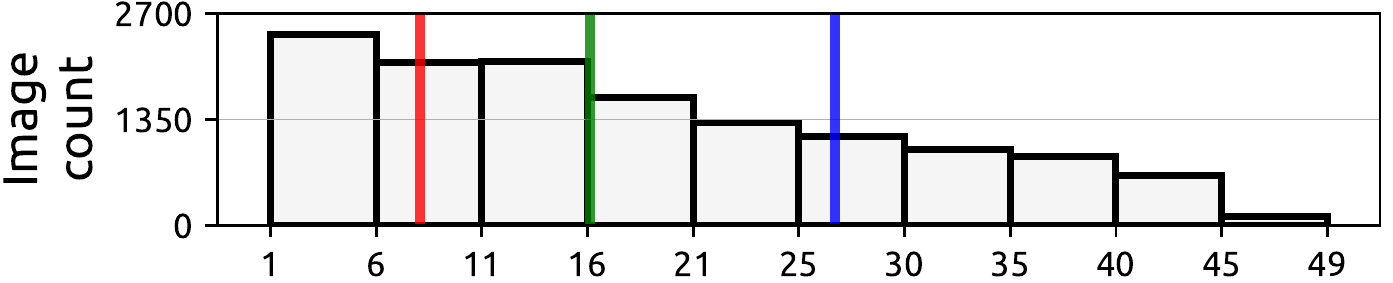}\hspace{1em}
\includegraphics[width=0.4\linewidth]{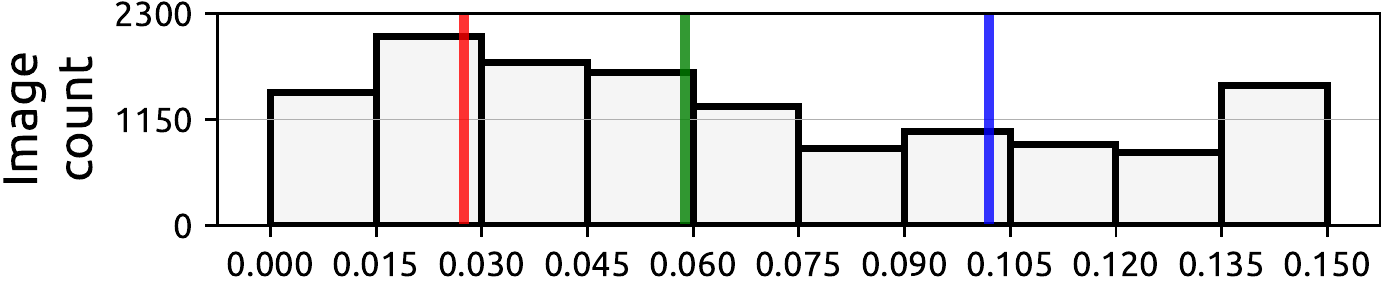}
\rotatebox[origin=l]{180}{\phantom{------}}
\\
\vspace{0.2em}
\rotatebox[origin=l]{90}{\phantom{--}\scriptsize\underline{$T(\cdot)\geq20$}}\hspace{0.5em}
\includegraphics[width=0.4\linewidth]{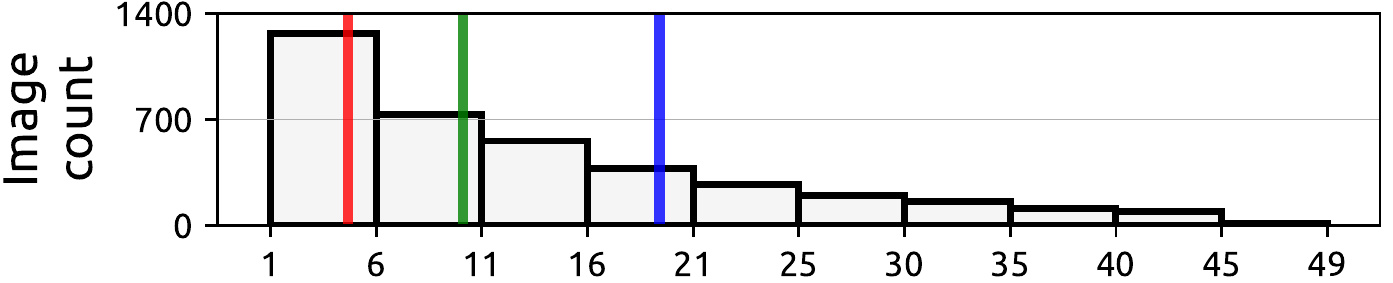}\hspace{1em}
\includegraphics[width=0.4\linewidth]{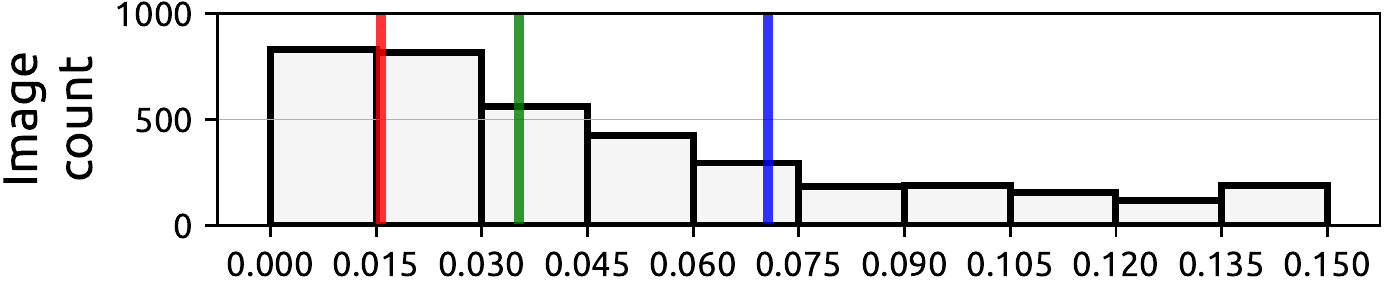}
\rotatebox[origin=l]{180}{\phantom{------}}
\\
\vspace{0.2em}
\rotatebox[origin=l]{90}{\phantom{----}\scriptsize\underline{$T(\cdot)\geq30$}}\hspace{0.5em}
\includegraphics[width=0.4\linewidth]{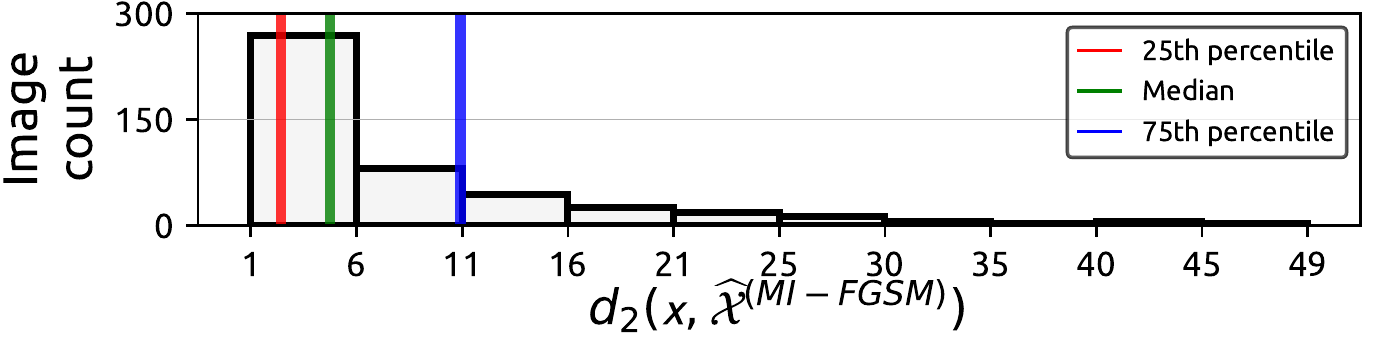}\hspace{1em}
\includegraphics[width=0.4\linewidth]{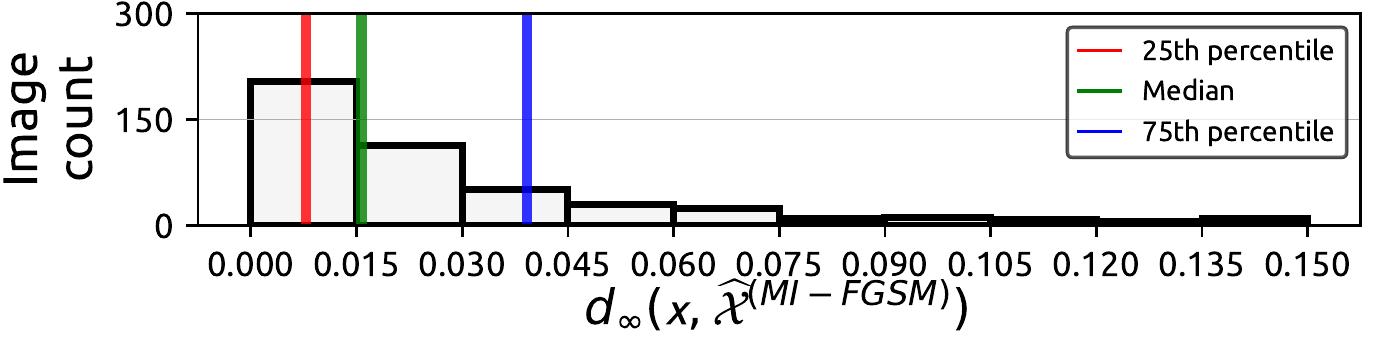}
\rotatebox[origin=l]{180}{\phantom{------}}
\\
\hspace{5.5em}
\\
(c) Adversarial examples transferred to \textbf{ViT-L} with \textbf{MI-FGSM}.
\vspace{2.5em}
\caption{Source images that achieved adversarial transferability to \textbf{ViT-L} are selected based on transferability count, with $T(\Theta, \widehat{\mathcal{X}}^{\text{(A)}}, \bm{y}) \geq \{1, 20, 30\}$. The minimum amount of perturbation required for creating adversarial examples from these source images is histogrammed, measuring the perturbation using $d_p(\bm{x}, \widehat{\mathcal{X}}^{\text{(A)}})$, with $p\in \{2,\infty\}$. The median perturbation, as well as the $25$th and the $75$th percentile, are provided in order to improve interpretability.}
\label{fig:pert_norm_vitl}
\end{figure*}

\clearpage
\section{Error estimates} 
In the main text, we briefly mentioned the usage of a number of error estimates in order to measure mistakes made in the prediction of source images. We denote with  $\bm{y}$ the true probabilistic categorical distribution associated with a data point $\bm{x}$ and assume that $c=\arg \max (\bm{y})$ is the true class and that $\hat{\bm{y}} = P (\theta, \bm{x})$ is the prediction obtained with a model described by its parameters $\theta$. The error estimates are then defined, in the context of ImageNet, as follows:
\begin{align}
    &\text{MSE}(\hat{\bm{y}}, \bm{y}) = \frac{1}{1,000} \sum_{k=0}^{1,000} \big(y_k - \hat{y}_k \big)^2 \,, \\
    &\text{Q}(\hat{\bm{y}}) = \displaystyle \frac{\max_{k\neq c}({\hat{y}_k})}{\max_c({\hat{y}_c})}\,, \\
    &\text{WD}(\hat{\bm{y}}, \bm{y}) = \inf_{\pi \in \mathcal{P}(\hat{\bm{y}}, \bm{y})} \int_{\mathbb{R}\times\mathbb{R}} |\hat{\bm{y}} - \bm{y}| d \pi (\hat{\bm{y}} , \bm{y}) \,,
\end{align}
with $\mathcal{P}(u,v)$ representing the set of probability distributions on $\mathbb{R}\times\mathbb{R}$, where the first factor has marginal distribution $u$ and the second one marginal distribution $v$. Note that the fourth estimate used in the main paper, $1- \max (P(\theta, \bm{x}))$, corresponds to $\frac{1}{2} \text{MAE}(\hat{\bm{y}}, \bm{y})$, since all source images in this study are initially correctly classified by all models. For this reason, we omit the mean absolute distance from the set of measured estimates.

From Table~\ref{tbl:big_Q_filter_transferability_alexnet} to Table~\ref{tbl:big_Q_filter_transferability_vitlarge}, and based on source image filtering, we provide results regarding the transferability and required perturbation for all models considered in this study, when the adversarial examples are generated from the model that has the highest transferability to the model under inspection according to Figure~\ref{fig:transferability_matrix_sup_all}.

\section{Categorical information}

We could observe that a large number of adversarial examples are misclassified into categories that are semantically close to the categories of their source images. This leads to the following question: does a misclassification made for ImageNet, where the prediction is a semantically similar class (i.e., a brown dog breed is misclassified as another brown dog breed), carry the same weight as a misclassification made by an automated system in a self-driving car scenario (i.e., a human or a vehicle not identified)? 

\begin{figure}[hbtp!]
\centering
\begin{tikzpicture}[thick,scale=0.8, every node/.style={scale=0.8}]

\def\xpos{2}
\def\ypos{-1}
\node[] at (\xpos-2.875, \ypos+0.8)  {(617) Lab coat};
\node[] at (\xpos+2.875, \ypos+0.8)  {(697) Pyjama};
\node[inner sep=0pt] (vid2) at (\xpos -4, \ypos-0.6)
    {\includegraphics[width=.18\textwidth]{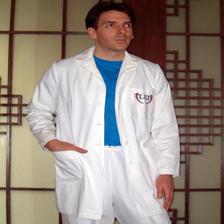}};
\node[inner sep=0pt] (vid2) at (\xpos -1.5, \ypos-0.6)
    {\includegraphics[width=.18\textwidth]{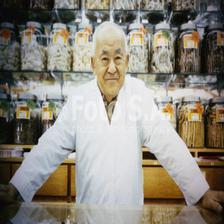}};
\node[inner sep=0pt] (vid2) at (\xpos +1.5, \ypos-0.6)
    {\includegraphics[width=.18\textwidth]{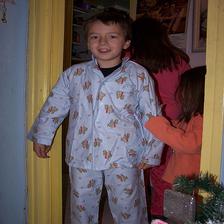}};
\node[inner sep=0pt] (vid2) at (\xpos +4, \ypos-0.6)
    {\includegraphics[width=.18\textwidth]{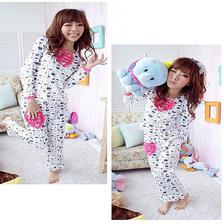}};
\draw[>=triangle 45, ->] (\xpos-1.5, \ypos+0.8) -- (\xpos+1.5, \ypos+0.8);
\def\ypos{-4.1}
\node[] at (\xpos-2.875, \ypos+0.8)  {(861) Toilet seat};
\node[] at (\xpos+2.875, \ypos+0.8)  {(999) Toilet tissue};
\node[inner sep=0pt] (vid2) at (\xpos -4, \ypos-0.6)
    {\includegraphics[width=.18\textwidth]{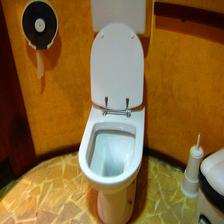}};
\node[inner sep=0pt] (vid2) at (\xpos -1.5, \ypos-0.6)
    {\includegraphics[width=.18\textwidth]{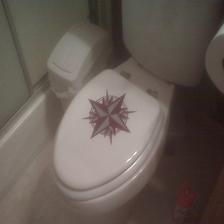}};
\node[inner sep=0pt] (vid2) at (\xpos +1.5, \ypos-0.6)
    {\includegraphics[width=.18\textwidth]{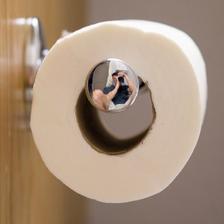}};
\node[inner sep=0pt] (vid2) at (\xpos +4, \ypos-0.6)
    {\includegraphics[width=.18\textwidth]{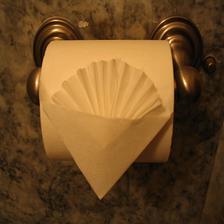}};
\draw[>=triangle 45, ->] (\xpos-1.5, \ypos+0.8) -- (\xpos+1.5, \ypos+0.8);
\def\ypos{-7.2}
\node[] at (\xpos-2.875, \ypos+0.8)  {(369) Siamang};
\node[] at (\xpos+3, \ypos+0.8)  {(381) Spider monkey};
\node[inner sep=0pt] (vid2) at (\xpos -4, \ypos-0.6)
    {\includegraphics[width=.18\textwidth]{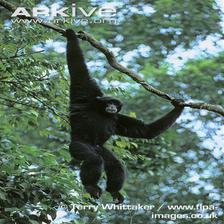}};
\node[inner sep=0pt] (vid2) at (\xpos -1.5, \ypos-0.6)
    {\includegraphics[width=.18\textwidth]{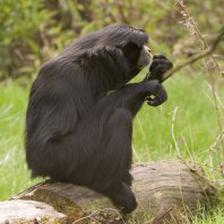}};
\node[inner sep=0pt] (vid2) at (\xpos +1.5, \ypos-0.6)
    {\includegraphics[width=.18\textwidth]{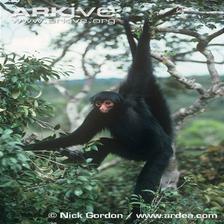}};
\node[inner sep=0pt] (vid2) at (\xpos +4, \ypos-0.6)
    {\includegraphics[width=.18\textwidth]{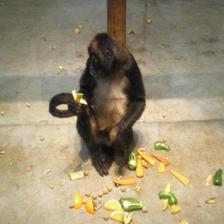}};
\draw[>=triangle 45, ->] (\xpos-1.5, \ypos+0.8) -- (\xpos+1.5, \ypos+0.8);
\def\ypos{-10.3}
\node[] at (\xpos-2.875, \ypos+0.8)  {(966) Red wine};
\node[] at (\xpos+2.875, \ypos+0.8)  {(572) Goblet};
\node[inner sep=0pt] (vid2) at (\xpos -4, \ypos-0.6)
    {\includegraphics[width=.18\textwidth]{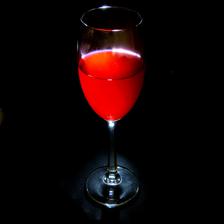}};
\node[inner sep=0pt] (vid2) at (\xpos -1.5, \ypos-0.6)
    {\includegraphics[width=.18\textwidth]{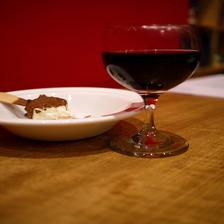}};
\node[inner sep=0pt] (vid2) at (\xpos +1.5, \ypos-0.6)
    {\includegraphics[width=.18\textwidth]{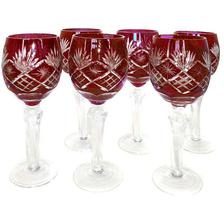}};
\node[inner sep=0pt] (vid2) at (\xpos +4, \ypos-0.6)
    {\includegraphics[width=.18\textwidth]{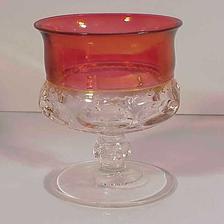}};
\draw[>=triangle 45, ->] (\xpos-1.5, \ypos+0.8) -- (\xpos+1.5, \ypos+0.8);
\def\ypos{-13.4}
\node[] at (\xpos-2.875, \ypos+0.8)  {(146) Albatross};
\node[] at (\xpos+2.875, \ypos+0.8)  {(128) Black stork};
\node[inner sep=0pt] (vid2) at (\xpos -4, \ypos-0.6)
    {\includegraphics[width=.18\textwidth]{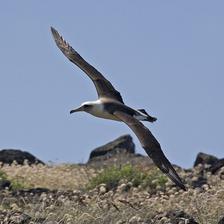}};
\node[inner sep=0pt] (vid2) at (\xpos -1.5, \ypos-0.6)
    {\includegraphics[width=.18\textwidth]{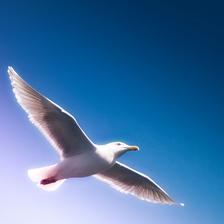}};
\node[inner sep=0pt] (vid2) at (\xpos +1.5, \ypos-0.6)
    {\includegraphics[width=.18\textwidth]{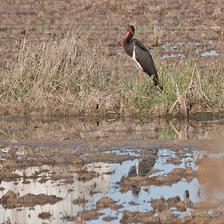}};
\node[inner sep=0pt] (vid2) at (\xpos +4, \ypos-0.6)
    {\includegraphics[width=.18\textwidth]{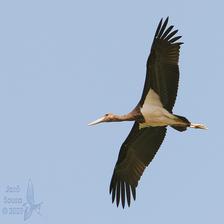}};
\draw[>=triangle 45, ->] (\xpos-1.5, \ypos+0.8) -- (\xpos+1.5, \ypos+0.8);
\def\ypos{-16.5}
\node[] at (\xpos-2.875, \ypos+0.8)  {(159) Rhodesian};
\node[] at (\xpos+2.875, \ypos+0.8)  {(168) Redbone};
\node[inner sep=0pt] (vid2) at (\xpos -4, \ypos-0.6)
    {\includegraphics[width=.18\textwidth]{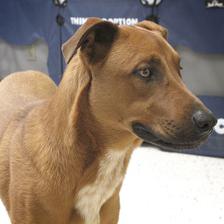}};
\node[inner sep=0pt] (vid2) at (\xpos -1.5, \ypos-0.6)
    {\includegraphics[width=.18\textwidth]{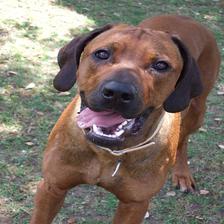}};
\node[inner sep=0pt] (vid2) at (\xpos +1.5, \ypos-0.6)
    {\includegraphics[width=.18\textwidth]{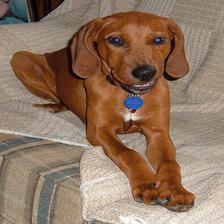}};
\node[inner sep=0pt] (vid2) at (\xpos +4, \ypos-0.6)
    {\includegraphics[width=.18\textwidth]{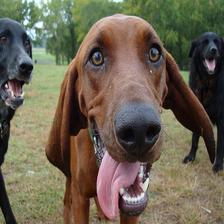}};
\draw[>=triangle 45, ->] (\xpos-1.5, \ypos+0.8) -- (\xpos+1.5, \ypos+0.8);
\def\ypos{-19.6}
\node[] at (\xpos-2.875, \ypos+0.8)  {(636) Maillot};
\node[] at (\xpos+2.875, \ypos+0.8)  {(748) Purse};
\node[inner sep=0pt] (vid2) at (\xpos -4, \ypos-0.6)
    {\includegraphics[width=.18\textwidth]{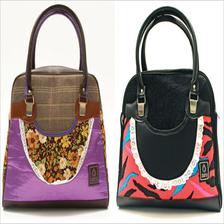}};
\node[inner sep=0pt] (vid2) at (\xpos -1.5, \ypos-0.6)
    {\includegraphics[width=.18\textwidth]{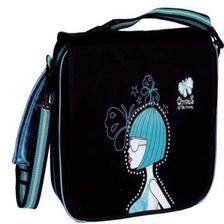}};
\node[inner sep=0pt] (vid2) at (\xpos +1.5, \ypos-0.6)
    {\includegraphics[width=.18\textwidth]{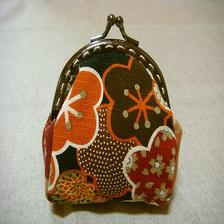}};
\node[inner sep=0pt] (vid2) at (\xpos +4, \ypos-0.6)
    {\includegraphics[width=.18\textwidth]{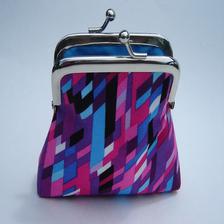}};
\draw[>=triangle 45, ->] (\xpos-1.5, \ypos+0.8) -- (\xpos+1.5, \ypos+0.8);
\end{tikzpicture}
\vspace{2em}
\caption{Adversarial examples shown on the left are misclassified into similar categories shown on the right, by multiple models used in this study.}
\label{fig:imagenet-similar-classes_sup2}  
\end{figure}

In Figure~\ref{fig:imagenet-similar-classes_sup2}, we provide a number of qualitative examples where the adversarial examples on the left are misclassified into the categories on the right. Note that both categories are semantically very similar to each other. As such, we believe an important item for future work is the analysis of misclassification categories, taking into account the semantic similarity of classes. 

\clearpage

\clearpage

\begin{table*}[t!]
\centering
\caption{The lowest, the highest, and the average transferability, as well as the $L_{\{2, \infty\}}$ perturbations, are provided for adversarial examples created by randomly sampling $1,000$ source images $10,000$ times from the datasets provided in the second row. Statistics are provided using adversarial examples that are created from ViT-L and tested on \textbf{AlexNet}.}
\tiny
\vspace{1em}
\begin{tabular}{lllc|cc|cc|cc}
\cmidrule[1pt]{1-10}
~ & ~ &   & \multicolumn{1}{c}{All images}  & \multicolumn{2}{ c}{Hard images} & \multicolumn{2}{c}{Easy (fragile) images} & \multicolumn{2}{c}{Filtered images} \\
    \cmidrule[0.25pt]{4-10}
~ & ~ &  ~ &  $\mathbb{S}$  & $\mathbb{S}_{Q<10}$ & $\mathbb{S}_{Q<25}$ & $\mathbb{S}_{Q>90}$ & $\mathbb{S}_{Q>75}$  & $\mathbb{S} \setminus (\mathbb{S}_{Q<10} \cup \mathbb{S}_{Q>90})$ & $\mathbb{S} \setminus (\mathbb{S}_{Q<25} \cup \mathbb{S}_{Q>75})$ \\
    \cmidrule[0.25pt]{4-10}
    \multicolumn{3}{c}{Source images in set:}  & $19,025$ & $1,904$ & $4,758$ & $1,904$ & $4,758$ & $15,219$ & $9,511$\\
    \cmidrule[0.25pt]{1-10}
    \multirow{9}{*}{\rotatebox[origin=c]{90}{Transferability}} & \multirow{3}{*}{PGD} 
           & Low & $28.1\%$ & $0.4\%$ & $2.2\%$ & $85.4\%$ & $71.0\%$ & $26.7\%$ & $25.8\%$\\
    ~ & ~  & Avg & $34.2\%$ & $1.7\%$ & $4.7\%$ & $88.2\%$ & $75.1\%$ & $31.5\%$ & $30.1\%$\\
    ~ & ~  & High & $40.4\%$ & $1.9\%$ & $6.2\%$ & $90.9\%$ & $79.4\%$ & $36.1\%$ & $33.4\%$\\
    \cmidrule[0.15pt]{2-10}
    ~ & \multirow{3}{*}{CW}
           & Low & $6.1\%$ & $0.0\%$ & $0.0\%$ & $58.7\%$ & $36.3\%$ & $9.1\%$ & $2.3\%$\\
    ~ & ~  & Avg & $12.5\%$ & $0.0\%$ & $0.2\%$ & $62.4\%$ & $41.5\%$ & $10.1\%$ & $4.0\%$\\
    ~ & ~  & High & $18.2\%$ & $0.0\%$ & $0.6\%$ & $66.2\%$ & $48.0\%$ & $12.3\%$ & $5.1\%$\\
    \cmidrule[0.25pt]{2-10}    
    ~ & \multirow{3}{*}{MI-FGSM}
           & Low & $89.5\%$ & $76.2\%$ & $80.3\%$ & $97.2\%$ & $96.2\%$ & $92.6\%$ & $93.4\%$\\
    ~ & ~  & Avg & $94.2\%$ & $80.6\%$ & $83.2\%$ & $98.9\%$ & $97.5\%$ & $94.1\%$ & $94.5\%$\\
    ~ & ~  & High & $98.4\%$ & $84.1\%$ & $85.5\%$ & $99.5\%$ & $99.0\%$ & $96.3\%$ & $96.1\%$\\
    \cmidrule[0.25pt]{1-10}
    \multirow{9}{*}{\rotatebox[origin=c]{90}{\shortstack{Perturbation\\($L_2$ / $L_{\infty}$)}}} & \multirow{3}{*}{PGD} 
           & Low & $7.15$ / $0.07$ & $7.81$ / $0.09$ & $9.08$ / $0.10$ & $5.43$ / $0.04$ & $6.40$ / $0.06$ &  $8.04$ / $0.09$ & $8.71$ / $0.10$\\
    ~ & ~  & Avg & $7.52$ / $0.08$ & $9.76$ / $0.12$ & $9.58$ / $0.12$ & $5.70$ / $0.05$ & $6.73$ / $0.06$ &  $8.59$ / $0.09$ & $9.01$ / $0.10$\\
    ~ & ~ & High & $8.50$ / $0.09$ & $11.3$ / $0.14$ & $10.75$ / $0.13$ & $5.95$ / $0.05$ & $7.01$ / $0.07$ &  $9.07$ / $0.10$ & $9.52$ / $0.11$\\
    \cmidrule[0.15pt]{2-10}
    ~ & \multirow{3}{*}{CW}
           & Low & $2.35$ / $0.07$ & $-$ / $-$ & $2.58$ / $2.58$ & $2.12$ / $0.06$ & $2.31$ / $0.07$ &  $2.78$ / $0.08$ & $2.93$ / $0.09$\\
    ~ & ~  & Avg & $2.69$ / $0.08$ & $-$ / $-$ & $2.95$ / $0.13$ & $2.23$ / $0.07$ & $2.54$ / $0.07$ &  $3.15$ / $0.08$ & $3.41$ / $0.09$\\
    ~ & ~ & High & $3.11$ / $0.09$ & $-$ / $-$ & $4.11$ / $0.14$ & $2.37$ / $0.07$ & $2.75$ / $0.08$ &  $3.41$ / $0.09$ & $3.78$ / $0.10$\\
    \cmidrule[0.25pt]{2-10}    
       ~ & \multirow{3}{*}{MI-FGSM}
           & Low & $18.1$ / $0.07$ & $26.7$ / $0.10$ & $26.1$ / $0.10$&  $10.1$ / $0.03$  & $12.1$ / $0.04$  & $19.3$ / $0.07$  & $19.3$ / $0.07$ \\
    ~ & ~  & Avg & $19.8$ / $0.07$ & $27.1$ / $0.10$ & $26.5$ / $0.10$ &  $10.7$ / $0.03$ & $13.3$ / $0.05$  & $19.8$ / $0.07$  & $19.9$ / $0.07$ \\
    ~ & ~ & High & $20.3$ / $0.07$ & $27.7$ / $0.11$& $27.1$ / $0.10$ &  $11.5$ / $0.04$  & $14.5$ / $0.05$  & $20.6$ / $0.07$  & $20.8$ / $0.07$ \\
\cmidrule[1pt]{1-10}
\end{tabular}
\label{tbl:big_Q_filter_transferability_alexnet}
\end{table*}

\begin{table*}[t!]
\centering
\caption{The lowest, the highest, and the average transferability, as well as the $L_{\{2, \infty\}}$ perturbations, are provided for adversarial examples created by randomly sampling $1,000$ source images $10,000$ times from the datasets provided in the second row. Statistics are provided using adversarial examples that are created from AlexNet and tested on \textbf{SqueezeNet}.}
\tiny
\vspace{1em}
\begin{tabular}{lllc|cc|cc|cc}
\cmidrule[1pt]{1-10}
~ & ~ &   & \multicolumn{1}{c}{All images}  & \multicolumn{2}{ c}{Hard images} & \multicolumn{2}{c}{Easy (fragile) images} & \multicolumn{2}{c}{Filtered images} \\
    \cmidrule[0.25pt]{4-10}
~ & ~ &  ~ &  $\mathbb{S}$  & $\mathbb{S}_{Q<10}$ & $\mathbb{S}_{Q<25}$ & $\mathbb{S}_{Q>90}$ & $\mathbb{S}_{Q>75}$  & $\mathbb{S} \setminus (\mathbb{S}_{Q<10} \cup \mathbb{S}_{Q>90})$ & $\mathbb{S} \setminus (\mathbb{S}_{Q<25} \cup \mathbb{S}_{Q>75})$ \\
    \cmidrule[0.25pt]{4-10}
    \multicolumn{3}{c}{Source images in set:}  & $19,025$ & $1,904$ & $4,758$ & $1,904$ & $4,758$ & $15,219$ & $9,511$\\
    \cmidrule[0.25pt]{1-10}
    \multirow{9}{*}{\rotatebox[origin=c]{90}{Transferability}} & \multirow{3}{*}{PGD} 
           & Low & $41.2\%$ & $3.9\%$ & $9.1\%$ & $92.5\%$ & $82.8\%$ & $41.9\%$ & $39.6\%$\\
    ~ & ~  & Avg & $47.8\%$ & $5.6\%$ & $13.0\%$ & $94.0\%$ & $86.6\%$ & $47.3\%$ & $45.8\%$\\
    ~ & ~  & High & $54.8\%$ & $7.2\%$ & $16.9\%$ & $96.4\%$ & $90.2\%$ & $51.2\%$ & $50.4\%$\\
    \cmidrule[0.15pt]{2-10}
    ~ & \multirow{3}{*}{CW}
           & Low & $61.3\%$ & $22.4\%$ & $34.3\%$ & $95.3\%$ & $90.8\%$  & $63.5\%$  & $65.5\%$\\
    ~ & ~  & Avg & $68.4\%$ & $26.3\%$ & $38.7\%$ & $97.0\%$ & $93.2\%$  & $70.0\%$  & $69.0\%$ \\
    ~ & ~  & High & $74.2\%$ & $30.7\%$ & $44.4\%$ & $98.2\%$ & $96.0\%$  & $73.5\%$  & $73.1\%$ \\
    \cmidrule[0.25pt]{2-10}
    ~ & \multirow{3}{*}{MI-FGSM}
           & Low & $94.1\%$ & $88.4\%$ & $90.1\%$ &  $98.9\%$ & $97.5\%$ & $95.6\%$ & $95.8\%$\\
    ~ & ~  & Avg & $96.2\%$ & $90.3\%$ & $92.2\%$ &  $99.5\%$ & $98.5\%$ & $96.4\%$ & $96.5\%$\\
    ~ & ~ & High & $97.5\%$ & $92.6\%$ & $93.8\%$ &  $99.9\%$ & $99.3\%$ & $97.2\%$ & $97.2\%$\\
    \cmidrule[0.25pt]{1-10}
    \multirow{9}{*}{\rotatebox[origin=c]{90}{\shortstack{Perturbation\\($L_2$ / $L_{\infty}$)}}} & \multirow{3}{*}{PGD} 
           & Low & $6.72$ / $0.05$ & $8.97$ / $0.09$ & $8.73$ / $0.09$ & $4.48$ / $0.03$ & $5.63$ / $0.04$ &  $7.55$ / $0.07$ & $8.03$ / $0.07$\\
    ~ & ~  & Avg & $7.34$ / $0.06$ & $9.61$ / $0.10$ & $9.28$ / $0.10$ & $4.73$ / $0.03$ & $5.95$ / $0.04$ &  $7.93$ / $0.07$ & $8.38$ / $0.07$\\
    ~ & ~ & High & $7.91$ / $0.07$ & $10.2$ / $0.11$ & $9.84$ / $0.11$ & $4.98$ / $0.03$ & $6.29$ / $0.05$ &  $8.30$ / $0.08$ & $8.60$ / $0.08$\\
    \cmidrule[0.15pt]{2-10}
    ~ & \multirow{3}{*}{CW}
           & Low & $7.61$ / $0.11$ & $10.4$ / $0.14$ & $10.26$ / $0.14$ & $4.35$ / $0.09$ & $5.99$ / $0.11$ & $8.34$ / $0.12$  & $8.97$ / $0.12$ \\
    ~ & ~  & Avg & $8.37$ / $0.13$ & $11.3$ / $0.14$ & $10.82$ / $0.14$ & $4.65$ / $0.09$ & $6.04$ / $0.11$ & $8.86$ / $0.13$  &  $9.23$ / $0.13$ \\
    ~ & ~ & High & $9.05$ / $0.13$ & $11.8$ / $0.14$ & $11.35$ / $0.14$ & $4.94$ / $0.10$ & $6.48$ / $0.12$ & $9.35$ / $0.13$  &  $9.61$ / $0.13$ \\
    \cmidrule[0.25pt]{2-10}    
       ~ & \multirow{3}{*}{MI-FGSM}
           & Low & $15.9$ / $0.07$ & $24.7$ / $0.08$ & $22.1$ / $0.07$ &  $7.1$ / $0.02$ & $10.3$ / $0.03$ & $16.7$ / $0.06$ & $16.7$ / $0.06$ \\
    ~ & ~  & Avg & $16.8$ / $0.07$ & $25.3$ / $0.09$ & $22.6$ / $0.08$ &  $8.1$ / $0.02$ & $10.8$ / $0.03$ & $17.1$ / $0.06$ & $17.2$ / $0.06$ \\
    ~ & ~ & High & $17.5$ / $0.08$ & $25.8$ / $0.09$ & $23.5$ / $0.08$ &  $8.6$ / $0.02$ & $11.5$ / $0.04$ & $17.4$ / $0.07$ & $17.5$ / $0.07$ \\
\cmidrule[1pt]{1-10}
\end{tabular}
\label{tbl:big_Q_filter_transferability_squeezenet}
\end{table*}

\begin{table*}[t!]
\centering
\caption{The lowest, the highest, and the average transferability, as well as the $L_{\{2, \infty\}}$ perturbations, are provided for adversarial examples created by randomly sampling $1,000$ source images $10,000$ times from the datasets provided in the second row. Statistics are provided using adversarial examples that are created from DenseNet-121 and tested on \textbf{VGG-16}.}
\tiny
\vspace{1em}
\begin{tabular}{lllc|cc|cc|cc}
\cmidrule[1pt]{1-10}
~ & ~ &   & \multicolumn{1}{c}{All images}  & \multicolumn{2}{ c}{Hard images} & \multicolumn{2}{c}{Easy (fragile) images} & \multicolumn{2}{c}{Filtered images} \\
    \cmidrule[0.25pt]{4-10}
~ & ~ &  ~ &  $\mathbb{S}$  & $\mathbb{S}_{Q<10}$ & $\mathbb{S}_{Q<25}$ & $\mathbb{S}_{Q>90}$ & $\mathbb{S}_{Q>75}$  & $\mathbb{S} \setminus (\mathbb{S}_{Q<10} \cup \mathbb{S}_{Q>90})$ & $\mathbb{S} \setminus (\mathbb{S}_{Q<25} \cup \mathbb{S}_{Q>75})$ \\
    \cmidrule[0.25pt]{4-10}
    \multicolumn{3}{c}{Source images in set:}  & $19,025$ & $1,904$ & $4,758$ & $1,904$ & $4,758$ & $15,219$ & $9,511$\\
    \cmidrule[0.25pt]{1-10}
    \multirow{9}{*}{\rotatebox[origin=c]{90}{Transferability}} & \multirow{3}{*}{PGD} 
           & Low & $27.2\%$ & $3.2\%$ & $7.3\%$ & $72.0\%$ & $56.4\%$ & $27.4\%$ & $28.5\%$\\
    ~ & ~  & Avg & $33.6\%$ & $5.4\%$ & $10.2\%$ & $75.4\%$ & $61.4\%$ & $31.9\%$ & $31.4\%$\\
    ~ & ~  & High & $39.8\%$ & $7.5\%$ & $14.4\%$ & $78.9\%$ & $66.3\%$ & $36.1\%$ & $36.0\%$\\
    \cmidrule[0.15pt]{2-10}
    ~ & \multirow{3}{*}{CW}
           & Low & $12.2\%$ & $0.1\%$ & $1.4\%$ & $51.5\%$ & $33.5\%$ & $9.4\%$ & $9.6\%$\\
    ~ & ~  & Avg & $16.7\%$ & $0.8\%$ & $2.8\%$ & $55.8\%$ & $38.6\%$ & $13.8\%$ & $12.7\%$\\
    ~ & ~  & High & $21.6\%$ & $1.5\%$ & $4.7\%$ & $60.1\%$ & $43.8\%$ & $18.3\%$ & $16.5\%$\\
    \cmidrule[0.25pt]{2-10}  
    ~ & \multirow{3}{*}{MI-FGSM}
           & Low & $87.4\%$ & $77.7\%$ & $80.3\%$ & $94.5\%$ & $91.4\%$ & $89.4\%$ & $89.6\%$\\
    ~ & ~  & Avg & $90.5\%$ & $80.0\%$ & $84.1\%$ & $95.6\%$ & $94.3\%$ & $90.2\%$ & $90.6\%$\\
    ~ & ~ & High & $92.4\%$ & $82.8\%$ & $88.4\%$ & $97.2\%$ & $96.5\%$ & $92.3\%$ & $92.2\%$\\
    \cmidrule[0.25pt]{1-10}
    \multirow{9}{*}{\rotatebox[origin=c]{90}{\shortstack{Perturbation\\($L_2$ / $L_{\infty}$)}}} & \multirow{3}{*}{PGD} 
           & Low & $6.33$ / $0.05$ & $7.95$ / $0.09$ & $7.87$ / $0.08$ & $4.80$ / $0.04$ & $5.61$ / $0.05$ &  $7.06$ / $0.06$ & $7.23$ / $0.06$\\
    ~ & ~  & Avg & $6.93$ / $0.06$ & $8.56$ / $0.09$ & $8.53$ / $0.09$ & $5.06$ / $0.04$ & $5.98$ / $0.05$ &  $7.44$ / $0.07$ & $7.62$ / $0.07$\\
    ~ & ~ & High & $7.41$ / $0.08$ & $9.16$ / $0.10$ & $8.86$ / $0.10$ & $5.30$ / $0.04$ & $6.32$ / $0.06$ &  $7.84$ / $0.08$ & $7.98$ / $0.07$\\
    \cmidrule[0.15pt]{2-10}
    ~ & \multirow{3}{*}{CW}
           & Low & $2.66$ / $0.07$ & $3.93$ / $0.08$ & $3.06$ / $0.08$ & $2.96$ / $0.06$ & $2.55$ / $0.07$ &  $3.08$ / $0.08$ & $3.22$ / $0.08$\\
    ~ & ~  & Avg & $3.10$ / $0.08$ & $4.75$ / $0.10$ & $3.74$ / $0.10$ & $2.46$ / $0.07$ & $2.77$ / $0.07$ &  $3.41$ / $0.08$ & $3.52$ / $0.08$\\
    ~ & ~ & High & $3.50$ / $0.09$ & $5.31$ / $0.14$ & $4.35$ / $0.11$ & $2.61$ / $0.08$ & $3.00$ / $0.08$ &  $3.74$ / $0.09$ & $3.82$ / $0.09$\\
    \cmidrule[0.15pt]{2-10}
       ~ & \multirow{3}{*}{MI-FGSM}
           & Low & $19.7$ / $0.06$ & $25.6$ / $0.09$ & $23.4$ / $0.08$ & $13.1$ / $0.04$ & $15.4$ / $0.05$ & $19.9$ / $0.06$ & $19.9$ / $0.06$ \\
    ~ & ~  & Avg & $20.4$ / $0.07$ & $26.1$ / $0.09$ & $24.7$ / $0.08$ & $13.5$ / $0.04$ & $16.0$ / $0.05$ & $20.3$ / $0.06$ & $20.4$ / $0.06$ \\
    ~ & ~ & High & $21.1$ / $0.07$ & $27.0$ / $0.09$ & $25.8$ / $0.09$ & $14.2$ / $0.05$ & $16.9$ / $0.06$ & $21.0$ / $0.07$ & $21.0$ / $0.07$ \\
\cmidrule[1pt]{1-10}
\end{tabular}
\label{tbl:big_Q_filter_transferability_vgg}
\end{table*}


\begin{table*}[t!]
\centering
\caption{The lowest, the highest, and the average transferability, as well as the $L_{\{2, \infty\}}$ perturbations, are provided for adversarial examples created by randomly sampling $1,000$ source images $10,000$ times from the datasets provided in the second row. Statistics are provided using adversarial examples that are created from DenseNet-121 and tested on \textbf{ResNet-50}.}
\tiny
\vspace{1em}
\begin{tabular}{lllc|cc|cc|cc}
\cmidrule[1pt]{1-10}
~ & ~ &   & \multicolumn{1}{c}{All images}  & \multicolumn{2}{ c}{Hard images} & \multicolumn{2}{c}{Easy (fragile) images} & \multicolumn{2}{c}{Filtered images} \\
    \cmidrule[0.25pt]{4-10}
~ & ~ &  ~ &  $\mathbb{S}$  & $\mathbb{S}_{Q<10}$ & $\mathbb{S}_{Q<25}$ & $\mathbb{S}_{Q>90}$ & $\mathbb{S}_{Q>75}$  & $\mathbb{S} \setminus (\mathbb{S}_{Q<10} \cup \mathbb{S}_{Q>90})$ & $\mathbb{S} \setminus (\mathbb{S}_{Q<25} \cup \mathbb{S}_{Q>75})$ \\
    \cmidrule[0.25pt]{4-10}
    \multicolumn{3}{c}{Source images in set:}  & $19,025$ & $1,904$ & $4,758$ & $1,904$ & $4,758$ & $15,219$ & $9,511$\\
    \cmidrule[0.25pt]{1-10}
    \multirow{9}{*}{\rotatebox[origin=c]{90}{Transferability}} & \multirow{3}{*}{PGD} 
           & Low & $23.9\%$ & $5.2\%$ & $6.9\%$ & $65.8\%$ & $50.1\%$ & $22.3\%$ & $21.5\%$\\
    ~ & ~  & Avg & $29.4\%$ & $7.4\%$ & $9.8\%$ & $69.2\%$ & $55.8\%$ & $27.1\%$ & $25.9\%$\\
    ~ & ~  & High & $35.2\%$ & $9.8\%$ & $13.1\%$ & $72.8\%$ & $61.2\%$ & $32.6\%$ & $30.6\%$\\
    \cmidrule[0.15pt]{2-10}
    ~ & \multirow{3}{*}{CW}
           & Low & $10.3\%$ & $0.8\%$ & $1.6\%$ & $43.8\%$ & $29.0\%$ & $8.7\%$ & $8.4\%$\\
    ~ & ~  & Avg & $15.0\%$ & $1.7\%$ & $3.2\%$ & $48.6\%$ & $33.7\%$ & $12.4\%$ & $11.5\%$\\
    ~ & ~  & High & $19.8\%$ & $2.8\%$ & $5.2\%$ & $52.5\%$ & $39.2\%$ & $16.1\%$ & $15.2\%$\\
    \cmidrule[0.25pt]{2-10}  
    ~ & \multirow{3}{*}{MI-FGSM}
           & Low & $63.1\%$ & $50.1\%$ & $53.3\%$ & $79.5\%$ & $75.7\%$ & $64.5\%$ & $65.6\%$\\
    ~ & ~  & Avg & $68.2\%$ & $53.2\%$ & $57.8\%$ & $81.7\%$ & $79.5\%$ & $69.7\%$ & $69.8\%$\\
    ~ & ~ & High & $72.5\%$ & $56.3\%$ & $62.7\%$ & $84.1\%$ & $82.1\%$ & $74.2\%$ & $72.9\%$\\
    \cmidrule[0.25pt]{1-10}
    \multirow{9}{*}{\rotatebox[origin=c]{90}{\shortstack{Perturbation\\($L_2$ / $L_{\infty}$)}}} & \multirow{3}{*}{PGD} 
           & Low & $6.41$ / $0.06$ & $7.50$ / $0.08$ & $7.47$ / $0.08$ & $5.28$ / $0.04$ & $5.86$ / $0.05$ &  $6.97$ / $0.07$ & $7.09$ / $0.07$\\
    ~ & ~  & Avg & $6.97$ / $0.07$ & $8.01$ / $0.09$ & $8.10$ / $0.09$ & $5.54$ / $0.05$ & $6.25$ / $0.06$ &  $7.39$ / $0.08$ & $7.52$ / $0.08$\\
    ~ & ~ & High & $7.50$ / $0.08$ & $8.53$ / $0.10$ & $8.65$ / $0.10$ & $5.78$ / $0.06$ & $6.49$ / $0.06$ &  $7.09$ / $0.08$ & $7.93$ / $0.08$\\
    \cmidrule[0.15pt]{2-10}
    ~ & \multirow{3}{*}{CW}
           & Low & $2.77$ / $0.07$ & $2.95$ / $0.08$ & $2.97$ / $0.8$ & $2.42$ / $0.05$ & $2.68$ / $0.07$ &  $3.15$ / $0.09$ & $3.22$ / $0.09$\\
    ~ & ~  & Avg & $3.21$ / $0.08$ & $3.41$ / $0.09$ & $3.58$ / $0.9$ & $2.59$ / $0.06$ & $2.91$ / $0.07$ &  $3.50$ / $0.09$ & $3.58$ / $0.09$\\
    ~ & ~ & High & $3.66$ / $0.10$ & $3.89$ / $0.10$ & $4.36$ / $0.11$ & $2.75$ / $0.07$ & $3.18$ / $0.08$ &  $3.83$ / $0.10$ & $3.90$ / $0.10$\\
    \cmidrule[0.15pt]{2-10}
       ~ & \multirow{3}{*}{MI-FGSM}
           & Low & $20.7$ / $0.07$ & $26.7$ / $0.09$ & $25.1$ / $0.09$ & $14.7$ / $0.05$ & $16.9$ / $0.06$ & $21.1$ / $0.07$ & $21.2$ / $0.07$ \\
    ~ & ~  & Avg & $22.2$ / $0.07$ & $27.7$ / $0.09$ & $26.5$ / $0.09$ & $15.5$ / $0.05$ & $17.9$ / $0.06$ & $22.5$ / $0.07$ & $22.6$ / $0.07$ \\
    ~ & ~ & High & $23.6$ / $0.08$ & $28.6$ / $0.10$ & $27.7$ / $0.10$ & $16.4$ / $0.05$ & $19.2$ / $0.06$ & $23.6$ / $0.08$ & $23.5$ / $0.08$ \\
\cmidrule[1pt]{1-10}
\end{tabular}
\label{tbl:big_Q_filter_transferability_resnet}
\end{table*}

\begin{table*}[t!]
\centering
\caption{The lowest, the highest, and the average transferability, as well as the $L_{\{2, \infty\}}$ perturbations, are provided for adversarial examples created by randomly sampling $1,000$ source images $10,000$ times from the datasets provided in the second row. Statistics are provided using adversarial examples that are created from ResNet-50 and tested on \textbf{DenseNet-121}.}
\tiny
\vspace{1em}
\begin{tabular}{lllc|cc|cc|cc}
\cmidrule[1pt]{1-10}
~ & ~ &   & \multicolumn{1}{c}{All images}  & \multicolumn{2}{ c}{Hard images} & \multicolumn{2}{c}{Easy (fragile) images} & \multicolumn{2}{c}{Filtered images} \\
    \cmidrule[0.25pt]{4-10}
~ & ~ &  ~ &  $\mathbb{S}$  & $\mathbb{S}_{Q<10}$ & $\mathbb{S}_{Q<25}$ & $\mathbb{S}_{Q>90}$ & $\mathbb{S}_{Q>75}$  & $\mathbb{S} \setminus (\mathbb{S}_{Q<10} \cup \mathbb{S}_{Q>90})$ & $\mathbb{S} \setminus (\mathbb{S}_{Q<25} \cup \mathbb{S}_{Q>75})$ \\
    \cmidrule[0.25pt]{4-10}
    \multicolumn{3}{c}{Source images in set:}  & $19,025$ & $1,904$ & $4,758$ & $1,904$ & $4,758$ & $15,219$ & $9,511$\\
    \cmidrule[0.25pt]{1-10}
    \multirow{9}{*}{\rotatebox[origin=c]{90}{Transferability}} & \multirow{3}{*}{PGD} 
           & Low & $21.3\%$ & $3.2\%$ & $4.7\%$ & $69.7\%$ & $50.8\%$ & $19.9\%$ & $18.3\%$\\
    ~ & ~  & Avg & $27.7\%$ & $5.4\%$ & $7.8\%$ & $73.2\%$ & $57.4\%$ & $24.8\%$ & $22.9\%$\\
    ~ & ~  & High & $34.0\%$ & $7.5\%$ & $10.7\%$ & $77.7\%$ & $63.0\%$ & $29.7\%$ & $27.0\%$\\
    \cmidrule[0.15pt]{2-10}
    ~ & \multirow{3}{*}{CW}
           & Low & $9.1\%$ & $0.3\%$ & $0.7\%$ & $47.9\%$ & $47.9\%$ & $7.2\%$ & $6.7\%$\\
    ~ & ~  & Avg & $13.6\%$ & $1.2\%$ & $1.9\%$ & $52.7\%$ & $52.7\%$ & $10.5\%$ & $8.7\%$\\
    ~ & ~  & High & $19.1\%$ & $2.3\%$ & $3.4\%$ & $56.6\%$ & $56.6\%$ & $14.0\%$ & $12.0\%$\\
    \cmidrule[0.25pt]{2-10}  
    ~ & \multirow{3}{*}{MI-FGSM}
           & Low & $64.2\%$ & $48.2\%$ & $52.2\%$ & $80.9\%$ & $76.9\%$ & $65.0\%$ & $64.1\%$\\
    ~ & ~  & Avg & $68.6\%$ & $51.7\%$ & $56.8\%$ & $83.5\%$ & $79.3\%$ & $68.8\%$ & $69.4\%$\\
    ~ & ~ & High & $72.3\%$ & $54.5\%$ & $61.5\%$ & $86.2\%$ & $83.5\%$ & $73.0\%$ & $74.6\%$\\
    \cmidrule[0.25pt]{1-10}
    \multirow{9}{*}{\rotatebox[origin=c]{90}{\shortstack{Perturbation\\($L_2$ / $L_{\infty}$)}}} & \multirow{3}{*}{PGD} 
           & Low & $6.09$ / $0.06$ & $7.08$ / $0.07$ & $7.15$ / $0.07$ & $4.83$ / $0.04$ & $5.60$ / $0.05$ &  $6.86$ / $0.07$ & $7.10$ / $0.07$\\
    ~ & ~  & Avg & $6.74$ / $0.07$ & $7.88$ / $0.08$ & $7.91$ / $0.08$ & $5.11$ / $0.04$ & $5.96$ / $0.05$ &  $7.31$ / $0.07$ & $7.51$ / $0.07$\\
    ~ & ~ & High & $7.35$ / $0.08$ & $8.62$ / $0.09$ & $8.58$ / $0.09$ & $5.37$ / $0.05$ & $6.30$ / $0.06$ &  $7.75$ / $0.08$ & $7.93$ / $0.08$\\
    \cmidrule[0.15pt]{2-10}
    ~ & \multirow{3}{*}{CW}
           & Low & $2.44$ / $0.07$ & $2.02$ / $0.06$ & $2.66$ / $0.07$ & $2.18$ / $0.06$ & $2.18$ / $0.06$ &  $2.83$ / $0.08$ & $2.88$ / $0.08$\\
    ~ & ~  & Avg & $2.85$ / $0.08$ & $3.03$ / $0.08$ & $3.39$ / $0.09$ & $2.32$ / $0.06$ & $2.32$ / $0.06$ &  $3.18$ / $0.09$ & $3.21$ / $0.09$\\
    ~ & ~ & High & $3.27$ / $0.09$ & $4.11$ / $0.10$ & $3.99$ / $0.11$ & $2.46$ / $0.07$ & $2.46$ / $0.07$ &  $3.53$ / $0.09$ & $3.58$ / $0.09$\\
    \cmidrule[0.15pt]{2-10}
       ~ & \multirow{3}{*}{MI-FGSM}
           & Low & $20.6$ / $0.07$ & $27.3$ / $0.09$ & $25.7$ / $0.09$ & $13.2$ / $0.04$ & $15.2$ / $0.05$ & $21.2$ / $0.07$ & $21.3$ / $0.07$ \\
    ~ & ~  & Avg & $22.0$ / $0.07$ & $28.5$ / $0.09$ & $27.4$ / $0.09$ & $14.1$ / $0.04$ & $17.7$ / $0.06$ & $22.1$ / $0.08$ & $22.2$ / $0.08$ \\
    ~ & ~ & High & $23.1$ / $0.08$ & $29.6$ / $0.10$ & $28.4$ / $0.10$ & $15.1$ / $0.05$ & $18.5$ / $0.06$ & $23.0$ / $0.08$ & $23.0$ / $0.08$ \\
\cmidrule[1pt]{1-10}
\end{tabular}
\label{tbl:big_Q_filter_transferability_densenet}
\end{table*}

\begin{table*}[t!]
\centering
\caption{The lowest, the highest, and the average transferability, as well as the $L_{\{2, \infty\}}$ perturbations, are provided for adversarial examples created by randomly sampling $1,000$ source images $10,000$ times from the datasets provided in the second row. Statistics are provided using adversarial examples that are created from ViT-L and tested on \textbf{ViT-B}.}
\tiny
\vspace{1em}
\begin{tabular}{lllc|cc|cc|cc}
\cmidrule[1pt]{1-10}
~ & ~ &   & \multicolumn{1}{c}{All images}  & \multicolumn{2}{ c}{Hard images} & \multicolumn{2}{c}{Easy (fragile) images} & \multicolumn{2}{c}{Filtered images} \\
    \cmidrule[0.25pt]{4-10}
~ & ~ &  ~ &  $\mathbb{S}$  & $\mathbb{S}_{Q<10}$ & $\mathbb{S}_{Q<25}$ & $\mathbb{S}_{Q>90}$ & $\mathbb{S}_{Q>75}$  & $\mathbb{S} \setminus (\mathbb{S}_{Q<10} \cup \mathbb{S}_{Q>90})$ & $\mathbb{S} \setminus (\mathbb{S}_{Q<25} \cup \mathbb{S}_{Q>75})$ \\
    \cmidrule[0.25pt]{4-10}
    \multicolumn{3}{c}{Source images in set:}  & $19,025$ & $1,904$ & $4,758$ & $1,904$ & $4,758$ & $15,219$ & $9,511$\\
    \cmidrule[0.25pt]{1-10}
    \multirow{9}{*}{\rotatebox[origin=c]{90}{Transferability}} & \multirow{3}{*}{PGD} 
           & Low & $61.7\%$ & $48.1\%$ & $49.9\%$ & $83.2\%$ & $76.5\%$ & $60.7\%$ & $61.0\%$\\
    ~ & ~  & Avg & $67.2\%$ & $52.6\%$ & $54.4\%$ & $86.0\%$ & $80.8\%$ & $66.6\%$ & $66.5\%$\\
    ~ & ~  & High & $74.0\%$ & $57.1\%$ & $60.3\%$ & $89.0\%$ & $84.7\%$ & $71.4\%$ & $71.1\%$\\
    \cmidrule[0.15pt]{2-10}
    ~ & \multirow{3}{*}{CW}
           & Low & $20.6\%$ & $9.5\%$ & $9.4\%$ & $52.8\%$ & $40.3\%$ & $19.9\%$ & $19.5\%$\\
    ~ & ~  & Avg & $26.7\%$ & $12.3\%$ & $13.4\%$ & $56.9\%$ & $45.5\%$ & $24.7\%$ & $23.9\%$\\
    ~ & ~  & High & $33.4\%$ & $15.2\%$ & $17.5\%$ & $61.4\%$ & $50.4\%$ & $29.5\%$ & $28.3\%$\\
    \cmidrule[0.25pt]{2-10}  
    ~ & \multirow{3}{*}{MI-FGSM}
           & Low & $80.1\%$ & $75.9\%$ & $76.9\%$ & $89.9\%$ & $86.2\%$ & $81.9\%$ & $82.0\%$\\
    ~ & ~  & Avg & $84.6\%$ & $78.4\%$ & $80.2\%$ & $91.2\%$ & $89.5\%$ & $85.2\%$ & $85.3\%$\\
    ~ & ~ & High & $89.2\%$ & $80.5\%$ & $83.5\%$ & $93.5\%$ & $92.1\%$ & $88.4\%$ & $87.7\%$\\

    \cmidrule[0.25pt]{1-10}
    \multirow{9}{*}{\rotatebox[origin=c]{90}{\shortstack{Perturbation\\($L_2$ / $L_{\infty}$)}}} & \multirow{3}{*}{PGD} 
           & Low & $6.49$ / $0.06$ & $7.40$ / $0.07$ & $7.38$ / $0.07$ & $4.94$ / $0.04$ & $5.57$ / $0.05$ &  $6.81$ / $0.06$ & $6.87$ / $0.06$\\
    ~ & ~  & Avg & $6.93$ / $0.07$ & $7.71$ / $0.07$ & $7.70$ / $0.08$ & $5.21$ / $0.04$ & $5.98$ / $0.05$ &  $7.14$ / $0.07$ & $7.20$ / $0.07$\\
    ~ & ~ & High & $7.35$ / $0.07$ & $8.03$ / $0.08$ & $8.04$ / $0.08$ & $5.54$ / $0.05$ & $6.34$ / $0.06$ &  $7.47$ / $0.07$ & $7.54$ / $0.07$\\
    \cmidrule[0.15pt]{2-10}
    ~ & \multirow{3}{*}{CW}
           & Low & $2.39$ / $0.07$ & $2.64$ / $0.08$ & $2.58$ / $0.07$ & $1.98$ / $0.06$ & $2.20$ / $0.06$ &  $2.54$ / $0.08$ & $2.63$ / $0.08$\\
    ~ & ~  & Avg & $2.64$ / $0.08$ & $2.87$ / $0.08$ & $2.88$ / $0.08$ & $2.11$ / $0.06$ & $2.37$ / $0.07$ &  $2.77$ / $0.09$ & $2.82$ / $0.08$\\
    ~ & ~ & High & $2.91$ / $0.09$ & $3.12$ / $0.09$ & $3.15$ / $0.09$ & $2.31$ / $0.07$ & $2.55$ / $0.08$ &  $2.99$ / $0.09$ & $3.05$ / $0.09$\\
    \cmidrule[0.15pt]{2-10}
       ~ & \multirow{3}{*}{MI-FGSM}
           & Low & $15.0$ / $0.05$ & $18.8$ / $0.06$ & $17.0$ / $0.06$ & $11.0$ / $0.04$ & $12.9$ / $0.04$ & $15.2 $ / $0.05$ & $15.0$ / $0.05$ \\
    ~ & ~  & Avg & $16.9$ / $0.05$ & $19.7$ / $0.07$ & $18.2$ / $0.06$ & $11.7$ / $0.04$ & $13.8$ / $0.05$ & $16.4 $ / $0.06$ & $16.2$ / $0.06$ \\
    ~ & ~ & High & $17.5$ / $0.06$ & $19.5$ / $0.07$ & $19.5$ / $0.06$ & $12.3$ / $0.04$ & $14.5$ / $0.05$ & $17.6 $ / $0.06$ & $17.5$ / $0.06$ \\
\cmidrule[1pt]{1-10}
\end{tabular}
\label{tbl:big_Q_filter_transferability_vitbase}
\end{table*}

\begin{table*}[t!]
\centering
\caption{The lowest, the highest, and the average transferability, as well as the $L_{\{2, \infty\}}$ perturbations, are provided for adversarial examples created by randomly sampling $1,000$ source images $10,000$ times from the datasets provided in the second row. Statistics are provided using adversarial examples that are created from ViT-B and tested on \textbf{ViT-L}.}
\tiny
\vspace{1em}
\begin{tabular}{lllc|cc|cc|cc}
\cmidrule[1pt]{1-10}
~ & ~ &   & \multicolumn{1}{c}{All images}  & \multicolumn{2}{ c}{Hard images} & \multicolumn{2}{c}{Easy (fragile) images} & \multicolumn{2}{c}{Filtered images} \\
    \cmidrule[0.25pt]{4-10}
~ & ~ &  ~ &  $\mathbb{S}$  & $\mathbb{S}_{Q<10}$ & $\mathbb{S}_{Q<25}$ & $\mathbb{S}_{Q>90}$ & $\mathbb{S}_{Q>75}$  & $\mathbb{S} \setminus (\mathbb{S}_{Q<10} \cup \mathbb{S}_{Q>90})$ & $\mathbb{S} \setminus (\mathbb{S}_{Q<25} \cup \mathbb{S}_{Q>75})$ \\
    \cmidrule[0.25pt]{4-10}
    \multicolumn{3}{c}{Source images in set:}  & $19,025$ & $1,904$ & $4,758$ & $1,904$ & $4,758$ & $15,219$ & $9,511$\\
    \cmidrule[0.25pt]{1-10}
    \multirow{9}{*}{\rotatebox[origin=c]{90}{Transferability}} & \multirow{3}{*}{PGD} 
           & Low & $38.7\%$ & $23.2\%$ & $27.7\%$ & $69.2\%$ & $57.9\%$ & $38.8\%$ & $37.8\%$\\
    ~ & ~  & Avg & $44.7\%$ & $27.5\%$ & $32.2\%$ & $72.8\%$ & $63.0\%$ &  $43.5\%$ & $42.0\%$\\
    ~ & ~  & High & $51.2\%$ & $30.8\%$ & $37.2\%$ & $77.4\%$ & $69.7\%$ &  $47.3\%$ & $45.4\%$\\
    \cmidrule[0.15pt]{2-10}
    ~ & \multirow{3}{*}{CW}
           & Low & $9.4\%$ & $2.0\%$ & $2.9\%$ & $40.1\%$ & $25.8\%$ & $10.1\%$ & $8.7\%$\\
    ~ & ~  & Avg & $14.6\%$ & $3.8\%$ & $5.3\%$ & $44.2\%$ & $30.8\%$ & $13.5\%$ & $11.0\%$\\
    ~ & ~  & High & $19.2\%$ & $5.4\%$ & $8.0\%$ & $49.7\%$ & $35.7\%$ & $17.7\%$ & $14.2\%$\\
    \cmidrule[0.25pt]{2-10}  
    ~ & \multirow{3}{*}{MI-FGSM}
           & Low & $59.1\%$ & $48.1\%$ & $52.9\%$ & $72.8\%$ & $67.1\%$ & $59.0\%$ & $58.9\%$\\
    ~ & ~  & Avg & $63.6\%$ & $50.3\%$ & $56.2\%$ & $75.7\%$ & $75.5\%$ & $63.5\%$ & $63.1\%$\\
    ~ & ~ & High & $68.2\%$ & $53.7\%$ & $59.5\%$ & $78.1\%$ & $76.6\%$ & $68.1\%$ & $67.4\%$\\
    \cmidrule[0.25pt]{1-10}
    \multirow{9}{*}{\rotatebox[origin=c]{90}{\shortstack{Perturbation\\($L_2$ / $L_{\infty}$)}}} & \multirow{3}{*}{PGD} 
           & Low & $6.00$ / $0.05$ & $6.79$ / $0.07$ & $6.67$ / $0.06$ & $4.68$ / $0.03$ & $5.31$ / $0.04$ &  $6.27$ / $0.06$ & $6.41$ / $0.06$\\
    ~ & ~  & Avg & $6.49$ / $0.06$ & $7.14$ / $0.07$ & $7.10$ / $0.07$ & $4.98$ / $0.04$ & $5.67$ / $0.05$ &  $6.76$ / $0.06$ & $6.88$ / $0.06$\\
    ~ & ~ & High & $7.01$ / $0.07$ & $7.54$ / $0.08$ & $7.49$ / $0.08$ & $5.26$ / $0.04$ & $6.01$ / $0.05$ &  $6.98$ / $0.07$ & $7.14$ / $0.07$\\
    \cmidrule[0.15pt]{2-10}
    ~ & \multirow{3}{*}{CW}
           & Low & $1.88$ / $0.06$ & $2.09$ / $0.08$ & $2.13$ / $0.07$ & $1.72$ / $0.05$ & $1.85$ / $0.06$ & $2.08$ / $0.06$ & $2.02$ / $0.06$\\
    ~ & ~  & Avg & $2.25$ / $0.08$ & $2.56$ / $0.09$ & $2.53$ / $0.08$ & $1.85$ /  $0.05$ & $2.05$ / $0.06$ & $2.42$ / $0.07$ & $2.40$ / $0.07$\\
    ~ & ~ & High & $2.71$ / $0.09$ & $2.91$ / $0.10$ & $2.84$ / $0.09$ & $1.94$ / $0.06$ & $2.87$ / $0.07$ & $2.74$ / $0.08$ & $2.63$ / $0.07$\\
    \cmidrule[0.15pt]{2-10}
       ~ & \multirow{3}{*}{MI-FGSM}
           & Low & $15.5$ / $0.05$ & $19.2$ / $0.07$ & $18.8$ / $0.06$ & $11.2$ / $0.04$ & $13.3$ / $0.04$ & $16.8$ / $0.05$ & $16.4$ / $0.05$ \\
    ~ & ~  & Avg & $17.6$ / $0.06$ & $21.7$ / $0.07$ & $19.2$ / $0.06$ & $12.6$ / $0.04$ & $14.6$ / $0.05$ & $17.5$ / $0.05$ & $17.1$ / $0.05$ \\
    ~ & ~ & High & $18.2$ / $0.06$ & $22.5$ / $0.08$ & $20.5$ / $0.07$ & $13.9$ / $0.04$ & $15.7$ / $0.05$ & $18.1$ / $0.06$ & $18.0$ / $0.06$ \\
\cmidrule[1pt]{1-10}
\end{tabular}
\label{tbl:big_Q_filter_transferability_vitlarge}
\vspace{-2em}
\end{table*}

\end{document}